\documentclass[10pt,twocolumn,letterpaper]{article}

\usepackage{cvpr}              

\usepackage{microtype}

\renewcommand{\paragraph}[1]{\vspace{.5em}\noindent\textbf{#1.}}

\usepackage{amsmath,amssymb,amsthm,booktabs,graphicx}
\usepackage{algorithm}
\usepackage{algpseudocode}
\usepackage{xcolor}
\usepackage{capt-of}   
\usepackage{colortbl}   
\usepackage{adjustbox}  
\usepackage{float}   
\usepackage[table]{xcolor}
\newcommand\best[1]{\cellcolor{red!30}\textbf{#1}}
\newcommand\second[1]{\cellcolor{yellow!45}\underline{#1}}

\definecolor{cvprblue}{rgb}{0.21,0.49,0.74}
\usepackage[pagebackref,breaklinks,colorlinks,allcolors=cvprblue]{hyperref}

\newcommand{\bbR}{\mathbb{R}}

\newcommand{\Enc}{\mathrm{Enc}}
\newcommand{\cG}{\mathcal{G}}          
\newcommand{\cZ}{\mathcal{Z}}          
\newcommand{\cC}{\mathcal{C}}          
\newcommand{\cB}{\mathcal{B}}  

\def\confName{CVPR}
\def\confYear{2027}

\title{Rank Before You Encode: Reading a Few Percent of a Whole-Slide Image}

\author{Haoyu He$^{1}$ \qquad Mahdi S. Hosseini$^{1,2}$\\
$^{1}$ Department of Computer Science and Software Engineering (CSSE), Concordia University, Montreal, Canada\\
$^{2}$ Mila - Quebec AI Institute, Montreal, Canada\\
{\tt\small haoyu.he@mail.concordia.ca}
}

\begin{document}
\maketitle
\begin{abstract}
Classifying a whole-slide image requires encoding thousands of high-resolution tiles with a pathology foundation model, which accounts for most of the inference cost. Existing efficient methods reduce this cost by selecting fewer tiles, but typically train a new model around their selection strategy. A largely unexplored alternative is to keep an existing full-bag model frozen and run it on selected tiles at the same encoder cost. We introduce this comparison, alongside random tile sampling at matched cost. To select tiles, we fit a lightweight selector with 3.3K parameters to the attention of a frozen model, without additional slide labels. The selector ranks regions from a slide thumbnail, encodes the top \(K\) regions at \(5\times\) magnification, and selects \(b\) for processing at \(20\times\). We evaluate seven frozen MIL aggregators on six tasks and compare them with nine efficient methods trained on the same features and splits. At a budget of \((K,b)=(1000,10)\), encoding costs about 6.6\% of the full bag and is \(10\)--\(14\times\) faster, excluding tiling. The AUROC drop is no more than 0.005 in 33 of 36 aggregator-task pairs. At \((500,10)\), the cost falls to about 4\% and inference is \(13\)--\(22\times\) faster, but AUROC drops by up to 0.035 on CAMELYON17, where small metastases are easily missed. Selection improves over random sampling by about 0.10 AUROC on tasks with focal lesions, while random sampling performs well on diffuse tasks. In our comparisons, the frozen model at (500,10) is ahead of nine efficient methods in 55 of 60 paired comparisons. Giving an existing WSI model fewer tiles is thus a baseline that a new efficient model should be compared with at matched encoder cost.

\end{abstract}
    
\section{Introduction}\label{sec:intro}

\begin{figure}[t]\centering\includegraphics[width=\linewidth]{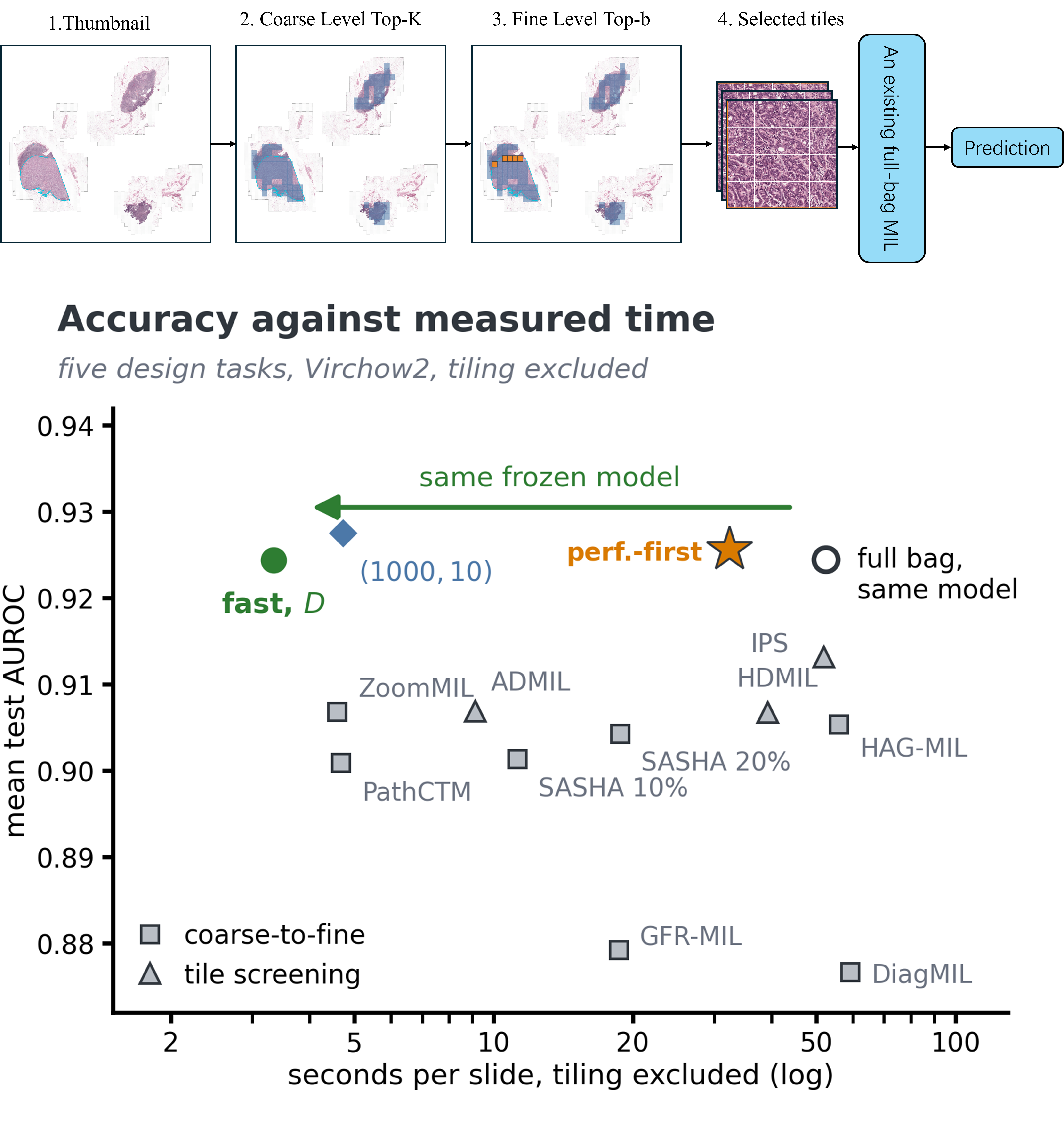}
\caption{(Top)~The testing process. A thumbnail ranking keeps $K$ regions at the coarse level and zooms into $b$, and an existing model reads only their $20\times$ tiles, weights unchanged. (Bottom)~Mean test AUROC against measured seconds per slide over five design tasks (in Table \ref{tab:main}). }\label{fig:teaser}
\end{figure}

\begin{table*}[t]\centering\footnotesize\setlength\tabcolsep{5pt}
\caption{\textbf{Comparison of efficient WSI methods.} Reads first: input used before selection. Budget by: how the budget is determined. Existing: whether the method works with a frozen full-bag model. Trained: trainable parameters (Virchow2, two classes). Labels: need for additional slide labels. Train min: median training time per run across six tasks (Supplementary Table~S10). F\%: encoder FLOPs as a percentage of full-bag cost (range across tasks in Supplementary Table~S1). Performance-first budgets are selected on training and validation slides to match full-bag performance (frozen gated ABMIL, mean over runs).}\label{tab:related}
\begin{tabular}{@{}llccrcrr@{}}
\toprule
Method & Reads first & Budget by & Existing & Trained & Labels & Train min & F\% $\downarrow$ \\
\midrule
ZoomMIL~\cite{thandiackal2022zoommil} & all $5\times$ regions & training & no & 3.4M & yes & 48 to 197 & 7.3 to 8.0 \\
ZoomMIL (3 lev.) & all $2.5\times$ regions & training & no & 5.2M & yes & 30 to 123 & 2.8 to 3.6 \\
HAG-MIL~\cite{xiong2023hagmil} & all levels & training & no & 63.5M & yes & 62 to 313 & 106 \\
GFR-MIL~\cite{gfrmil} & all $5\times$ regions & training & no & 13.3M & yes & 37 to 90 & 34 \\
SASHA~\cite{sasha2025} & all $5\times$ regions & training & no & 56.4M & yes & 661 to 1672 & 15 to 27 \\
PathCTM~\cite{pathctm} & all $5\times$ regions & per slide & no & 179.0M & yes & 107 to 445 & 7.4 to 7.8 \\
DiagMIL~\cite{diagmil} & all $5\times$ regions & per slide & no & 2.4M & yes & 34 to 93 & 106 \\
HDMIL~\cite{dong2025hdmil} & all tiles & training & no & 1.6M & yes & 112 to 1370 & 63 to 94 \\
IPS~\cite{bergner2023ips} & all tiles & training & no & 4.5M & yes & 30 to 115 & 99 \\
ADMIL~\cite{stothers2026admil} & all tiles & training & no & 9.3M & yes & 164 to 831 & 0.2 to 2.4 \\
\midrule
Ours, fast $D$ & a thumbnail & design & yes & \textbf{3.3K} & \textbf{no} & 4 to 26 & 4.1 to 4.7 \\
Ours, performance-first rule & a thumbnail & train+val & yes & \textbf{3.3K} & \textbf{no} & 4 to 26 & 14 to 86 \\
\bottomrule
\end{tabular}
\end{table*}


A whole-slide image (WSI) is commonly classified by encoding every tile at a fine magnification with a pathology foundation model and aggregating the resulting features using multiple instance learning (MIL)~\cite{ilse2018attention,lu2021clam,chen2024uni,zimmermann2024virchow2, yang2026gated}. On our six design tasks a slide has on average $12.2$ to $15.4$ thousand tiles at $20\times$ magnification, and encoding them costs $4.1$ to $5.2$\,PFLOPs and a median of $35$ to $74$ seconds per slide (Supplementary Table~S1 and S2). The encoding accounts for more than $99.9\%$ of the computation, so the way to save is to encode fewer tiles.

Efficient MIL methods do this by training a new aggregator around their own tile selection for one budget~\cite{thandiackal2022zoommil,xiong2023hagmil,pathctm,shin2025wisefuse,dong2025hdmil,stothers2026admil,bergner2023ips}. The saving is uneven, because these methods encode every coarse region or every tile before they choose. In our reruns four of the nine methods (HDMIL, IPS, HAG-MIL and DiagMIL) still cost $63$ to $106\%$ of the full bag (Table~\ref{tab:main}). One comparison is also missing: a laboratory usually already has a full-bag model, trained and checked on its own slides. The question it faces is whether a new efficient model is better than that model, left unchanged and given fewer tiles at the same encoder cost. Two baselines answer it, namely the existing model on tiles picked by a cheap ranking and the existing model on random tiles of the same cost. Without them, a reported saving cannot be separated from what any cheap subset of tiles would give.

\begin{figure*}[t]
  \centering
  \includegraphics[width=\linewidth]{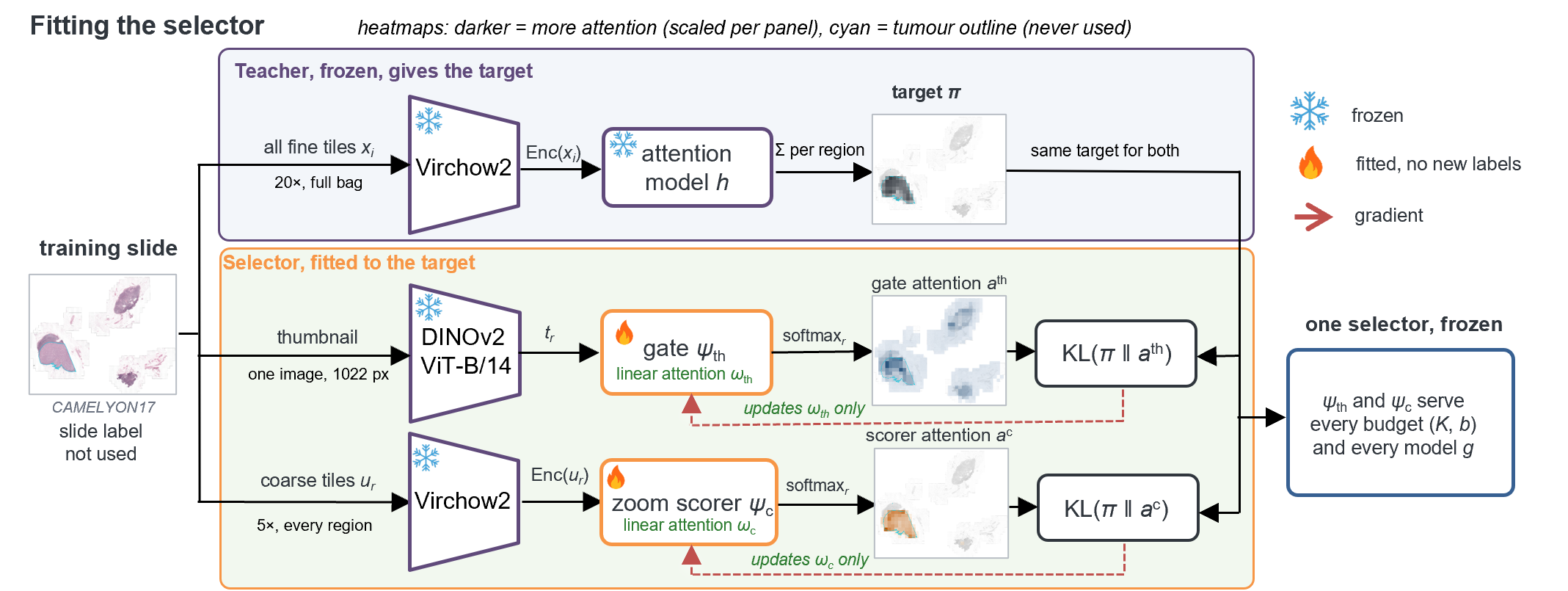}
  \caption{\textbf{Method pipeline}. The two scorers are fitted once per training split, with no new labels. A frozen attention model $h$ (the teacher, the split's gated ABMIL) reads the full bag of fine tiles of a training slide, and its attention, summed over each region's fine tiles and normalised over regions, is the target $\pi$. The thumbnail scorer $\psi_{\mathrm{th}}$, on DINOv2 thumbnail tokens $t_r$, and the coarse scorer $\psi_{\mathrm c}$, on coarse encodings $\Enc(u_r)$, are linear attention models fitted to this one target by minimising $\mathrm{KL}(\pi\,\|\,a)$ (Alg.~\ref{alg:fit}). Gradients (dashed) update only the weights $\omega_{\mathrm{th}}$ and $\omega_{\mathrm c}$, and the encoders and $h$ stay frozen. The fitted selector is then frozen and serves every budget $(K,b)$ and every model $g$ (Fig.~\ref{fig:teaser}, top). Shown is a CAMELYON17 training slide of split~1 ($468$ regions, $6{,}077$ fine tiles), darker meaning more attention, scaled per panel. The tumour outline (cyan) is for illustration and is never seen by any model.}
  \label{fig:pipeline}
\end{figure*}

We study how few tiles an existing model needs. A frozen attention model already marks the regions it relies on, and we find that a linear scorer can predict this attention before the tiles are encoded. Our selector has two such scorers ($3{,}328$ weights). The first ranks the regions of a slide from its thumbnail, and the top $K$ regions are encoded at $5\times$. The second ranks those regions from their $5\times$ features, and the top $b$ are encoded at $20\times$. The existing model sees only these $20\times$ tiles, with its weights unchanged (Fig.~\ref{fig:teaser}, Fig.~\ref{fig:pipeline}). Both scorers are fitted once to the attention of a frozen gated ABMIL on the training slides, with no new labels, and the same fit serves every budget $(K,b)$ and every aggregator. Fitting takes 4 to 26 minutes per split, compared with 30 minutes to 28 hours for efficient baselines trained for a single budget (Table~\ref{tab:related}).

We evaluate seven frozen MIL aggregators on six design tasks with two encoders, and on three held-out tasks fixed in advance. We compare each reading with its full-bag counterpart, random tile selection at the same encoder cost, the model's own attention, and nine efficient methods trained on identical features and splits. We report three findings.

First, existing models maintain most of their performance at a small fraction of encoder cost (Table~\ref{tab:main}). At $(K,b)=(1000,10)$, about $6.6\%$ of the full-bag encoder cost and $10\times$ to $14\times$ faster with tiling excluded, six frozen aggregators remain within $0.005$ AUROC of their full-bag performance in 33 of 36 comparisons. The budget was chosen on the design results of gated ABMIL alone and the other five aggregators, read only after it was fixed, are within $0.005$ in 27 of 30 pairs (Supplementary section I).

Second, going below the coarse level costs the smallest lesions. At $(500,10)$, about $4\%$ of the cost and $13\times$ to $22\times$ faster, the count falls to 22 of 36. On CAMELYON17 the frozen models lose up to $0.035$ AUROC, mostly on slides whose only lesions are isolated tumour cells, and, at a validation threshold set for $98\%$ sensitivity, $0.02$ to $0.19$ of specificity (Supplementary Table S27). A budget chosen on a model's own validation slides holds on average out of sample, but when the selector was moved between lymph-node cohorts it chose a budget that lost $0.017$ AUROC on CAMELYON16.

Third, efficient methods rarely outperform the model they would replace (Table~\ref{tab:main}). In our reruns, the frozen gated ABMIL at about $4\%$ is ahead of ten settings of nine efficient methods in 55 of 60 paired comparisons. Its own full bag is ahead in 59, so the selection keeps the lead of the full-bag model and does not create it. Part of the gap may be tuning, which this comparison does not separate.

We suggest that an efficient WSI method be compared with the existing full-bag model on selected and on random tiles at the same encoder cost. Code, splits, and the selector will be released upon acceptance.

\section{Related Work}\label{sec:related}

\textbf{Coarse-to-fine reading.}
ZoomMIL zooms with a differentiable top-$k$~\cite{thandiackal2022zoommil}, HAG-MIL builds an attention hierarchy~\cite{xiong2023hagmil}, GFR-MIL by glance, focus and reflect steps~\cite{gfrmil}, DiagMIL by an uncertainty gate between a coarse and a fine stage~\cite{diagmil} and PathCTM by reasoning across scales with a stopping rule~\cite{pathctm}, and agents choose regions sequentially~\cite{zhao2023rlogist,sasha2025}. WISE-FUSE scores every coarse patch against class descriptions with a pathology vision-language model and encodes the fine tiles of the best regions~\cite{shin2025wisefuse}, and the multiscale form of DSMIL pairs each coarse feature with the features of every fine tile inside it~\cite{li2021dsmil}.
Despite their different selection strategies, existing coarse-to-fine WSI methods share three limitations. First, they encode every coarse region before deciding which regions to process at fine resolution. This creates a minimum encoder cost determined by the coarse magnification, equal to $(s_{\mathrm c}/s_{\mathrm f})^2$ of the full-bag cost. With a $20\times$ fine level, this corresponds to $6.25\%$ at $5\times$ and $1.6\%$ at $2.5\times$. WISE-FUSE has a similar cost floor, with an additional vision-language encoder, while multiscale DSMIL encodes all tiles at both magnifications. Second, their budgets are either fixed during training or determined by the model's own stopping rule. Changing the accuracy-cost trade-off may therefore require another model or additional training. Third, each method trains its own aggregator around its selection strategy, so it cannot be directly added to an existing full-bag model. Our method addresses these limitations by ranking regions from a thumbnail before encoding any tiles. This allows us to use different budgets without encoding every coarse region or changing the weights of an existing MIL model.

\textbf{Screening tiles before encoding.}
HDMIL screens tiles with a small network distilled from a full-bag teacher~\cite{dong2025hdmil}, the prior work closest to fitting our selector to a frozen model, ADMIL scores the pixels of every tile with a small network and trains an aggregator per budget~\cite{stothers2026admil}, and IPS selects patches iteratively over all of them~\cite{bergner2023ips}.
\emph{Limitation:} they read and score every fine tile, and their budget is again fixed when they are trained. ADMIL, the cheapest network on every tile, reaches lower FLOPs than our default ($0.2$ to $2.4\%$), but with its scorer on $20\times$ tiles, as we run it, its measured time is bound by reading and decoding every tile ($3.6$ to $13.5\times$ faster than the full bag, tiling excluded, against $13$ to $22\times$ for ours). A distilled encoder~\cite{filiot2025h0mini} makes each evaluation cheaper but meets the same bound. Looking at every tile can find lesions a coarse ranking misses. Our method first ranks regions using a thumbnail and then selects tiles at the coarse level. The two scorers are fitted once and can be used with an existing frozen MIL aggregator without retraining it. We also evaluate ADMIL in settings where it performs better than our method and where it does not (Sec.~\ref{sec:results}).


\section{Protocol and instrument}\label{sec:method}

\textbf{Comparing at the same cost. }\label{sec:protocol}
An efficient WSI method reads a slide at some encoder cost and replaces a full-bag model $g$ that a laboratory already has. We compare these approaches with three readings of the same test slides. First, we measure the performance of $g$ when it processes the full bag of fine tiles. Second, we run the same mode $g$, with its weight unchanged on fine tiles selected by an efficient ranking method at the same encoder cost. This tells us whether training a separate efficient model provides an advantage. Third, we run $g$ on random fine tiles of the same encoder cost. This allows us to measure the benefit of selecting tiles intelligently rather than randomly. Cost counts encoder evaluations and every other network a method runs, per slide (Sec.~\ref{sec:setup}). We consider the mean AUROC to be maintained if it is no more than 0.005 below the full-bag result. For each individual aggregator and task, we use a non-inferiority margin of 0.010, which was fixed in advance. The second comparison requires a ranking method that is inexpensive and does not change the existing model $g$. Our selector is described below.
We also test the frozen model's own attention and the tile selections made by four existing efficient methods when they are used with $g$ at matched costs (Supplementary section G).

\textbf{Setting and cost}\label{sec:setting}
We consider a two-level reading process. The first level provides a coarse view, while the second provides a finer view that is $s$ times higher in resolution.
Each slide is divided into $N$ coarse regions with tissue. Every region $r$ has a coarse view $u_r$ (one tile at the coarse level) and up to $m=s^2$ fine patches $\{x_i: i\in\cC(r)\}$. A frozen encoder $\Enc$ maps both to $\bbR^d$. A region is zoomable if it has at least one fine patch. In our experiments the coarse level is $5\times$ ($256$ pixels, about $0.5$\,mm), the fine level $20\times$ ($256$ pixels, about $128\,\mu$m). Therefore, the fine level is four times higher in resolution, giving $s=4$ and $m=16$, and the encoder is Virchow2~\cite{zimmermann2024virchow2}, with UNI2-h~\cite{chen2024uni,mahmood2025uni2h} as the second encoder.
The full bag encodes every fine patch. Under a budget defined by $(K,b)$, the selector first keeps $K$ regions and then chooses $b$ of those regions for high-resolution processing. The number of tile-encoder evaluations is:
\begin{equation}\label{eq:cost}
  \operatorname{cost}(K,b)=K+\textstyle\sum_{r\in\cZ_{K,b}}|\cC(r)|\;\le\;K+mb
\end{equation}
If every zoomable gated region is zoomed, their coarse views are not needed. We use $T$ to represent the budget that processes every region retained by the gate that contains fine patches ($99$ to $100\%$ of the full bag). The small difference occurs because some fine patches do not have a corresponding coarse region when the two magnification levels are tiled independently.
A method that encodes every coarse region before zooming costs at least $N/\sum_r|\cC(r)|\approx1/m$ of the full bag, which is about $1/16$ of the full-bag with our two levels. This lower bound becomes smaller when the first level is even coarser. It does not apply to methods that rank regions without first encoding every coarse region. Our method uses a thumbnail to perform this initial ranking in a single pass. Its cost is fixed for each slide rather than increasing with the number of regions.

\textbf{Choosing where to look}
\emph{Thumbnail gate.} The slide thumbnail is encoded once at $1022$ pixels on the long side by DINOv2 ViT-B/14~\cite{oquab2024dinov2} (about $2$\,TFLOPs, the cost of six Virchow2 evaluations), and each region takes the token containing its centre. A linear attention scorer $\psi_{\mathrm{th}}$ ranks all regions and the gate keeps the top $K$, $\cG_K$. The thumbnail provides a low-resolution overview of the slide, but it cannot reliably locate individual lesions. Its purpose is to keep potentially informative tissue regions, not to identify lesions directly.
\emph{Coarse scorer and zoom set.} After applying the thumbnail gate, we encode the retained regions at the coarse level using $\Enc(u_r)$.
A second scorer $\psi_{\mathrm c}$ ranks these regions and the $b$ zoomable regions of $\cG_K$ with the highest score are zoomed,
\begin{equation}\label{eq:select}
\begin{aligned}
  \cG_K&=\operatorname{top}_K\{\psi_{\mathrm{th}}(t_r)\},\\
  \cZ_{K,b}&=\operatorname{top}_b\{\psi_{\mathrm c}(\Enc(u_r)): r\in\cG_K,\ \cC(r)\neq\emptyset\},
\end{aligned}
\end{equation}
with $t_r$ the region's thumbnail token.
The two scorers work at different levels of detail. The thumbnail is inexpensive but too coarse to reliably identify individual lesions. It is used to retain regions that may contain useful information. The coarse scorer then uses pathology features to decide which of these regions should be tested at high resolution. In this way, the thumbnail gate prioritizes coverage, while the coarse scorer makes a more detailed selection. Neither scorer evaluates regions outside the thumbnail gate.

\textbf{Fitting the scorers.} Both scorers are attention-pooling models, $a_r=\operatorname{softmax}_r(\omega^\top v_r)$, fitted on the training slides of each split and then frozen. We fit them to a frozen attention MIL model $h$ of the task, the teacher, which need not be the model $g$ that reads the selection. The target of a training slide is $h$'s attention on its fully zoomed bag $T$, summed over each region's fine tiles and normalised over its regions, and both scorers minimise $\mathrm{KL}(\text{target}\,\|\,a)$ (Alg.~\ref{alg:fit}, Fig.~\ref{fig:pipeline}). This needs no labels beyond those $h$ was trained with, but it needs the training slides' fine features, and it teaches the selector where $h$'s fine-level attention falls, seen from the thumbnail and the coarse level. We use the frozen gated ABMIL of each split as $h$, and the same selector is read by every aggregator $g$, those without attention included (Sec.~\ref{sec:results}). If the deployed model has no attention, a gated ABMIL trained on its cached features serves as $h$. Fitted to each aggregator's own importance instead it is no better (Supplementary section F).

\begin{algorithm}[t]
\caption{Fitting the selector (once per training split, no new labels)}\label{alg:fit}
\begin{algorithmic}[1]
\Require frozen attention model $h$, training slides
\For{each training slide}
  \State $\alpha\gets$ attention of $h$ on its fully zoomed bag $T$
  \State $\pi_r\gets\sum_{i\in\cC(r)}\alpha_i$, normalised over $r$ \Comment{no label}
\EndFor
\State $\omega_{\mathrm{th}}\gets\arg\min_\omega\sum\mathrm{KL}\big(\pi\,\|\,\operatorname{softmax}_r(\omega^\top t_r)\big)$
\State $\omega_{\mathrm c}\gets\arg\min_\omega\sum\mathrm{KL}\big(\pi\,\|\,\operatorname{softmax}_r(\omega^\top\Enc(u_r))\big)$
\State \Return $\psi_{\mathrm{th}},\psi_{\mathrm c}$, frozen, for every model $g$
\end{algorithmic}
\end{algorithm}

\begin{figure*}[t]\centering\includegraphics[width=\linewidth]{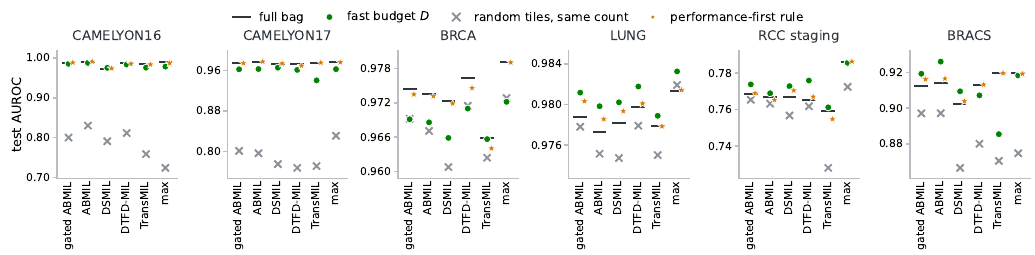}
\caption{\textbf{Frozen full-bag models read through our selection}, no retraining and no new labels (Virchow2, test AUROC, mean over 25 runs, BRACS 5). Per task and aggregator, four reads: the model's own full bag, the fast budget $D$ (about $4\%$ of the encoder FLOPs) through our selection, as many random $20\times$ tiles as $D$ reads (matched in count, not FLOPs), and the budget the performance-first rule chose.TransMIL reads a bag as a sequence, here in our gate order, so its full bag also depends slightly on the ranking. Mean pooling, whose full bag fails on lymph nodes, and all numbers are in Supplementary Table S14.}\label{fig:frozen}
\end{figure*}

\textbf{Reading an existing model}\label{sec:frozen}
A model $g$ trained on full bags of fine patches reads, at budget $(K,b)$, only the fine patches of the zoomed regions, with the same weights, so the aggregator applies unchanged. Its cost is Eq.~\eqref{eq:cost}, since the $K$ coarse views are needed by the coarse scorer. $T$ ($K=N$, every zoomable region zoomed) is $g$'s full bag up to the $\le1\%$ of fine patches without a coarse parent, and is what we report as the frozen model's full bag. As controls, fine patches are drawn uniformly from the slide, as many as the selection reads, or as many as its whole encoder cost buys (coarse encodings and thumbnail included).
\begin{equation}\label{eq:finebag}
  \cB^{\mathrm f}_{K,b}=\{\Enc(x_i) : i\in\cC(r),\ r\in\cZ_{K,b}\},
\end{equation}

\textbf{Inference.} With the budget fixed before deployment, a test slide is read in four steps. Its thumbnail is encoded and $\psi_{\mathrm{th}}$ keeps $\cG_K$, the $K$ regions are encoded at the coarse level and $\psi_{\mathrm c}$ picks $\cZ_{K,b}$ (Eq.~\eqref{eq:select}), their fine patches are encoded, and $g$ reads $\cB^{\mathrm f}_{K,b}$ (Eq.~\eqref{eq:finebag}).

\textbf{Two ways to set the budget.}\label{sec:budgetrule} We evaluate two strategies for choosing the budget. The fast budget reads every slide at $D=(500,10)$, fixed once for all tasks during design and before the held-out tasks were trained. It is not tuned. A second named point, $(1000,10)$, doubles the gate. We selected this second budget based on the development results of gated ABMIL and then used it for the other aggregators and held-out settings (Supplementary section I). The performance-first rule chooses each model's budget from its own training and validation slides, walking down a ladder of budgets from the full bag while a bootstrap bound on the AUROC loss stays within a margin. This approach follows the general idea of fixed-sequence testing ~\cite{maurer1995multiple,angelopoulos2021ltt} but without its error-rate guarantee (Supplementary section H). We fixed the selection rule before reporting its results.

\section{Experimental Setup}\label{sec:setup}

\begin{table*}[htbp]\centering\footnotesize\setlength\tabcolsep{2.4pt}\renewcommand\arraystretch{0.85}
\caption{\textbf{Accuracy and cost} (Virchow2, UNI2-h in supplement Table S16), on five tasks (BRCA in supplement Table S15). AUROC: test, mean over 25 runs (5 on BRACS). F\%: FLOPs per slide as a share of the full bag's, every network included. $\times$: measured speed-up over the full bag, from reading tiles to the prediction, tissue segmentation and tiling excluded, as the ratio of summed seconds over 20 test slides (with tiling in Supplementary Table S7). Our method uses the existing full-bag gated ABMIL model, without retraining, on the $20\times$ tiles selected at $D=(500,10)$. Random tiles use the same encoder FLOPs of our selection at $D$, including coarse encodings and the thumbnail. ZoomMIL, 3 levels: $2.5\times\!\to\!5\times\!\to\!20\times$.
Both HAG-MIL and DiagMIL encode every $5\times$ region and then all $20\times$ tiles within those regions, requiring a full-bag pass plus an additional coarse-level pass. Consequently, their computational cost reaches 106\% of the full-bag baseline.
The two full-bag gated ABMIL rows are separately trained models, using different input pipelines. BRCA, the one task where efficient methods (IPS, ADMIL, SASHA) are ahead of the frozen model at $D$, is in Supplementary Table S15. Bold on red and underline on yellow indicate the best and second best result per task. Mean is averaged over the five tasks shown. Subscripts give the standard deviation over runs. }\label{tab:main}
\resizebox{\textwidth}{!}{\begin{tabular}{@{}l@{\hspace{4pt}}crr|@{\hspace{4pt}}crr|@{\hspace{4pt}}crr|@{\hspace{4pt}}crr|@{\hspace{4pt}}crr|@{\hspace{4pt}}c@{}}
\toprule
 & \multicolumn{3}{c}{LUNG} & \multicolumn{3}{c}{RCC staging} & \multicolumn{3}{c}{BRACS} & \multicolumn{3}{c}{CAMELYON16} & \multicolumn{3}{c}{CAMELYON17} & \\ \cmidrule(lr){2-4}\cmidrule(lr){5-7}\cmidrule(lr){8-10}\cmidrule(lr){11-13}\cmidrule(lr){14-16}
Method & AUROC  & F\% $\downarrow$ & $\times$ $\uparrow$ & AUROC  & F\% $\downarrow$ & $\times$ $\uparrow$ & AUROC  & F\% $\downarrow$ & $\times$ $\uparrow$ & AUROC  & F\% $\downarrow$ & $\times$ $\uparrow$ & AUROC  & F\% $\downarrow$ & $\times$ $\uparrow$ & Mean $\uparrow$\\
\midrule\multicolumn{17}{@{}c}{\textbf{Virchow2}} \\ \midrule

gated ABMIL (baseline code) & 0.979$_{\pm.018}$ & 100 & 1.0 & 0.770$_{\pm.048}$ & 100 & 1.0 & 0.905$_{\pm.013}$ & 100 & 1.0 & \best{0.990}$_{\pm.006}$ & 100 & 1.0 & \second{0.975}$_{\pm.023}$ & 100 & 1.0 & 0.9238 \\
\midrule
ZoomMIL~\cite{thandiackal2022zoommil} & 0.970$_{\pm.030}$ & 8.0 & 10.9 & 0.762$_{\pm.052}$ & 7.3 & 11.0 & 0.904$_{\pm.028}$ & 7.8 & 11.8 & 0.975$_{\pm.007}$ & 7.7 & 13.0 & 0.923$_{\pm.037}$ & 7.7 & 12.5 & 0.9068 \\
ZoomMIL, 3 levels & 0.975$_{\pm.016}$ & 3.6 & 20.8 & 0.745$_{\pm.052}$ & 2.8 & 24.3 & 0.888$_{\pm.015}$ & 3.4 & 27.2 & 0.890$_{\pm.044}$ & 3.5 & 28.0 & 0.884$_{\pm.052}$ & 3.4 & 27.2 & 0.8765 \\
PathCTM~\cite{pathctm} & 0.970$_{\pm.031}$ & 7.8 & 10.2 & 0.745$_{\pm.055}$ & 7.4 & 10.3 & 0.887$_{\pm.018}$ & 7.6 & 11.6 & 0.975$_{\pm.013}$ & 7.6 & 12.5 & 0.927$_{\pm.046}$ & 7.6 & 12.5 & 0.9009 \\
HDMIL~\cite{dong2025hdmil} & 0.974$_{\pm.021}$ & 81 & 1.2 & 0.759$_{\pm.054}$ & 73 & 1.4 & 0.870$_{\pm.029}$ & 94 & 1.1 & 0.975$_{\pm.035}$ & 64 & 1.5 & 0.955$_{\pm.036}$ & 63 & 1.6 & 0.9067 \\
ADMIL~\cite{stothers2026admil} & 0.965$_{\pm.021}$ & 0.3 & 3.6 & 0.758$_{\pm.060}$ & 0.7 & 3.6 & 0.887$_{\pm.012}$ & 0.2 & 4.6 & 0.964$_{\pm.026}$ & 1.1 & 13.5 & 0.961$_{\pm.030}$ & 2.4 & 8.2 & 0.9069 \\
HAG-MIL~\cite{xiong2023hagmil} & 0.968$_{\pm.024}$ & 106 & 0.9 & 0.746$_{\pm.043}$ & 106 & 0.9 & 0.868$_{\pm.033}$ & 106 & 0.9 & 0.979$_{\pm.008}$ & 106 & 0.9 & 0.965$_{\pm.030}$ & 106 & 0.9 & 0.9053 \\
IPS~\cite{bergner2023ips} & 0.969$_{\pm.029}$ & 99 & 1.0 & 0.763$_{\pm.054}$ & 99 & 1.0 & 0.895$_{\pm.010}$ & 99 & 1.0 & 0.980$_{\pm.011}$ & 99 & 1.0 & 0.958$_{\pm.037}$ & 99 & 1.0 & 0.9132 \\
GFR-MIL~\cite{gfrmil} & 0.972$_{\pm.022}$ & 34 & 2.7 & 0.758$_{\pm.058}$ & 34 & 2.6 & 0.880$_{\pm.045}$ & 34 & 2.7 & 0.883$_{\pm.063}$ & 34 & 2.9 & 0.904$_{\pm.047}$ & 34 & 3.1 & 0.8792 \\
DiagMIL~\cite{diagmil} & 0.976$_{\pm.018}$ & 106 & 0.9 & 0.728$_{\pm.068}$ & 106 & 0.9 & 0.857$_{\pm.016}$ & 106 & 0.9 & 0.897$_{\pm.122}$ & 106 & 0.9 & 0.925$_{\pm.037}$ & 106 & 0.9 & 0.8767 \\
SASHA~\cite{sasha2025}, 10\% & 0.976$_{\pm.016}$ & 17 & 4.3 & 0.761$_{\pm.075}$ & 15 & 4.3 & 0.873$_{\pm.014}$ & 17 & 4.6 & 0.952$_{\pm.019}$ & 17 & 5.4 & 0.945$_{\pm.040}$ & 17 & 5.0 & 0.9013 \\
SASHA~\cite{sasha2025}, 20\% & 0.976$_{\pm.019}$ & 27 & 2.6 & 0.760$_{\pm.068}$ & 25 & 2.6 & 0.878$_{\pm.012}$ & 26 & 2.7 & 0.953$_{\pm.019}$ & 26 & 3.3 & 0.953$_{\pm.024}$ & 27 & 3.0 & 0.9043 \\
\midrule

gated ABMIL in our pipeline (frozen), full bag & 0.979$_{\pm.020}$ & 100 & 1.0 & 0.768$_{\pm.056}$ & 100 & 1.0 & 0.913$_{\pm.014}$ & 100 & 1.0 & \second{0.988}$_{\pm.009}$ & 100 & 1.0 & 0.974$_{\pm.020}$ & 100 & 1.0 & 0.9245 \\
frozen gated ABMIL, random tiles, same FLOPs & 0.978$_{\pm.020}$ & 4.7 & 18.3 & 0.768$_{\pm.056}$ & 4.3 & 17.1 & 0.908$_{\pm.013}$ & 4.5 & 21.3 & 0.887$_{\pm.025}$ & 4.4 & 22.3 & 0.868$_{\pm.038}$ & 4.2 & 22.7 & 0.8819 \\

frozen gated ABMIL, our selection, $D$ & \best{0.981}$_{\pm.016}$ & 4.7 & 13.3 & \best{0.774}$_{\pm.039}$ & 4.3 & 13.3 & \best{0.919}$_{\pm.012}$ & 4.5 & 17.8 & 0.985$_{\pm.011}$ & 4.4 & 21.1 & 0.962$_{\pm.033}$ & 4.2 & 21.8 & 0.9244 \\

frozen gated ABMIL, our selection, $(1000,10)$ & \best{0.981}$_{\pm.016}$ & 6.8 & 10.0 & \second{0.771}$_{\pm.040}$ & 6.5 & 10.2 & \second{0.918}$_{\pm.012}$ & 6.7 & 12.5 & \best{0.990}$_{\pm.005}$ & 6.6 & 14.2 & \best{0.978}$_{\pm.022}$ & 6.6 & 13.9 & \best{0.9275} \\

frozen gated ABMIL, performance-first rule & \second{0.980}$_{\pm.018}$ & 17 & 4.8 & 0.769$_{\pm.056}$ & 86 & 1.1 & 0.916$_{\pm.016}$ & 76 & 1.3 & \second{0.988}$_{\pm.009}$ & 14 & 6.9 & 0.974$_{\pm.020}$ & 81 & 1.2 & \second{0.9256} \\

\bottomrule
\end{tabular}}
\end{table*}

\textbf{Tasks.}
Six design tasks were used to build the selection and the fast budget. They are lymph-node metastasis detection on CAMELYON16~\cite{bejnordi2017camelyon16} (399 slides) and CAMELYON17~\cite{litjens2018camelyon} (499, where slides with only isolated tumour cells, ITC, are positive), ductal \vs\ lobular TCGA-BRCA (913) and adenocarcinoma \vs\ squamous TCGA-LUNG (1{,}053)~\cite{tcga2013pancancer}, benign, atypical and malignant BRACS~\cite{brancati2022bracs} (545), and RCC staging (921 TCGA kidney slides, pT1 and pT2 \vs\ pT3 and pT4, as in~\cite{pathctm}). RCC staging is site-confounded, so its differences are counted in our tallies and means.
Three held-out tasks took no part in building the selection, the models or the fast budget, and their splits, endpoints and analysis code were fixed in writing before their models were trained. They are PANDA~\cite{bulten2022panda} cancer \vs\ benign and ISUP grade (10{,}615 biopsies) and TCGA-KIRC nuclear grade (510 slides of the RCC staging cohort with a new label).

\textbf{Protocol.}
Design tasks use five patient-level $80/10/10\%$ splits $\times$ five seeds (25 runs, with BRACS on its official split $\times$ 5 seeds and CAMELYON16 on its official test set). Held-out tasks use five splits and one seed per encoder. Every method reads the same tiles and splits (Sec.~\ref{sec:setting}).
The frozen models are seven aggregators trained on full bags, namely gated and plain attention~\cite{ilse2018attention,lu2021clam}, DSMIL~\cite{li2021dsmil}, DTFD-MIL~\cite{zhang2022dtfd}, TransMIL~\cite{shao2021transmil}, max and mean pooling. 


\textbf{Statistics.}
AUROC (macro one-vs-rest beyond two classes) is the mean over splits of the seed means. Comparisons are paired by run and test slide~\cite{obuchowski1997clustered,nadeau2003inference}.  Holm is also applied to the 42-pair superiority family. Cost and wall-clock time are defined with Table~\ref{tab:main}.

\section{Results}\label{sec:results}



\paragraph{One selector serves every aggregator.} As shown in Fig.~\ref{fig:frozen}, each panel is one design task and shows six frozen aggregators, each with the same weights under four reads, namely its full bag, the fast budget $D$, as many random $20\times$ tiles as $D$ reads, and the budget chosen by the performance-first rule. Three situations can be read from it. First, on the two lymph-node tasks $D$ stays near the full bag while random tiles fall far below. Second, on BRCA and LUNG $D$ and random tiles lie within $0.005$ of each other. Third, the performance-first rule returns to the full bag on every task. Under the rule all 36 pairs of aggregator and task are within $0.004$ of the full bag, at a mean cost of $14$ to $100\%$ of the full bag's, depending on aggregator and task (Supplementary Table S14). The largest gaps at $D$ are those of TransMIL, $-0.035$ on CAMELYON17 and $-0.034$ on BRACS.

\begin{table*}[t]\centering\footnotesize\setlength\tabcolsep{4.5pt}
\caption {Held-out tasks and transfer. Test AUROC of frozen gated ABMIL models read through our selection at $D$ and at $(1000,10)$, minus their own full bag and minus as many random tiles, with cost as \% of the full bag's encoder evaluations. Under the performance-first rule each model's budget is chosen on its own training and validation slides (on SLN-Breast the budget chosen on CAMELYON17), registered before these reads except for the two transfers between CAMELYON16 and CAMELYON17. $^{*}$In the lower block, the $95\%$ interval excludes zero.  On PANDA no slide has more than $500$ coarse regions, so both budgets read the same tiles. C16 and C17 are CAMELYON16 and CAMELYON17. KIRC grade reuses the RCC staging slides with a new label. Intervals and details in Supplementary Tables S24, S26, S28 and S31.}
\begin{tabular}{@{}lrccrccrc@{}}\toprule
 & \multicolumn{3}{c}{$D=(500,10)$} & \multicolumn{3}{c}{$(1000,10)$} & \multicolumn{2}{c}{performance-first rule} \\ \cmidrule(lr){2-4}\cmidrule(lr){5-7}\cmidrule(l){8-9}
Frozen gated ABMIL & cost & minus full bag & minus random & cost & minus full bag & minus random & cost & minus full bag \\ \midrule
\multicolumn{9}{@{}l}{\emph{Held-out tasks}} \\
\quad PANDA cancer & 37\% & $-0.001$ & $+0.011$ & 37\% & $-0.001$ & $+0.011$ & 46\% & $0.000$ \\
\quad PANDA ISUP & 37\% & $-0.003$ & $+0.011$ & 37\% & $-0.003$ & $+0.010$ & 37\% & $-0.003$ \\
\quad KIRC grade & 4.1\% & $+0.022$ & $+0.026$ & 6.4\% & $+0.020$ & $+0.025$ & 60\% & $+0.007$ \\
\multicolumn{9}{@{}l}{\emph{Across centres and datasets}} \\
\quad C16 selector, C17 models & 4.2\% & $-0.025^{*}$ & $+0.146^{*}$ & 6.5\% & $-0.004$ & $+0.167^{*}$ & 91\% & $+0.001$ \\
\quad C17 selector, C16 models & 4.4\% & $-0.006$ & $+0.170^{*}$ & 6.6\% & $+0.008$ & $+0.184^{*}$ & 13\% & $-0.017$ \\
\quad SLN-Breast, C17 models & 5.9\% & $-0.007$ & $+0.052^{*}$ & 8.0\% & $-0.003$ & $+0.059^{*}$ & 87\% & $0.000$ \\
\quad C17 centre left out & 4.7\% & $+0.003$ & $+0.152^{*}$ & 6.9\% & $+0.013$ & $+0.168^{*}$ & 82\% & $0.000$ \\
\bottomrule\end{tabular}
\label{tab:general}\end{table*}

\paragraph{An existing model keeps its accuracy at a few percent of its encoder cost.}
As Table ~\ref{tab:main} shown, at $(1000,10)$ the frozen model reads $6.3$ to $6.8\%$ of the encoder cost, $10.0\times$ to $14.3\times$ faster than the full bag for gated ABMIL (Table~\ref{tab:main}), and the six frozen aggregators are on average within $0.005$ AUROC of their full bags in 33 of 36 pairs (Supplementary Table S23). Reading every coarse region without the gate and zooming $10$ by the fitted scorer is within $0.005$ in 32 of 36 pairs, so at $K=1000$ the gate removes little. The gate matters at $D$. In an exploratory count at a validation threshold of $95\%$ specificity, $(1000,10)$ misses no more CAMELYON17 slides with only isolated tumour cells (ITC) than the full bag for any aggregator, while $D$ misses more (Supplementary section K). The fast budget $D=(500,10)$ reads $4.1$ to $4.7\%$ of the FLOPs, $13.3\times$ to $21.8\times$ faster than the full bag with tiling excluded ($10.0$ to $17.1\times$ with it), and is within $0.005$ in 22 of 36 pairs and within $0.010$ in 29 pairs, against 14 for as many random tiles, but it pays on the smallest lesions (Supplementary Table S30). On CAMELYON17 five aggregators lose $0.012$ to $0.035$ at $D$, and every efficient method in Table~\ref{tab:main} that reads less than the full bag at a fixed budget loses on CAMELYON17, by $0.013$ to $0.090$.


\paragraph{Efficient methods rarely beat the model they would replace.}
Paired by split and seed against ten settings of nine efficient methods (ZoomMIL, HAG-MIL, PathCTM, GFR-MIL, DiagMIL, SASHA at $10$ and $20\%$, HDMIL, IPS and ADMIL) over the six tasks, the frozen gated ABMIL is ahead in 55 of 60 comparisons at $D$, but its own full bag is ahead in 59 of the same paired comparisons as well, so the reading keeps the full-bag model's lead. It is behind as point estimates only on BRCA (Supplementary Table S15) and against HAG-MIL on CAMELYON17 ($-0.003$). The efficient methods ran with their authors' code at published settings without per-task tuning, while our methods were also run without tuning.

\paragraph{Choosing a budget with the performance-first rule.} A performance-first rule walks down the budget ladder from the full bag on each model's own training and validation slides and keeps the cheapest budget within a margin of at most $0.005$. Over six aggregators it is never more than $0.005$ below the full bag on average, the largest losses $0.004$ (Supplementary Table S14). It controls AUROC, not sensitivity by lesion size. On the held-out tasks, SLN-Breast and the CAMELYON17 centre left out, gated ABMIL under the rule is $-0.003$ to $+0.007$ from its full bag at $37$ to $87\%$ of the cost (Table~\ref{tab:general}). The two transfers between CAMELYON16 and CAMELYON17, not registered, are $+0.001$ and $-0.017$ (Supplementary section H).

\begin{figure}[t]\centering\includegraphics[width=\linewidth]{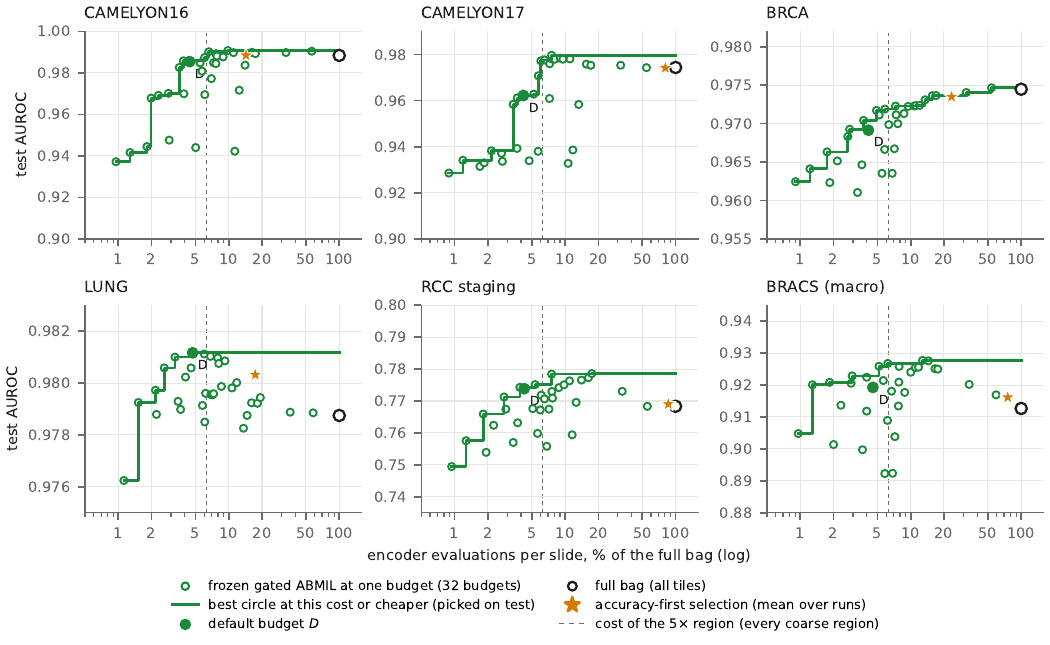}
\caption{Budget curves of a frozen model (Virchow2, test AUROC, mean over 25 runs, BRACS 5). Green line: the highest AUROC that any budget at most this expensive reaches. It is chosen with the test results, so it is an optimistic upper bound.  Cost: encoder evaluations, thumbnail included, \% of the full bag.}
\label{fig:budget}
\end{figure}

\paragraph{Held-out tasks.}  As Table~\ref{tab:general} shown,  On PANDA the frozen gated ABMIL at $D$ is $0.001$ below its full bag for cancer detection and $0.003$ below for ISUP grade, non-inferior at $0.010$ on both, and $0.011$ above as many random tiles on each (5 runs, Supplementary Table S31). Biopsies have few regions, so $D$ reads about $37\%$ of the encoder evaluations there, against about $4\%$ on resections. On KIRC grade it is $0.022$ above its full bag, which we do not read as a gain from selection, since the task reuses the RCC staging slides with a new label. The performance-first rule is within $0.003$ of the full bag on PANDA and $0.007$ above it on KIRC grade, at $37$ to $60\%$ of the cost. DSMIL and TransMIL on the held-out tasks are in Supplementary Table S31.

\paragraph{Across centres and cohorts.}The selector retains its advantage over random tiles under transfer (Table~\ref{tab:general}). Transferring from one lymph-node cohort to the other yields gains of $0.146$ and $0.170$ over random selection at budget $D$. Relative to the full bag, performance drops by $0.006$ on CAMELYON16 and $0.025$ on CAMELYON17. At $(1000,10)$, the differences are $+0.008$ and $-0.004$. On SLN-Breast, an external sentinel-node cohort from a different hospital and scanner, CAMELYON17 models are $0.007$ below the full bag at $D$ and $0.003$ below at $(1000,10)$. Leaving one CAMELYON17 centre out of training yields differences of $+0.003$ at $D$ and $+0.013$ at $(1000,10)$. The performance-first rule matches the full bag on SLN-Breast and the held-out centre, but retains $87\%$ and $82\%$ of the full computational cost. Under cohort transfer, it matches the full bag on CAMELYON17 but loses $0.017$ on CAMELYON16 despite selecting a budget of only $13\%$ of the cost based on validation slides (Supplementary section H).

\paragraph{Budget curve.} Figure ~\ref{fig:budget} shows the frozen gated ABMIL at every budget of the rule's ladder. On every task a budget of at most 6\% of the slide comes within 0.005 of the full bag. These points are picked on the test results and are optimistic. The fast budget $D$, fixed in advance, is within 0.007 of the full bag on five tasks and above it on LUNG, RCC staging and BRACS, but 0.012 below it on CAMELYON17. The performance-first rule, chosen without the test slides, stays close to the full bag on every task at 14 to 86\% of the cost, so the gap between the star and the line is what choosing a budget without the test results costs.

\begin{figure*}[t]
\centering
\includegraphics[width=\textwidth]{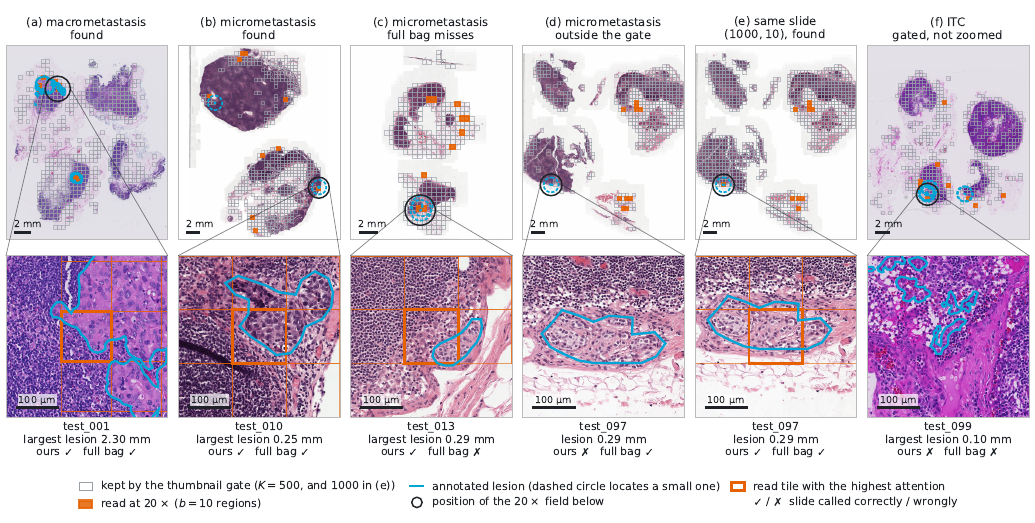}
\caption{
\textbf{What the selection reads and what it misses on CAMELYON16 test slides} (selector and frozen gated ABMIL of split 0, Virchow2). Each column is one read of one slide, with the outcome in its title. \emph{Top}, the slide, with the regions kept by the thumbnail gate in grey, the ten regions read at $20\times$ in orange and the annotated lesions in cyan. A dashed circle locates a lesion too small to see at this scale. \emph{Bottom}, the $20\times$ field at the black ring, $0.375$\,mm wide. If the read calls the slide positive, the field is centred on the tile with the highest attention among the tiles read, drawn as a thick orange square, with the other tiles read in thin orange. If the read misses the slide, the field is centred on the largest lesion, and the absence of orange squares shows that the lesion was not read. The tick or cross under each field says whether our read and the full bag call the slide correctly at $95\%$ specificity on the validation negatives. All reads are at $D=(500,10)$ except (e), which repeats the slide of (d) at $(1000,10)$. The slides are chosen by rule, as (a) the first slide by id with a macrometastasis, (b) the smallest micrometastasis that is zoomed and called, (c) the only positive slide that our read calls and the full bag misses, and (d) and (f) the only two annotated positive slides with no lesion among the zoomed regions.}\label{fig:qual}
\end{figure*}

\paragraph{What the selection reads.} Among the 48 positive test slides with annotations, the ten regions read at $D$ contain an annotated lesion on 46, and the frozen model calls all 46 (Fig.~\ref{fig:qual}). Over all test slides it calls 47 of 49 positive and 3 of 79 negative slides at its validation threshold, against 48 and 2 at $(1000,10)$ and 47 and 7 for the full bag.

Inside the selection the frozen model finds the lesion, whether it fills the bag or only a few tiles. On test\_010 (b) only 5 of the 159 selected tiles overlap the $0.25$\,mm lesion, yet they receive $98\%$ of the attention. On test\_001 (a) 146 of 160 selected tiles overlap a lesion and receive almost all of it. A smaller bag can also help the model. On test\_013 (c) 4 of the 146 selected tiles overlap the two lesions, each at most $0.29$\,mm, and receive $65\%$ of the attention. The tumour probability is $0.16$, well above the threshold of $0.0026$. On the full bag of the same slide it is $8.9\times10^{-6}$, just below the full bag's threshold of $1.2\times10^{-5}$.

The two misses at $D$ have different causes. On test\_097 (d) the gate ranks the region containing the $0.29$\,mm lesion outside its top 500 of 876 regions, so no later stage can recover it, and the full bag calls the slide. At $K=1000$ the gate keeps every region of this slide, the coarse scorer ranks the lesion among its ten, and 5 of the 155 selected tiles receive $99.7\%$ of the attention (e). On test\_099 (f) the gate keeps 12 of the 14 lesions, all isolated tumour cells of at most $0.10$\,mm, but the coarse scorer ranks none of them among its ten. The full bag misses this slide as well. No negative slide is shown. On the negative slide with the highest score at $D$, test\_088, a single tile receives $52\%$ of the attention and the full bag is above its threshold too, so this false positive comes from the frozen model and not from the selection. These slides are one model on one split and serve as an illustration.

\section{Conclusion}

A selector with $3{,}328$ weights, fitted to a frozen model's attention with no new labels, lets an existing MIL model read about $6.6\%$ of its encoder cost at $(1000,10)$ and stay within $0.005$ AUROC of its full bag in 33 of 36 aggregator-task pairs, without retraining. At the fast budget $D$, about $4\%$, the count is 22 of 36 and CAMELYON17 loses up to $0.035$, which the performance-first rule recovers only at $64$ to $94\%$ of the cost. At $D$ the model is still ahead of nine efficient methods in 55 of 60 paired comparisons, so we suggest that an efficient WSI method report the existing model on selected and on random tiles at matched encoder cost. Fitting takes 4 to 26 minutes per split, so a laboratory can check its own model before training a new one.


\section*{LLM Usage Statement}
We used Claude Code throughout this work to assist with coding, experiments, and writing. Under our direction, it wrote and edited scripts for data handling, evaluation, statistical analysis, job scheduling, and table generation. It also collected experiment outputs. It did not generate or modify any data, labels, predictions, or images. All reported results were computed from saved model outputs. It further assisted with drafting the paper, searching and summarising related work, and refining the study's framing.

{
    \small
    \bibliographystyle{ieeenat_fullname}
    \bibliography{main}

@String(CVPR= {IEEE Conf. Comput. Vis. Pattern Recog.})

@String(ECCV= {Eur. Conf. Comput. Vis.})

@String(ICLR = {Int. Conf. Learn. Represent.})

@String(IJCAI = {IJCAI})

@String(AAAI = {AAAI})

@String(CVPR  = {CVPR})

@String(ECCV  = {ECCV})

@String(ICLR  = {ICLR})

@InProceedings{ilse2018attention,
  title = 	 {Attention-based Deep Multiple Instance Learning},
  author =       {Ilse, Maximilian and Tomczak, Jakub and Welling, Max},
  booktitle = 	 {Proceedings of the 35th International Conference on Machine Learning},
  pages = 	 {2127--2136},
  year = 	 {2018},
  editor = 	 {Dy, Jennifer and Krause, Andreas},
  volume = 	 {80},
  series = 	 {Proceedings of Machine Learning Research},
  month = 	 {10--15 Jul},
  publisher =    {PMLR},
  url = 	 {https://proceedings.mlr.press/v80/ilse18a.html}
}

@article{lu2021clam, title={Data-efficient and weakly supervised computational pathology on whole-slide images}, author={Lu, Ming Y. and Williamson, Drew F. K. and Chen, Tiffany Y. and Chen, Richard J. and Barbieri, Matteo and Mahmood, Faisal}, journal={Nature Biomedical Engineering}, volume={5}, pages={555--570}, year={2021}, publisher={Nature Publishing Group}, doi={10.1038/s41551-020-00682-w}}

@inproceedings{li2021dsmil, title={Dual-stream multiple instance learning network for whole slide image classification with self-supervised contrastive learning}, author={Li, Bin and Li, Yin and Eliceiri, Kevin W.}, booktitle={Proceedings of the IEEE/CVF Conference on Computer Vision and Pattern Recognition}, pages={14318--14328}, year={2021}}

@inproceedings{shao2021transmil, title={{Transmil}: Transformer based correlated multiple instance learning for whole slide image classification}, author={Shao, Zhuchen and Bian, Hao and Chen, Yang and Wang, Yifeng and Zhang, Jian and Ji, Xiangyang and others}, journal={Advances in Neural Information Processing Systems}, booktitle={NeurIPS}, volume={34}, pages={2136--2147}, year={2021}}

@inproceedings{zhang2022dtfd, title={{DTFD-MIL}: Double-Tier Feature Distillation Multiple Instance Learning for Histopathology Whole Slide Image Classification}, author={Hongrun Zhang and Yanda Meng and Yitian Zhao and Yihong Qiao and Xiaoyun Yang and Sarah E. Coupland and Yalin Zheng}, booktitle={CVPR}, year={2022}}

@article{chen2024uni, title={Towards a general-purpose foundation model for computational pathology}, author={Chen, Richard J. and Ding, Tong and Lu, Ming Y. and Williamson, Drew F. K. and others}, journal={Nature Medicine}, volume={30}, pages={850--862}, year={2024}}

@article{zimmermann2024virchow2, title={Virchow2: Scaling self-supervised mixed magnification models in pathology}, author={Zimmermann, Eric and Vorontsov, Eugene and Viret, Julian and others}, journal={arXiv:2408.00738}, year={2024}}

@article{oquab2024dinov2, title={{DINOv2}: Learning robust visual features without supervision}, author={Oquab, Maxime and Darcet, Timoth{\'e}e and Moutakanni, Th{\'e}o and others}, journal={TMLR}, year={2024}}

@inproceedings{thandiackal2022zoommil, title={Differentiable zooming for multiple instance learning on whole-slide images}, author={Thandiackal, Kevin and Chen, Boqi and Pati, Pushpak and Jaume, Guillaume and Williamson, Drew F. K. and Gabrani, Maria and Goksel, Orcun}, booktitle={ECCV}, year={2022}}

@inproceedings{xiong2023hagmil, title={Diagnose like a pathologist: Transformer-enabled hierarchical attention-guided multiple instance learning for whole slide image classification}, author={Xiong, Conghao and Chen, Hao and Sung, Joseph J. Y. and King, Irwin}, booktitle={IJCAI}, year={2023}}

@inproceedings{bergner2023ips, title={Iterative patch selection for high-resolution image recognition}, author={Bergner, Benjamin and Lippert, Christoph and Mahendran, Aravindh}, booktitle={ICLR}, year={2023}}

@inproceedings{dong2025hdmil, title={Fast and accurate gigapixel pathological image classification with hierarchical distillation multi-instance learning}, author={Dong, Jiuyang and Jiang, Junjun and Jiang, Kui and Li, Jiahan and Zhang, Yongbing}, booktitle={CVPR}, year={2025}}

@misc{gfrmil, author="Qi, Mingxin
and Li, Hongyu
and Lu, Yinuo
and Mu, Wei",
editor="Yang, Guang
and Adeli, Ehsan
and de Bruijne, Marleen
and Papie{\.{z}}, Bart{\l}omiej W.
and Speidel, Stefanie
and Tiwari, Pallavi
and Zheng, Guoyan
and Yaqub, Mohammad
and Dou, Qi
and Rekik, Islem",
title="{GFR-MIL}: Glance--Focus--Reflect Based Multiple Instance Learning for Multi-Scale Whole Slide Image Analysis",
booktitle="Medical Image Computing and Computer Assisted Intervention -- MICCAI 2026",
year="2027",
publisher="Springer Nature Switzerland",
address="Cham",
pages="171--181",
isbn="978-3-032-38224-5"}

@inproceedings{diagmil, author="Jeong, Jiwon
and Hong, Sungrae
and Han, Donghee
and Shin, Jisu
and Kim, Kyungeun
and Yi, Mun Yong",
editor="Yang, Guang
and Adeli, Ehsan
and de Bruijne, Marleen
and Papie{\.{z}}, Bart{\l}omiej W.
and Speidel, Stefanie
and Tiwari, Pallavi
and Zheng, Guoyan
and Yaqub, Mohammad
and Dou, Qi
and Rekik, Islem",
title="DiagMIL: Uncertainty-Gated Coarse-to-Fine Multi-resolution MIL for WSI Diagnosis",
booktitle="Medical Image Computing and Computer Assisted Intervention -- MICCAI 2026",
year="2027",
publisher="Springer Nature Switzerland",
address="Cham",
pages="572--582",
isbn="978-3-032-38224-5"}

@inproceedings{pathctm, title={Thinking in scales: Accelerating gigapixel pathology image analysis via adaptive continuous reasoning}, author={Ge, J. and Zhan, Y. and Zhao, W. and Zhang, D. and Wang, K. and Liu, J. and Yang, C. and Li, C. and Zhang, J. and Dong, Y. and Zhang, N. and Liu, Q. and Crispin-Ortuzar, M. and Fu, H. and Li, C. and Gao, Z.}, booktitle={ICML}, year={2026}, eprint={2605.19491}}

@inproceedings{zhao2023rlogist, title={{RLogist}: Fast observation strategy on whole-slide images with deep reinforcement learning}, author={Zhao, Boxuan and Zhang, Jun and Ye, Deheng and Cao, Jian and Han, Xiao and Fu, Qiang and Yang, Wei}, booktitle={AAAI}, year={2023}}

@inproceedings{sasha2025, title={Sequential attention-based sampling for histopathological analysis}, author={Gogisetty, Tarun and Malpani, Naman and Thoppe, Gugan and Devarajan, Sridharan}, booktitle={NeurIPS}, year={2025}, eprint={2507.05077}}

@incollection{maurer1995multiple, title={Multiple comparisons in drug clinical trials and preclinical assays: a-priori ordered hypotheses}, author={Maurer, Willi and Hothorn, Ludwig A. and Lehmacher, Walter}, booktitle={Biometrie in der chemisch-pharmazeutischen Industrie}, editor={Vollmar, J.}, volume={6}, pages={3--18}, publisher={Gustav Fischer}, address={Stuttgart}, year={1995}}

@article{stothers2026admil, title={{ADMIL}: Attention-distilled multiple instance learning for selective foundation model inference in pathology}, author={Stothers, Duncan and Wu, Ren-Chin and Lotter, William}, journal={arXiv:2608.22066}, year={2026}}

@article{angelopoulos2021ltt, title={Learn then test: Calibrating predictive algorithms to achieve risk control}, author={Angelopoulos, Anastasios N. and Bates, Stephen and Cand{\`e}s, Emmanuel J. and Jordan, Michael I. and Lei, Lihua}, journal={arXiv:2110.01052}, year={2021}}

@article{bejnordi2017camelyon16, title={Diagnostic assessment of deep learning algorithms for detection of lymph node metastases in women with breast cancer}, author={Ehteshami Bejnordi, Babak and others}, journal={JAMA}, volume={318}, number={22}, pages={2199--2210}, year={2017}}

@article{litjens2018camelyon, title={1399 {H\&E}-stained sentinel lymph node sections of breast cancer patients: the {CAMELYON} dataset}, author={Litjens, Geert and Bandi, Peter and Ehteshami Bejnordi, Babak and Geessink, Oscar and Balkenhol, Maschenka and Bult, Peter and Halilovic, Altuna and Hermsen, Meyke and van de Loo, Rob and Vogels, Rob and Manson, Quirine F. and Stathonikos, Nikolas and Baidoshvili, Alexi and van Diest, Paul and Wauters, Carla and van Dijk, Marcory and van der Laak, Jeroen}, journal={GigaScience}, volume={7}, number={6}, pages={giy065}, year={2018}}

@article{brancati2022bracs, title={{BRACS}: A dataset for breast carcinoma subtyping in {H\&E} histology images}, author={Brancati, Nadia and Anniciello, Anna Maria and Pati, Pushpak and Riccio, Daniel and Scognamiglio, Giosu{\`e} and Jaume, Guillaume and De Pietro, Giuseppe and Di Bonito, Maurizio and Foncubierta, Antonio and Botti, Gerardo and Gabrani, Maria and Feroce, Florinda and Frucci, Maria}, journal={Database}, volume={2022}, pages={baac093}, year={2022}}

@article{tcga2013pancancer, title={The {Cancer Genome Atlas} pan-cancer analysis project}, author={{The Cancer Genome Atlas Research Network} and Weinstein, John N. and Collisson, Eric A. and Mills, Gordon B. and Shaw, Kenna R. Mills and Ozenberger, Brad A. and Ellrott, Kyle and Shmulevich, Ilya and Sander, Chris and Stuart, Joshua M.}, journal={Nature Genetics}, volume={45}, number={10}, pages={1113--1120}, year={2013}}

@article{filiot2025h0mini, title={Distilling foundation models for robust and efficient models in digital pathology}, author={Filiot, Alexandre and Dop, Nicolas and Tchita, Oussama and Riou, Auriane and Dubois, R{\'e}my and Peeters, Thomas and others}, journal={arXiv:2501.16239}, year={2025}}

@article{nadeau2003inference, title={Inference for the generalization error}, author={Nadeau, Claude and Bengio, Yoshua}, journal={Machine Learning}, volume={52}, pages={239--281}, year={2003}}

@article{obuchowski1997clustered, title={Nonparametric analysis of clustered {ROC} curve data}, author={Obuchowski, Nancy A.}, journal={Biometrics}, volume={53}, pages={567--578}, year={1997}}

@article{bulten2022panda, title={Artificial intelligence for diagnosis and {Gleason} grading of prostate cancer: the {PANDA} challenge}, author={Bulten, Wouter and Kartasalo, Kimmo and Chen, Po-Hsuan Cameron and Str{\"o}m, Peter and Pinckaers, Hans and Nagpal, Kunal and Cai, Yuannan and Steiner, David F. and van Boven, Hester and Vink, Robert and others}, journal={Nature Medicine}, volume={28}, number={1}, pages={154--163}, year={2022}}

@article{shin2025wisefuse, title={{WISE-FUSE}: Efficient whole slide image encoding via coarse-to-fine patch selection with {VLM} and {LLM} knowledge fusion}, author={Shin, Yonghan and Kim, SeungKyu and Jeong, Won-Ki}, journal={arXiv:2508.14537}, year={2025}}

@misc{mahmood2025uni2h, title={{UNI2-h}}, author={{Mahmood Lab}}, howpublished={Hugging Face model card, MahmoodLab/UNI2-h}, year={2025}}

@misc{yang2026gated,
      title={Gated Spatial Redundancy Projection for Pathology Transformer Attentions}, 
      author={Zhiyuan Yang and Jiahao Cheng and Vincent Quoc-Huy Trinh and Mahdi S. Hosseini},
      year={2026},
      eprint={2608.08374},
      archivePrefix={arXiv},
      primaryClass={cs.CV},
      url={https://arxiv.org/abs/2608.08374}, 
}
}


\clearpage
\setcounter{page}{1}
\onecolumn
{\centering\Large\textbf{\thetitle}\\\vspace{0.5em}Supplementary Material\\\vspace{1.0em}\par}
\label{supp:start}
\renewcommand{\thetable}{S\arabic{table}}\renewcommand{\thefigure}{S\arabic{figure}}\renewcommand{\theequation}{S\arabic{equation}}\renewcommand{\thealgorithm}{S\arabic{algorithm}}
\setcounter{table}{0}\setcounter{figure}{0}\setcounter{equation}{0}\setcounter{algorithm}{0}
\appendix
\section{FLOPs and wall-clock time of every method}\label{app:costall}
This section puts the cost of every method in one place. It covers the full bag, each efficient baseline, and our selection at its three operating points, the fast budget $D=(500,10)$, the budget $(1000,10)$, and the performance-first rule, which chooses a budget for each model from its training and validation slides. All numbers are for Virchow2 and the frozen gated ABMIL, per test slide, and are the ones behind Table~\ref{tab:main} of the main paper. Table~\ref{tab:costflops} counts computation, Table~\ref{tab:costtime} gives measured time, and Table~\ref{tab:costours} gives the intervals for our three operating points.
\paragraph{FLOPs.} Every encoder evaluation of a coarse or a fine tile counts $340$\,GFLOPs (Virchow2), and the networks of each baseline are measured with the same FLOP counter (Sec.~\ref{app:impl}) and added. Our selection is counted in encoder evaluations as in our budget tables, the tiles it encodes plus the thumbnail pass at $5.76$ evaluations ($1.96$\,TFLOPs over $340$\,GFLOPs). The full bag encodes $12.2$ to $15.4$ thousand tiles per slide, which is $4.1$ to $5.2$\,PFLOPs. Our selection at $D$ needs $0.19$ to $0.22$\,PFLOPs and at $(1000,10)$ $0.28$ to $0.34$\,PFLOPs. The rule's cost depends on the budget it chooses, from $14\%$ to $86\%$ of the full bag, and it is expensive on the tasks where it often keeps the full bag (Table~\ref{tab:costours}). HAG-MIL and DiagMIL encode every $5\times$ region on top of every $20\times$ tile under it, so they cost more than the full bag.
\begin{table}[H]\centering\scriptsize\setlength\tabcolsep{4pt}
\caption{FLOPs per test slide in PFLOPs ($10^{15}$) / as a percentage of the full bag (F\%), mean over the test slides. F\% is the cost column of Table~\ref{tab:main} of the main paper. Lower is cheaper.}\label{tab:costflops}
\adjustbox{max width=\linewidth}{\begin{tabular}{@{}lcccccc@{}}\toprule
Method & CAMELYON16 & CAMELYON17 & BRCA & LUNG & RCC staging & BRACS \\ \midrule
full bag (gated ABMIL) & 4.94\,/\,100 & 5.24\,/\,100 & 5.24\,/\,100 & 4.14\,/\,100 & 5.04\,/\,100 & 4.92\,/\,100 \\
ours, fast budget $D=(500,10)$ & 0.22\,/\,4.4 & 0.22\,/\,4.2 & 0.21\,/\,4.1 & 0.19\,/\,4.7 & 0.21\,/\,4.3 & 0.22\,/\,4.5 \\
ours, $(1000,10)$ & 0.32\,/\,6.6 & 0.34\,/\,6.6 & 0.33\,/\,6.3 & 0.28\,/\,6.8 & 0.32\,/\,6.5 & 0.33\,/\,6.7 \\
ours, performance-first rule & 0.70\,/\,14 & 4.20\,/\,81 & 1.21\,/\,23 & 0.70\,/\,17 & 4.33\,/\,86 & 3.68\,/\,76 \\ \midrule
ZoomMIL & 0.38\,/\,7.7 & 0.41\,/\,7.7 & 0.40\,/\,7.7 & 0.33\,/\,8.0 & 0.37\,/\,7.3 & 0.38\,/\,7.8 \\
ZoomMIL, 3 levels$^{a}$ & 0.17\,/\,3.5 & 0.18\,/\,3.4 & n/a & 0.15\,/\,3.6 & 0.14\,/\,2.8 & 0.17\,/\,3.4 \\
PathCTM & 0.37\,/\,7.6 & 0.40\,/\,7.6 & 0.39\,/\,7.5 & 0.32\,/\,7.8 & 0.37\,/\,7.4 & 0.37\,/\,7.6 \\
ADMIL$^{b}$ & 0.055\,/\,1.1 & 0.12\,/\,2.4 & 0.056\,/\,1.1 & 0.011\,/\,0.3 & 0.033\,/\,0.7 & 0.011\,/\,0.2 \\
SASHA 10\% & 0.82\,/\,17 & 0.87\,/\,17 & 0.86\,/\,16 & 0.69\,/\,17 & 0.77\,/\,15 & 0.81\,/\,17 \\
SASHA 20\% & 1.30\,/\,26 & 1.40\,/\,27 & 1.35\,/\,26 & 1.10\,/\,27 & 1.28\,/\,25 & 1.30\,/\,26 \\
GFR-MIL & 1.66\,/\,34 & 1.77\,/\,34 & 1.79\,/\,34 & 1.42\,/\,34 & 1.72\,/\,34 & 1.68\,/\,34 \\
HDMIL & 3.18\,/\,64 & 3.31\,/\,63 & 4.45\,/\,85 & 3.36\,/\,81 & 3.68\,/\,73 & 4.64\,/\,94 \\
IPS & 4.89\,/\,99 & 5.18\,/\,99 & 5.20\,/\,99 & 4.11\,/\,99 & 5.00\,/\,99 & 4.88\,/\,99 \\
HAG-MIL$^{c}$ & 5.21\,/\,106 & 5.53\,/\,106 & 5.54\,/\,106 & 4.38\,/\,106 & 5.33\,/\,106 & 5.20\,/\,106 \\
DiagMIL & 5.21\,/\,106 & 5.53\,/\,106 & 5.54\,/\,106 & 4.38\,/\,106 & 5.33\,/\,106 & 5.20\,/\,106 \\ \midrule
ours without the gate, $(N,10)$ & 0.38\,/\,7.8 & 0.42\,/\,8.1 & 0.39\,/\,7.6 & 0.33\,/\,7.9 & 0.38\,/\,7.7 & 0.38\,/\,7.7 \\
random $20\times$ tiles at the FLOPs of $D$ & 0.22\,/\,4.4 & 0.22\,/\,4.2 & 0.21\,/\,4.1 & 0.19\,/\,4.7 & 0.21\,/\,4.3 & 0.22\,/\,4.5 \\
\bottomrule\end{tabular}}
\par\vspace{2pt}\parbox{\linewidth}{\scriptsize\raggedright $^{a}$Not run on BRCA. $^{b}$At the number of tiles chosen on validation, and its scorer still reads and decodes every tile. $^{c}$An upper bound on RCC staging, where five test slides have more than 2500 regions and HAG-MIL zooms into 2500 of them. Our percentages are taken against the $20\times$ tiles under the coarse grid, which are $98.9$ to $99.3\%$ of the full bag, and they average the ratio over runs, so they can differ from the ratio of the two PFLOPs entries by up to one point.}
\end{table}
\paragraph{Wall-clock time.} Time is measured end to end on the hardware of Sec.~\ref{app:impl}, on 20 test slides per task, the same slides for every method. It includes reading and decoding every tile a method reads, encoding, every other step the method needs, and its own networks. Tissue segmentation and tiling are outside the timer, and Table~\ref{tab:tiling} adds each method's share of them (Sec.~\ref{app:tiling}). The speed-up is the summed seconds of the full bag divided by the summed seconds of the method. Over the six tasks our selection is $17.7\times$ faster than the full bag at $D$, $12.4\times$ at $(1000,10)$ and $2.5\times$ under the performance-first rule (geometric means). At $D$ the measured speed-up is below the ratio of FLOPs on every task, and most on LUNG and RCC staging ($13.3\times$ against $21\times$ on LUNG), since reading the thumbnail and decoding the $5\times$ tiles take time that FLOPs do not count.
\begin{table}[H]\centering\scriptsize\setlength\tabcolsep{4pt}
\caption{End-to-end seconds per test slide (median over 20 slides) / speed-up over the full bag (ratio of summed seconds). The last column is the geometric mean of the speed-up over the tasks. Higher speed-up is faster.}\label{tab:costtime}
\adjustbox{max width=\linewidth}{\begin{tabular}{@{}lccccccc@{}}\toprule
Method & CAMELYON16 & CAMELYON17 & BRCA & LUNG & RCC staging & BRACS & geo.\ mean \\ \midrule
full bag (gated ABMIL) & 61.2\,s\,/\,1.0$\times$ & 64.3\,s\,/\,1.0$\times$ & 73.5\,s\,/\,1.0$\times$ & 45.1\,s\,/\,1.0$\times$ & 35.1\,s\,/\,1.0$\times$ & 56.5\,s\,/\,1.0$\times$ & 1.0$\times$ \\
ours, fast budget $D=(500,10)$ & 3.0\,s\,/\,21.1$\times$ & 3.0\,s\,/\,21.8$\times$ & 3.6\,s\,/\,21.2$\times$ & 3.5\,s\,/\,13.3$\times$ & 3.5\,s\,/\,13.3$\times$ & 3.6\,s\,/\,17.8$\times$ & 17.7$\times$ \\
ours, $(1000,10)^{c}$ & 4.9\,s\,/\,14.2$\times$ & 5.2\,s\,/\,13.9$\times$ & 5.8\,s\,/\,14.3$\times$ & 5.3\,s\,/\,10.0$\times$ & 4.0\,s\,/\,10.2$\times$ & 5.2\,s\,/\,12.5$\times$ & 12.4$\times$ \\
ours, performance-first rule$^{d}$ & 9.0\,s\,/\,6.9$\times$ & 54.1\,s\,/\,1.2$\times$ & 16.2\,s\,/\,4.4$\times$ & 9.7\,s\,/\,4.8$\times$ & 38.5\,s\,/\,1.1$\times$ & 46.9\,s\,/\,1.3$\times$ & 2.5$\times$ \\ \midrule
ZoomMIL & 4.9\,s\,/\,13.0$\times$ & 5.2\,s\,/\,12.5$\times$ & 6.2\,s\,/\,12.1$\times$ & 4.5\,s\,/\,10.9$\times$ & 3.4\,s\,/\,11.0$\times$ & 4.9\,s\,/\,11.8$\times$ & 11.8$\times$ \\
ZoomMIL, 3 levels$^{a}$ & 2.2\,s\,/\,28.0$\times$ & 2.4\,s\,/\,27.2$\times$ & n/a & 2.2\,s\,/\,20.8$\times$ & 1.8\,s\,/\,24.3$\times$ & 2.3\,s\,/\,27.2$\times$ & 25.3$\times$ \\
PathCTM & 4.9\,s\,/\,12.5$\times$ & 5.1\,s\,/\,12.5$\times$ & 6.1\,s\,/\,12.2$\times$ & 4.6\,s\,/\,10.2$\times$ & 3.7\,s\,/\,10.3$\times$ & 5.0\,s\,/\,11.6$\times$ & 11.5$\times$ \\
ADMIL$^{b}$ & 4.5\,s\,/\,13.5$\times$ & 7.3\,s\,/\,8.2$\times$ & 14.2\,s\,/\,5.3$\times$ & 11.4\,s\,/\,3.6$\times$ & 9.9\,s\,/\,3.6$\times$ & 12.5\,s\,/\,4.6$\times$ & 5.7$\times$ \\
SASHA 10\% & 11.7\,s\,/\,5.4$\times$ & 13.2\,s\,/\,5.0$\times$ & 15.5\,s\,/\,4.7$\times$ & 11.1\,s\,/\,4.3$\times$ & 8.4\,s\,/\,4.3$\times$ & 12.0\,s\,/\,4.6$\times$ & 4.7$\times$ \\
SASHA 20\% & 19.2\,s\,/\,3.3$\times$ & 22.5\,s\,/\,3.0$\times$ & 25.8\,s\,/\,2.8$\times$ & 17.6\,s\,/\,2.6$\times$ & 13.7\,s\,/\,2.6$\times$ & 21.1\,s\,/\,2.7$\times$ & 2.8$\times$ \\
GFR-MIL & 21.1\,s\,/\,2.9$\times$ & 21.1\,s\,/\,3.1$\times$ & 27.3\,s\,/\,2.8$\times$ & 16.8\,s\,/\,2.7$\times$ & 13.6\,s\,/\,2.6$\times$ & 20.7\,s\,/\,2.7$\times$ & 2.8$\times$ \\
HDMIL & 39.7\,s\,/\,1.5$\times$ & 40.6\,s\,/\,1.6$\times$ & 62.7\,s\,/\,1.2$\times$ & 36.9\,s\,/\,1.2$\times$ & 26.0\,s\,/\,1.4$\times$ & 53.1\,s\,/\,1.1$\times$ & 1.3$\times$ \\
IPS & 60.8\,s\,/\,1.0$\times$ & 63.1\,s\,/\,1.0$\times$ & 72.8\,s\,/\,1.0$\times$ & 44.4\,s\,/\,1.0$\times$ & 35.0\,s\,/\,1.0$\times$ & 56.2\,s\,/\,1.0$\times$ & 1.0$\times$ \\
HAG-MIL & 65.0\,s\,/\,0.9$\times$ & 68.0\,s\,/\,0.9$\times$ & 78.7\,s\,/\,0.9$\times$ & 48.7\,s\,/\,0.9$\times$ & 37.8\,s\,/\,0.9$\times$ & 60.0\,s\,/\,0.9$\times$ & 0.9$\times$ \\
DiagMIL & 68.7\,s\,/\,0.9$\times$ & 72.7\,s\,/\,0.9$\times$ & 84.2\,s\,/\,0.9$\times$ & 51.5\,s\,/\,0.9$\times$ & 39.5\,s\,/\,0.9$\times$ & 63.8\,s\,/\,0.9$\times$ & 0.9$\times$ \\ \midrule
ours without the gate, $(N,10)$ & 4.8\,s\,/\,12.9$\times$ & 5.2\,s\,/\,12.2$\times$ & 6.1\,s\,/\,12.3$\times$ & 4.5\,s\,/\,10.6$\times$ & 3.4\,s\,/\,10.6$\times$ & 4.9\,s\,/\,11.9$\times$ & 11.7$\times$ \\
random $20\times$ tiles at the FLOPs of $D$ & 2.9\,s\,/\,22.3$\times$ & 2.9\,s\,/\,22.7$\times$ & 3.0\,s\,/\,25.3$\times$ & 2.9\,s\,/\,18.3$\times$ & 2.9\,s\,/\,17.1$\times$ & 3.0\,s\,/\,21.3$\times$ & 21.0$\times$ \\
\bottomrule\end{tabular}}
\par\vspace{2pt}\parbox{\linewidth}{\scriptsize\raggedright $^{a}$Not run on BRCA, mean over five tasks. $^{b}$Timed at $K=128$, and its time is dominated by scoring every tile. $^{c}$Timed in a separate session with the same harness, slides and GPU, against the full bag of that session, whose summed seconds are within $1\%$ of the full bag of this table on every task. $^{d}$Seconds are the mean over runs and slides, since the rule's budget differs between runs. Every budget it chose was timed in the session of the other rows.}
\end{table}
\begin{table}[H]\centering\scriptsize\setlength\tabcolsep{3pt}
\caption{Our three operating points, measured speed-up over the full bag with a slide-bootstrap $95\%$ interval. The last column is the share of runs in which the rule keeps the full bag.}\label{tab:costours}
\begin{tabular}{@{}lcccc@{}}\toprule
Task & $D$ & $(1000,10)$ & rule & rule keeps full bag \\ \midrule
CAMELYON16 & 21.1 [17.6, 24.7] & 14.2 [12.7, 15.8] & 6.86 [5.82, 7.87] & 0\% \\
CAMELYON17 & 21.8 [18.1, 25.5] & 13.9 [12.3, 15.7] & 1.23 [1.18, 1.27] & 60\% \\
BRCA & 21.2 [17.1, 25.0] & 14.3 [12.3, 16.0] & 4.39 [3.60, 5.09] & 0\% \\
LUNG & 13.3 [10.1, 16.8] & 10.0 [8.2, 11.9] & 4.81 [3.91, 5.64] & 4\% \\
RCC staging & 13.3 [10.0, 16.4] & 10.2 [8.5, 11.7] & 1.14 [1.09, 1.17] & 72\% \\
BRACS & 17.8 [14.3, 21.5] & 12.5 [11.0, 14.1] & 1.33 [1.22, 1.43] & 40\% \\
\midrule geometric mean & 17.7 & 12.4 & 2.54 & \\
\bottomrule\end{tabular}\end{table}

\paragraph{Measured time of our selection.} Table~\ref{tab:timelf} times the existing gated ABMIL read through our selection at $D$ and on its full bag, in the session of every method of Table~\ref{tab:main} on its 20 test slides per task (random $20\times$ tiles at $D$'s FLOPs, timed in the same session, are in Table~\ref{tab:costtime}), with no other process on the GPU. Seconds include each method's own networks. The speed-up, the ratio of these summed seconds, is in Table~\ref{tab:costours} with its slide-bootstrap $95\%$ interval.
\begin{table}[H]\centering\scriptsize\setlength\tabcolsep{3pt}\begin{tabular}{@{}lccc@{}}\toprule Task & full bag, summed (s) & ours $D$, summed (s) & ours $D$, median (s) \\ \midrule
CAMELYON16 & 1234 & 58.5 & 3.05 \\
CAMELYON17 & 1327 & 60.9 & 3.03 \\
BRCA & 1424 & 67.3 & 3.56 \\
LUNG & 935 & 70.2 & 3.52 \\
RCC staging & 876 & 66.1 & 3.52 \\
BRACS & 1248 & 70.1 & 3.58 \\
\bottomrule\end{tabular}\caption{Measured time of our selection at $D$ and of the full bag over the 20 timed slides per task. The speed-ups (ratios of the summed seconds) and their intervals are in Table~\ref{tab:costours}, and the random reader in Table~\ref{tab:costtime}.}\label{tab:timelf}\end{table}
\paragraph{Where $D$'s time goes.} Table~\ref{tab:stages} gives the mean seconds per slide of each stage of $D$ on the 20 timed slides per task, against random $20\times$ tiles at the same FLOPs. On the TCGA slides the $5\times$ tiles are decoded from the $10\times$ pyramid level and downsampled, and on these and BRACS they cost more per tile than $20\times$ tiles, and the thumbnail pass costs more on the larger slides. FLOPs count neither cost.
\begin{table}[H]\centering\scriptsize\setlength\tabcolsep{3pt}\begin{tabular}{@{}lcccccc@{}}\toprule Task & thumbnail (s) & $5\times$ (s, ms per tile) & $20\times$ (s, ms per tile) & $D$ total (s) & random tiles (s, ms per tile) & $D$ / random \\ \midrule
CAMELYON16 & 0.17 & 2.06, 4.4 & 0.67, 4.4 & 2.90 & 2.77, 4.4 & 1.05 \\
CAMELYON17 & 0.15 & 2.22, 4.5 & 0.72, 4.6 & 3.09 & 2.95, 4.5 & 1.05 \\
BRCA & 0.38 & 2.30, 4.9 & 0.75, 4.7 & 3.43 & 2.87, 4.5 & 1.20 \\
LUNG & 0.59 & 2.14, 5.4 & 0.77, 4.9 & 3.50 & 2.56, 4.6 & 1.37 \\
RCC staging & 0.41 & 2.10, 5.3 & 0.77, 4.9 & 3.29 & 2.56, 4.6 & 1.29 \\
BRACS & 0.38 & 2.40, 5.0 & 0.75, 4.7 & 3.53 & 2.94, 4.5 & 1.20 \\
\bottomrule\end{tabular}\caption{Stage times of our selection at $D$ and the FLOPs-matched random reader, from an earlier session with the same harness and slides, whose totals differ from those of Table~\ref{tab:timelf} by at most $0.07$\,s per slide.}\label{tab:stages}\end{table}
\paragraph{Preparation cost and break-even.} Table~\ref{tab:prep} costs the coarse encodings and thumbnails of the training slides at the measured per-tile and per-slide rates of Table~\ref{tab:stages} (median regions per training slide times the $5\times$ time per tile, plus the thumbnail step), against the seconds a new slide saves at $D$ (full bag minus $D$, from a different earlier session with the same harness, so its seconds differ slightly from those of Table~\ref{tab:stages}). Fitting the scorers adds $4$ to $26$ minutes per split, including one pass of the frozen teacher over the training slides, and the threshold re-set reads the validation slides once at $D$.
\begin{table}[H]\centering\scriptsize\setlength\tabcolsep{3pt}\begin{tabular}{@{}lcccccc@{}}\toprule Task & training slides & regions per slide & preparation (GPU-min) & full bag (s) & $D$ (s) & break-even (slides) \\ \midrule
CAMELYON16 & 214 & 980 & 16 & 61.0 & 2.72 & 16 \\
CAMELYON17 & 400 & 963 & 29 & 66.5 & 2.95 & 28 \\
BRCA & 732 & 1000 & 63 & 71.5 & 3.35 & 56 \\
LUNG & 841 & 832 & 70 & 46.5 & 3.46 & 98 \\
RCC staging & 742 & 965 & 67 & 43.9 & 3.18 & 99 \\
BRACS & 394 & 854 & 30 & 62.4 & 3.51 & 30 \\
\bottomrule\end{tabular}\caption{One-off preparation of the selector per task (coarse encodings and thumbnails of the training slides of split 0, split 1 on CAMELYON17; fitting the scorers, $4$ to $26$ minutes per split, is not included) and the number of new slides after which it is repaid.}\label{tab:prep}\end{table}
\subsection{Tissue segmentation and tiling}\label{app:tiling}
Table~\ref{tab:tiling} is measured with the pipeline that produced our tiles (a SAM2 tissue mask on the thumbnail, then patch coordinates), batch one, on the 20 slides per task of the wall-clock study, each evicted from the page cache, on an uncontended GPU. Methods that need the whole $20\times$ grid (full bags, IPS, HDMIL, ADMIL) are charged the $20\times$ coordinates, coarse-first methods and ours the $5\times$ coordinates plus their share of $20\times$ coordinates (ours on CAMELYON17 the whole $20\times$ grid, from which its aligned $5\times$ grid is built).
\begin{table}[H]\centering\scriptsize\setlength\tabcolsep{3pt}\begin{tabular}{@{}lcccccc@{}}\toprule
Task & opening (s) & segmentation (s) & $20\times$ grid (s) & $5\times$ grid (s) & ours $D$ vs full bag, with tiling & vs ZoomMIL, with tiling \\ \midrule
CAMELYON16 & 0.02 & 0.80 & 0.23 & 0.02 & 17.1$\times$ & 1.50$\times$ \\
CAMELYON17 & 0.01 & 0.87 & 0.24 & 0.02 & 15.9$\times$ & 1.45$\times$ \\
BRCA & 0.01 & 0.60 & 0.38 & 0.03 & 15.9$\times$ & 1.55$\times$ \\
LUNG & 0.01 & 0.56 & 0.22 & 0.02 & 10.9$\times$ & 1.18$\times$ \\
RCC staging & 0.01 & 0.52 & 0.16 & 0.02 & 10.0$\times$ & 1.15$\times$ \\
BRACS & 0.01 & 0.57 & 0.36 & 0.03 & 13.5$\times$ & 1.38$\times$ \\
\bottomrule\end{tabular}\caption{Segmentation and tiling time (median per slide) and speed-ups with it counted (per slide, ours adds opening, segmentation, the $5\times$ grid and its share of the $20\times$ grid, a median $0.5$ to $1.4$\,s, as medians of per-slide sums, which need not equal the sum of the column medians), as ratios of summed seconds over the 20 timed slides (the statistic of Table~\ref{tab:main}). Without tiling the speed-ups are those of Table~\ref{tab:costtime}.}\label{tab:tiling}\end{table}

\subsection{Cost per slide}
The cost in the main paper is a ratio of cohort means. Table~\ref{tab:perslidecost} gives, per test slide (pooled over runs, Virchow2), the mean, 95th percentile and maximum of the share of the slide's full bag, and the share of slides that cost more than their full bag. Any reader with a fixed number of coarse regions or zoomed regions exceeds the full bag on a few very small slides. Reading the full bag whenever the budget would cost more (a fallback) leaves the means almost unchanged.
\begin{table}[H]\centering\scriptsize\setlength\tabcolsep{2.5pt}\adjustbox{max width=\linewidth}{\begin{tabular}{@{}l@{\hspace{6pt}}ccccc@{\hspace{6pt}}ccccc@{\hspace{6pt}}ccccc@{\hspace{6pt}}ccccc@{}}\toprule
 & \multicolumn{5}{c}{LUNG} & \multicolumn{5}{c}{BRACS} & \multicolumn{5}{c}{RCC staging} & \multicolumn{5}{c}{CAMELYON17} \\ \cmidrule(lr){2-6}\cmidrule(lr){7-11}\cmidrule(lr){12-16}\cmidrule(lr){17-21}
reader & ratio of means & mean & p95 & max & over 100\% & ratio of means & mean & p95 & max & over 100\% & ratio of means & mean & p95 & max & over 100\% & ratio of means & mean & p95 & max & over 100\% \\ \midrule
ours, $D$ & 4.7 & 8.6 & 19.0 & 115.9 & 0.8\% & 4.5 & 5.6 & 9.9 & 14.1 & 0.0\% & 4.3 & 5.9 & 12.9 & 28.1 & 0.0\% & 4.2 & 5.4 & 11.6 & 18.2 & 0.0\% \\
ZoomMIL & 8.0 & 10.8 & 19.8 & 109.7 & 0.8\% & 7.9 & 8.3 & 10.1 & 15.2 & 0.0\% & 7.4 & 7.9 & 10.7 & 26.3 & 0.0\% & 7.7 & 8.2 & 10.8 & 17.2 & 0.0\% \\
ZoomMIL, 3 levels & 3.6 & 6.9 & 17.5 & 112.9 & 0.9\% & 3.4 & 3.9 & 6.1 & 11.8 & 0.0\% & 2.8 & 3.6 & 7.5 & 28.0 & 0.0\% & 3.4 & 4.0 & 7.2 & 13.9 & 0.0\% \\
\bottomrule\end{tabular}}\caption{Cost per test slide, \% of that slide's full bag.}\label{tab:perslidecost}\end{table}
\subsection{Training and inference requirements}\label{sec:supp_req}
All methods are trained on the same Virchow2 features of the training slides, so the encoding of the training set is shared and excluded. The table lists what each method needs beyond that, and what it must read and compute for one new slide. Inference cost is in Virchow2-equivalent FLOPs as \% of the full bag, range over the six tasks (Table~\ref{tab:mainsix}).

\begin{table*}[h]\centering\scriptsize\setlength\tabcolsep{3pt}
\caption{\textbf{Requirements per method} (settings as run here). Other compute ``none'' means nothing beyond the networks listed under training. ``Trainings'' counts the models that must be trained to cover several operating points of the method's budget knob.}\label{tab:req}
\adjustbox{max width=\linewidth}{\begin{tabular}{@{}l l l l l r l@{}}\toprule
 & \multicolumn{3}{c}{Training} & \multicolumn{3}{c}{Inference on one new slide} \\ \cmidrule(lr){2-4}\cmidrule(l){5-7}
Method & features & other inputs, networks & trainings & pixels read & cost (\%) & other compute \\ \midrule
Full-bag models (7) & all $20\times$ & one MIL network & one & every $20\times$ tile & 100 & MIL head \\
ZoomMIL & all $5\times$, $20\times$ & differentiable top-$k$ MIL & one per $k$ & every $5\times$ region + $k$ regions at $20\times$ & 7.3 to 8.0 & none \\
PathCTM & all $5\times$, $20\times$ & recurrent zoom network & one per top-$n$ & every $5\times$ region + $n$ regions at $20\times$ & 7.4 to 7.8 & 40 recurrent steps \\
GFR-MIL & all $5\times$, $20\times$ & two-step selection & one per share & every $5\times$ region + ${\approx}28\%$ at $20\times$, two dependent steps & 34 & none \\
SASHA (0.1 / 0.2) & all $5\times$, $20\times$ & four stages incl.\ an RL agent & one per fraction & every $5\times$ region + 10 to 20\% at $20\times$, one region per step & 15 to 27 & agent, state update per step \\
DiagMIL (soft / hard) & all $5\times$, $20\times$ & per-slide graphs, two stages & one & every $5\times$ region + all or routed $20\times$ & 106 / 7 & graph construction (kNN, MST) \\
HAG-MIL & all $5\times$, $20\times$ & multi-stage & one per $t_\text{eval}$ & every $5\times$ region + most $20\times$ & 106 & none \\
HDMIL & all $20\times$ & \textbf{pixels} of every tile (low resolution), pixel CNN & one per keep ratio & low-resolution image of every tile + kept $20\times$ & 63 to 94 & CNN on every tile \\
IPS & all $20\times$ & iterative selection & one per memory size & every $20\times$ tile & 99 & iterative selection \\
ADMIL$^\dagger$ & all $20\times$ & \textbf{pixels} of $20\times$ tiles, EfficientNet-B0 & one student per $K$ & \textbf{every $20\times$ tile} (scored) + $K$ encoded & 0.2 to 2.4$^\ddagger$ & CNN on every $20\times$ tile \\
\textbf{Ours} & all $5\times$, $20\times$ & thumbnail, 2 linear scorers fitted to a frozen attention model, an existing full-bag model & \textbf{one for all 32 budgets} & \textbf{thumbnail + $K$ regions at $5\times$ + $b$ regions at $20\times$} & \textbf{4.1 to 4.7} & thumbnail encoder, linear scorers \\ \bottomrule
\end{tabular}}
{\scriptsize $^\dagger$Our re-implementation (no public code), and the magnification of its scorer is not stated in its paper; we use $20\times$. $^\ddagger$At the $K$ chosen on validation (per task as in Table~\ref{tab:mainsix}), encoder evaluations plus the scorer on every tile, ratio of cohort means, and every tile is still read and decoded. Ours costs 4.1 to 4.7\% at $D$. Official settings differ for some baselines (ZoomMIL $10\times\!\to\!20\times$ on CAMELYON16, DiagMIL $10\times/20\times$, GFR-MIL $5\times\!\to\!40\times$, HAG-MIL three levels, PathCTM four scales, HDMIL $1.25\times$), and we report the settings that produced every number in this paper.}
\end{table*}

Two points follow. In training, no method needs less than the full $20\times$ features, and the coarse-to-fine methods, ours included, also need all $5\times$ features. At inference, apart from its thumbnail, ours is the only method that reads neither every region at the coarse level nor every $20\times$ tile (on slides with at most $K$ regions it does read all of them at $5\times$), whereas the coarse-to-fine baselines read every $5\times$ region first, HDMIL a low-resolution image of every tile, and IPS and ADMIL every $20\times$ tile.

\subsection{Training time of every method}
Table~\ref{tab:traintime} gives the minutes each method took to train for one run, as logged by the runs of Table~\ref{tab:mainsix} (median over runs). These runs were trained in different sessions on shared GPUs, often several at a time, so only orders of magnitude are meaningful. For ours the deployed full-bag model already exists, so its cost is fitting the two scorers once per split, including one pass of the frozen teacher over the training slides, and one fit serves every budget. Each baseline trains its own aggregator, once for every setting of its budget knob, and the deployed model is replaced. ADMIL is the sum of its teacher, its PriorNet (one per split, trained on the pixels of the tiles) and one student, excluding the time to sample tiles and to score every tile with PriorNet.
\begin{table*}[h]\centering\scriptsize\setlength\tabcolsep{3pt}\adjustbox{max width=\linewidth}{\begin{tabular}{@{}lrrrrrrl@{}}\toprule
Method & CAMELYON16 & CAMELYON17 & BRCA & LUNG & RCC staging & BRACS & trainings to cover the budget \\ \midrule
\textbf{Ours}, fitting the selector & 4 & 9 & 25 & 14 & 26 & 4 & \textbf{one for all 32 budgets} \\
the existing full-bag gated ABMIL & 18 & 18 & 23 & 30 & 25 & 9 & one \\
the existing full-bag DSMIL & 12 & 21 & 38 & 39 & 40 & 19 & one \\
\midrule
ZoomMIL & 48 & 80 & 155 & 132 & 197 & 81 & one per $k$ \\
ZoomMIL, 3 levels & 30 & 53 & n/a & 101 & 103 & 123 & one per $k$ \\
HAG-MIL & 65 & 81 & 313 & 279 & 163 & 62 & one per $t_\text{eval}$ \\
PathCTM & 107 & 219 & 372 & 421 & 445 & 186 & one per top-$n$ \\
GFR-MIL & 38 & 46 & 90 & 82 & 82 & 37 & one per share \\
DiagMIL & 52 & 63 & 80 & 93 & 93 & 34 & one \\
SASHA, both fractions in one run & 661 (11\,h) & 1184 (20\,h) & 1672 (28\,h) & 1668 (28\,h) & 1529 (25\,h) & 897 (15\,h) & one run per pair of fractions \\
HDMIL & 112 & 172 & 319 & 555 & 669 (11\,h) & 1370 (23\,h) & one per keep ratio \\
IPS & 30 & 48 & 98 & 80 & 115 & 48 & one per memory size \\
ADMIL, teacher, PriorNet and one student & 164 & 289 & 624 (10\,h) & 831 (14\,h) & 552 & 253 & one student per $K$ \\
\bottomrule\end{tabular}}\caption{Training time per run in minutes (Virchow2, median over the runs, 25 per task and 5 on BRACS, SASHA 5, ours once per split), from the runs' own logs on shared GPUs, so indicative. Ours needs an existing full-bag model (the next two rows), which it does not retrain. n/a where no run was trained.}\label{tab:traintime}\end{table*}

\section{Deviations of the baselines from the published settings}\label{sec:supp_dev}
Every baseline uses the same Virchow2 (or UNI2-h) features, splits and seeds as ours and two levels ($5\times\!\to\!20\times$ for the multi-resolution methods, except the three-level ZoomMIL variant at $2.5\times\!\to\!5\times\!\to\!20\times$), which changes the input magnification of several methods (footnote of Table~\ref{tab:req}), and HDMIL's low-resolution input is at about $2.5\times$. Beyond that, the table lists every other difference from the authors' published setting that we found. We did not tune any baseline per task. In the main text, ``their authors' code at published settings'' means each method's released code and published hyperparameters apart from the shared features, the two-level input and the differences listed here. Two methods differ from this. ADMIL has no public code and is our re-implementation (Sec.~\ref{app:admil}), and GFR-MIL's training loop is re-implemented with the logic of its repository. ``Without per-task tuning'' means that no hyperparameter was tuned per task. ADMIL's $K$ is still chosen per task on validation slides, by the rule of Sec.~\ref{app:admil}. The budget setting of every baseline is in Table~\ref{tab:budgets}.

\begin{table*}[h]\centering\scriptsize\setlength\tabcolsep{3pt}
\caption{\textbf{Deviations from the published settings} (besides features and levels). Direction is the method a deviation is likely to disadvantage, where it can be judged. Uniform means it applies to all full-bag models alike, and none that we see no likely direction.}\label{tab:dev}
\adjustbox{max width=\linewidth}{\begin{tabular}{@{}l p{11.5cm} l@{}}\toprule
Method & Deviation & Direction \\ \midrule
Full-bag models (7) & One shared recipe for all, Adam, lr $10^{-4}$, weight decay $10^{-5}$, cosine schedule, dropout $0.25$, class-weighted cross-entropy, 100 epochs, early stopping on validation loss (patience 20). The ``gated ABMIL, raw features'' row is Ilse's gated attention on the raw features (no projection). ``gated ABMIL'' is CLAM-SB without its instance-clustering loss. DSMIL uses softmax cross-entropy for its bag and max-instance terms. DTFD-MIL is the AFS variant with 4 pseudo-bags and one optimiser (random pseudo-bags at evaluation change AUROC by $\le0.001$). TransMIL uses Adam instead of RAdam/Lookahead. & uniform \\
ZoomMIL & Two levels instead of three, 16 children per zoomed region (published 4 per level). & none \\
PathCTM & Two scales instead of four. The released code saves no checkpoint, and ours keeps the best validation AUROC with early stopping (patience 20). On CAMELYON16, 21 of 25 runs reach validation AUROC $1.0$ early, so the kept checkpoint is from passes 4 to 15 of 100. & against PathCTM (possible) \\
GFR-MIL & Epochs 100, patience 20, gradient accumulation 4 (repository defaults 50, 10, 1), and its training loop is re-implemented with the same logic. Training is not deterministic (fold 0, seed 0 gave $0.805$ and $0.624$ on a rerun). & none \\
HAG-MIL & Two levels, early stopping with \texttt{min\_delta} $10^{-3}$ (official $0$), $1750/2500$ regions at train/test, the same number of $20\times$ tokens as the official $7000/10000$ at $10\times$. The official checkpoint rule (lowest validation loss) keeps epochs 1 to 3 on BRACS while validation AUROC still rises. & against HAG-MIL on BRACS \\
IPS & Last epoch, as in the official protocol (UNI2-h RCC staging $0.741$, against $0.787$ at the best validation checkpoint). & against IPS \\
SASHA & The CAMELYON16 configuration is used on every task, including BRCA, LUNG and RCC staging, for which the repository has TCGA configurations (about ten settings differ), and the threshold $\tau$ is fixed at the repository default $0.95$, while the SASHA paper tunes it on validation ($0.90$ on CAMELYON16 and TCGA-NSCLC). & against SASHA \\
DiagMIL & Graphs on $\ell_2$-normalised features, with exact parent to child edges between levels instead of $k$NN ($k=4$), loss weights from the paper ($\lambda_H=0.05$, $\lambda_\text{align}=0.1$, the repository script swaps them). On CAMELYON16 3 of its 25 runs collapse, identified by validation AUROC below $0.80$ ($0.59$ to $0.77$) without reading test results. Without them its CAMELYON16 test AUROC is $0.938$ instead of $0.897$, still $0.047$ below ours at $D$. On the other tasks no run falls below this level, except on RCC staging, where every method is near $0.75$ and the cutoff does not mark a collapse. & none \\
HDMIL & Low-resolution input at about $2.5\times$ (published $1.25\times$, so more pixels). LUNG uses the CAMELYON16/BRCA learning rate and KAN degree ($3\cdot10^{-4}$, 12) instead of the README's NSCLC setting ($3\cdot10^{-5}$, 16). When the selector keeps no patch for a class, our adapter falls back to all patches for that class (the official forward does not need it). It fires on 7 to 26\% of test slides and adds 1 to 8 points to HDMIL's cost. & cost against HDMIL \\
ADMIL & Our re-implementation (no public code), with the scorer at $20\times$ (not stated in the paper), PriorNet trained on JPEG-compressed tiles and applied to raw tiles, $K$ chosen on validation (the smallest $K$ whose validation AUROC reaches ADMIL's own full-bag teacher, else the $K$ with the highest validation AUROC, with values equal at 4 decimals tied to the smallest $K$. This tie-break was fixed after the CAMELYON16 and BRACS test results had been seen). On CAMELYON16 it reaches $0.964$ at $K{=}128$, below the published $0.978$. & against ADMIL \\
\bottomrule
\end{tabular}}
\end{table*}

\section{ADMIL (re-implementation)}\label{app:admil}
ADMIL~\cite{stothers2026admil} has no public code. Our re-implementation (Table~\ref{tab:dev}) scores every $20\times$ tile with a small network trained on the pixels and trains one aggregator per $K$. $K$ is chosen on validation slides as the smallest $K$ whose mean validation AUROC reaches ADMIL's own full-bag teacher, else the $K$ with the highest validation AUROC (the smallest $K$ on ties at 4 decimals). Cost counts encoder evaluations of the $K$ tiles plus the scorer on every tile, as a ratio of cohort means (Virchow2).
\begin{table}[!htbp]\centering\scriptsize\begin{tabular}{@{}lcccccc@{}}\toprule
 & CAMELYON16 & CAMELYON17 & BRCA & LUNG & RCC staging & BRACS \\ \midrule
runs & 25 & 25 & 25 & 25 & 25 & 5 \\
$K$ & 128 & 330 & 128 & 4 & 64 & 1 \\
AUROC & 0.964 & 0.961 & 0.972 & 0.965 & 0.758 & 0.887 \\
cost (\%) & 1.11 & 2.37 & 1.06 & 0.26 & 0.66 & 0.23 \\
\bottomrule\end{tabular}\caption{ADMIL (re-implementation) at the $K$ chosen on validation slides, test AUROC and cost.}\label{tab:admil}\end{table}
\paragraph{ADMIL's frontier.} ADMIL trains one student per $K$ (tiles kept after screening every tile). Table~\ref{tab:admilfrontier} gives its test AUROC and cost (ratio of cohort means, Virchow2) at every $K$ trained, with the $K$ we report (chosen on validation slides) in bold.
\begin{table}[H]\centering\scriptsize\setlength\tabcolsep{2.5pt}\begin{tabular}{@{}lcccc@{}}\toprule
$K$ & LUNG & BRACS & RCC staging & CAMELYON17 \\ \midrule
1 & 0.955 (0.2) & \textbf{0.887 (0.2)} & 0.683 (0.2) & 0.892 (0.2) \\
2 & 0.954 (0.2) & 0.897 (0.2) & 0.709 (0.2) & 0.910 (0.2) \\
4 & \textbf{0.965 (0.3)} & 0.895 (0.3) & 0.725 (0.3) & 0.898 (0.3) \\
8 & 0.963 (0.3) & 0.895 (0.3) & 0.757 (0.3) & 0.906 (0.3) \\
16 & 0.966 (0.4) & 0.905 (0.3) & 0.747 (0.3) & 0.933 (0.3) \\
32 & 0.968 (0.5) & 0.884 (0.4) & 0.736 (0.4) & 0.932 (0.4) \\
64 & 0.972 (0.8) & 0.903 (0.7) & \textbf{0.758 (0.7)} & 0.919 (0.6) \\
128 & 0.969 (1.3) & 0.906 (1.1) & 0.772 (1.1) & 0.956 (1.1) \\
180 & 0.969 (1.7) & 0.908 (1.5) & 0.778 (1.4) & 0.958 (1.4) \\
330 & 0.975 (2.9) & 0.905 (2.5) & 0.775 (2.5) & \textbf{0.961 (2.4)} \\
660 & 0.978 (5.6) & 0.899 (4.8) & 0.779 (4.7) & 0.964 (4.5) \\
\bottomrule\end{tabular}\caption{ADMIL at every $K$, test AUROC (cost, \%).}\label{tab:admilfrontier}\end{table}

\section{Frozen full-bag models read through our selection}\label{app:frozen}
Table~\ref{tab:frozen} gives every number behind Fig.~\ref{fig:frozen} of the main paper, including mean pooling.
\begin{table*}[t]\centering\scriptsize\setlength\tabcolsep{2.4pt}
\caption{\textbf{A frozen full-bag model read through our selection}, no retraining and no new labels (Virchow2, test AUROC, mean over 25 runs, BRACS 5). The column full is the model on its full bag, $\Delta_{\mathrm{full}}$ is the model at $D$ (about $4\%$ of the encoder FLOPs) minus the full bag, and $\Delta_{\mathrm{rand}}$ is the model at $D$ minus random $20\times$ tiles matched to the selector's tile count, bold if better after Holm over the pairs shown (within-run bootstrap). The column rule is the budget the performance-first rule chose on each model's training and validation slides, minus the full bag, with its mean cost in \% of the full bag. Mean pooling, whose full bag fails on lymph nodes, is shown but outside the claims, and the rule was not run for it (n/a). RCC staging is site-confounded, so its differences are counted in the tallies but not read as evidence. TransMIL reads a bag as a sequence, here in our gate order, so its full bag also depends slightly on the ranking.}\label{tab:frozen}
\adjustbox{max width=\linewidth}{\begin{tabular}{@{}l@{\hspace{5pt}}cccc@{\hspace{5pt}}cccc@{\hspace{5pt}}cccc@{\hspace{5pt}}cccc@{\hspace{5pt}}cccc@{\hspace{5pt}}cccc@{}}\toprule
 & \multicolumn{4}{c}{CAMELYON16} & \multicolumn{4}{c}{CAMELYON17} & \multicolumn{4}{c}{BRCA} & \multicolumn{4}{c}{LUNG} & \multicolumn{4}{c}{RCC staging} & \multicolumn{4}{c}{BRACS} \\ \cmidrule(lr){2-5}\cmidrule(lr){6-9}\cmidrule(lr){10-13}\cmidrule(lr){14-17}\cmidrule(lr){18-21}\cmidrule(lr){22-25}
Aggregator & full & $\Delta_{\mathrm{full}}$ & $\Delta_{\mathrm{rand}}$ & rule (\%) & full & $\Delta_{\mathrm{full}}$ & $\Delta_{\mathrm{rand}}$ & rule (\%) & full & $\Delta_{\mathrm{full}}$ & $\Delta_{\mathrm{rand}}$ & rule (\%) & full & $\Delta_{\mathrm{full}}$ & $\Delta_{\mathrm{rand}}$ & rule (\%) & full & $\Delta_{\mathrm{full}}$ & $\Delta_{\mathrm{rand}}$ & rule (\%) & full & $\Delta_{\mathrm{full}}$ & $\Delta_{\mathrm{rand}}$ & rule (\%) \\ \midrule
gated ABMIL & 0.988 & $-$0.003 & \textbf{+0.185} & +0.000 (14) & 0.974 & $-$0.012 & \textbf{+0.161} & $-$0.000 (81) & 0.974 & $-$0.005 & $-$0.000 & $-$0.001 (23) & 0.979 & +0.002 & +0.003 & +0.002 (17) & 0.768 & +0.006 & +0.008 & +0.001 (86) & 0.913 & +0.007 & +0.022 & +0.003 (76) \\
ABMIL & 0.990 & $-$0.002 & \textbf{+0.158} & +0.001 (14) & 0.977 & $-$0.014 & \textbf{+0.166} & +0.000 (67) & 0.974 & $-$0.005 & +0.002 & $-$0.000 (24) & 0.977 & +0.002 & +0.005 & +0.001 (15) & 0.767 & +0.002 & +0.006 & $-$0.001 (87) & 0.914 & +0.012 & +0.029 & +0.002 (62) \\
DSMIL & 0.973 & +0.003 & \textbf{+0.185} & +0.001 (19) & 0.973 & $-$0.008 & \textbf{+0.191} & +0.000 (75) & 0.972 & $-$0.006 & +0.005 & $-$0.000 (54) & 0.978 & +0.002 & +0.005 & +0.001 (31) & 0.767 & +0.006 & +0.016 & +0.003 (80) & 0.902 & +0.007 & +0.043 & +0.001 (84) \\
DTFD-MIL & 0.988 & $-$0.004 & \textbf{+0.172} & $-$0.003 (14) & 0.973 & $-$0.012 & \textbf{+0.193} & $-$0.004 (64) & 0.976 & $-$0.005 & $-$0.000 & $-$0.002 (25) & 0.980 & +0.002 & +0.004 & +0.000 (15) & 0.765 & +0.011 & +0.014 & +0.002 (84) & 0.913 & $-$0.006 & +0.027 & +0.000 (100) \\
TransMIL & 0.985 & $-$0.010 & \textbf{+0.217} & $-$0.002 (18) & 0.975 & $-$0.035 & \textbf{+0.169} & +0.000 (94) & 0.966 & $-$0.000 & +0.003 & $-$0.002 (76) & 0.978 & +0.001 & +0.004 & $-$0.000 (45) & 0.759 & +0.002 & +0.033 & $-$0.004 (66) & 0.920 & $-$0.034 & +0.015 & +0.000 (100) \\
max & 0.990 & $-$0.012 & \textbf{+0.255} & $-$0.003 (14) & 0.976 & $-$0.013 & \textbf{+0.132} & +0.000 (77) & 0.979 & $-$0.007 & $-$0.001 & +0.000 (100) & 0.981 & +0.002 & +0.001 & +0.000 (84) & 0.786 & $-$0.001 & +0.013 & +0.000 (100) & 0.919 & $-$0.001 & +0.043 & $-$0.000 (92) \\
mean & 0.587 & +0.256 & \textbf{+0.255} & n/a & 0.785 & +0.113 & \textbf{+0.118} & n/a & 0.962 & +0.007 & +0.008 & n/a & 0.973 & +0.012 & \textbf{+0.013} & n/a & 0.772 & $-$0.003 & $-$0.001 & n/a & 0.813 & +0.016 & +0.018 & n/a \\
\bottomrule\end{tabular}}\end{table*}

\section{Accuracy and cost on all six tasks}\label{app:results}
All analyses in this section use saved predictions of the trained models. Nothing was retrained.
The main text (Table~\ref{tab:main}) shows five design tasks with Virchow2 (UNI2-h in Table~\ref{tab:mainuni}). Table~\ref{tab:mainsix} adds BRCA and the other full-bag aggregators. Three gated ABMIL models appear in these tables. ``Gated ABMIL, raw features'' is Ilse's gated attention on the raw features, ``gated ABMIL (baseline code)'' is CLAM-SB without its clustering loss, trained with the baselines' code, and the frozen gated ABMIL is the model of our pipeline, with an input projection, that our selection is read into.
\begin{table*}[t]\centering\scriptsize\setlength\tabcolsep{1.1pt}
\caption{\textbf{Accuracy and cost on all six design tasks} (Virchow2; Table~\ref{tab:main} of the main paper shows five of them). AUROC is the test AUROC (macro one-vs-rest on BRACS), mean over 25 runs (5 splits $\times$ 5 seeds, on BRACS the official split $\times$ 5 seeds). F\% is the FLOPs per test slide as a share of the full bag's, counting every encoder evaluation and every other network a method runs (ours with the thumbnail pass). $\times$ is the measured end-to-end speed-up over the full bag, from reading tiles to the prediction (ratio of summed times over 20 test slides, Table~\ref{tab:costtime}). $^\ddagger$Not timed, reads and encodes the same tiles as the timed full bag (gated ABMIL), shown as $1.0$. Frozen gated ABMIL is the full-bag model in our pipeline, read unchanged through our selection at $D$, every row timed in one session on the same slides and GPU. $^\S$GFR-MIL and DiagMIL lose most on the lymph-node tasks, and DiagMIL collapses in some runs (Table~\ref{tab:dev}), and DiagMIL's AUROC uses soft routing, which runs both stages on every slide. $^\dagger$SASHA at its two published budgets, 5 runs per task. $^\P$No public code, our re-implementation (Sec.~\ref{app:admil}), with $K$ chosen on validation slides per task (128, 330, 128, 4, 64, 1 in the order of the columns), timed at $K=128$ on every task, so its $\times$ is not at the $K$ of its AUROC and F\%. The row gated ABMIL, raw features is Ilse's gated attention on the raw features, and ABMIL elsewhere is the plain attention aggregator of our pipeline. The full-bag rows above are trained with the baselines' code and settings on raw features, so they are different models from the frozen aggregators of the same name in Table~\ref{tab:frozen} and the other tables, and their AUROC differs. Bold marks the best AUROC on the task. Mean is the test AUROC over the six tasks. Subscripts give the standard deviation over runs. BRCA at $(1000,10)$ and under the performance-first rule, with intervals against random tiles, is in Table~\ref{tab:brcaall}.}\label{tab:mainsix}
\adjustbox{max width=\linewidth}{\begin{tabular}{@{}l@{\hspace{4pt}}crr@{\hspace{4pt}}crr@{\hspace{4pt}}crr@{\hspace{4pt}}crr@{\hspace{4pt}}crr@{\hspace{4pt}}crr@{\hspace{4pt}}c@{}}
\toprule
 & \multicolumn{3}{c}{CAMELYON16} & \multicolumn{3}{c}{CAMELYON17} & \multicolumn{3}{c}{BRCA} & \multicolumn{3}{c}{LUNG} & \multicolumn{3}{c}{RCC staging} & \multicolumn{3}{c}{BRACS} & \\ \cmidrule(lr){2-4}\cmidrule(lr){5-7}\cmidrule(lr){8-10}\cmidrule(lr){11-13}\cmidrule(lr){14-16}\cmidrule(lr){17-19}
Method & AUROC & F\% & $\times$ & AUROC & F\% & $\times$ & AUROC & F\% & $\times$ & AUROC & F\% & $\times$ & AUROC & F\% & $\times$ & AUROC & F\% & $\times$ & Mean \\
\midrule
\multicolumn{20}{@{}l}{\emph{Full bag}} \\
\quad mean pooling$^\ddagger$ & 0.561$_{\pm.026}$ & 100 & 1.0 & 0.773$_{\pm.035}$ & 100 & 1.0 & 0.963$_{\pm.025}$ & 100 & 1.0 & 0.973$_{\pm.020}$ & 100 & 1.0 & 0.770$_{\pm.058}$ & 100 & 1.0 & 0.790$_{\pm.004}$ & 100 & 1.0 & 0.8051 \\
\quad max pooling$^\ddagger$ & 0.697$_{\pm.027}$ & 100 & 1.0 & 0.806$_{\pm.059}$ & 100 & 1.0 & 0.957$_{\pm.016}$ & 100 & 1.0 & 0.970$_{\pm.022}$ & 100 & 1.0 & \textbf{0.780}$_{\pm.043}$ & 100 & 1.0 & 0.835$_{\pm.008}$ & 100 & 1.0 & 0.8409 \\
\quad gated ABMIL, raw features~\cite{ilse2018attention}$^\ddagger$ & 0.978$_{\pm.010}$ & 100 & 1.0 & 0.975$_{\pm.024}$ & 100 & 1.0 & \textbf{0.978}$_{\pm.017}$ & 100 & 1.0 & 0.980$_{\pm.016}$ & 100 & 1.0 & 0.768$_{\pm.045}$ & 100 & 1.0 & \textbf{0.923}$_{\pm.012}$ & 100 & 1.0 & 0.9338 \\
\quad gated ABMIL (baseline code) & \textbf{0.990}$_{\pm.006}$ & 100 & 1.0 & 0.975$_{\pm.023}$ & 100 & 1.0 & 0.974$_{\pm.018}$ & 100 & 1.0 & 0.979$_{\pm.018}$ & 100 & 1.0 & 0.770$_{\pm.048}$ & 100 & 1.0 & 0.905$_{\pm.013}$ & 100 & 1.0 & 0.9322 \\
\quad DSMIL~\cite{li2021dsmil}$^\ddagger$ & 0.985$_{\pm.009}$ & 100 & 1.0 & 0.974$_{\pm.025}$ & 100 & 1.0 & 0.973$_{\pm.020}$ & 100 & 1.0 & 0.979$_{\pm.018}$ & 100 & 1.0 & 0.768$_{\pm.061}$ & 100 & 1.0 & 0.901$_{\pm.019}$ & 100 & 1.0 & 0.9298 \\
\quad TransMIL~\cite{shao2021transmil}$^\ddagger$ & 0.980$_{\pm.008}$ & 100 & 1.0 & 0.965$_{\pm.029}$ & 100 & 1.0 & 0.967$_{\pm.020}$ & 100 & 1.0 & 0.978$_{\pm.016}$ & 100 & 1.0 & 0.733$_{\pm.068}$ & 100 & 1.0 & 0.898$_{\pm.022}$ & 100 & 1.0 & 0.9199 \\
\quad DTFD-MIL~\cite{zhang2022dtfd}$^\ddagger$ & 0.988$_{\pm.006}$ & 100 & 1.0 & \textbf{0.979}$_{\pm.021}$ & 100 & 1.0 & 0.976$_{\pm.020}$ & 100 & 1.0 & \textbf{0.983}$_{\pm.015}$ & 100 & 1.0 & 0.773$_{\pm.053}$ & 100 & 1.0 & 0.915$_{\pm.006}$ & 100 & 1.0 & 0.9354 \\
\multicolumn{20}{@{}l}{\emph{Multi-resolution and tile screening}} \\
\quad ZoomMIL~\cite{thandiackal2022zoommil} & 0.975$_{\pm.007}$ & 7.7 & 13.0 & 0.923$_{\pm.037}$ & 7.7 & 12.5 & 0.968$_{\pm.012}$ & 7.7 & 12.1 & 0.970$_{\pm.030}$ & 8.0 & 10.9 & 0.762$_{\pm.052}$ & 7.3 & 11.0 & 0.904$_{\pm.028}$ & 7.8 & 11.8 & 0.9170 \\
\quad PathCTM~\cite{pathctm} & 0.975$_{\pm.013}$ & 7.6 & 12.5 & 0.927$_{\pm.046}$ & 7.6 & 12.5 & 0.965$_{\pm.026}$ & 7.5 & 12.2 & 0.970$_{\pm.031}$ & 7.8 & 10.2 & 0.745$_{\pm.055}$ & 7.4 & 10.3 & 0.887$_{\pm.018}$ & 7.6 & 11.6 & 0.9115 \\
\quad HDMIL~\cite{dong2025hdmil} & 0.975$_{\pm.035}$ & 64 & 1.5 & 0.955$_{\pm.036}$ & 63 & 1.6 & 0.966$_{\pm.019}$ & 85 & 1.2 & 0.974$_{\pm.021}$ & 81 & 1.2 & 0.759$_{\pm.054}$ & 73 & 1.4 & 0.870$_{\pm.029}$ & 94 & 1.1 & 0.9166 \\
\quad ADMIL~\cite{stothers2026admil}$^\P$ & 0.964$_{\pm.026}$ & 1.1 & 13.5 & 0.961$_{\pm.030}$ & 2.4 & 8.2 & 0.972$_{\pm.022}$ & 1.1 & 5.3 & 0.965$_{\pm.021}$ & 0.3 & 3.6 & 0.758$_{\pm.060}$ & 0.7 & 3.6 & 0.887$_{\pm.012}$ & 0.2 & 4.6 & 0.9178 \\
\quad HAG-MIL~\cite{xiong2023hagmil} & 0.979$_{\pm.008}$ & 106 & 0.9 & 0.965$_{\pm.030}$ & 106 & 0.9 & 0.959$_{\pm.024}$ & 106 & 0.9 & 0.968$_{\pm.024}$ & 106 & 0.9 & 0.746$_{\pm.043}$ & 106 & 0.9 & 0.868$_{\pm.033}$ & 106 & 0.9 & 0.9142 \\
\quad IPS~\cite{bergner2023ips} & 0.980$_{\pm.011}$ & 99 & 1.0 & 0.958$_{\pm.037}$ & 99 & 1.0 & 0.975$_{\pm.013}$ & 99 & 1.0 & 0.969$_{\pm.029}$ & 99 & 1.0 & 0.763$_{\pm.054}$ & 99 & 1.0 & 0.895$_{\pm.010}$ & 99 & 1.0 & 0.9234 \\
\quad GFR-MIL~\cite{gfrmil}$^\S$ & 0.883$_{\pm.063}$ & 34 & 2.9 & 0.904$_{\pm.047}$ & 34 & 3.1 & 0.964$_{\pm.017}$ & 34 & 2.8 & 0.972$_{\pm.022}$ & 34 & 2.7 & 0.758$_{\pm.058}$ & 34 & 2.6 & 0.880$_{\pm.045}$ & 34 & 2.7 & 0.8933 \\
\quad DiagMIL~\cite{diagmil}$^\S$ & 0.897$_{\pm.122}$ & 106 & 0.9 & 0.925$_{\pm.037}$ & 106 & 0.9 & 0.966$_{\pm.020}$ & 106 & 0.9 & 0.976$_{\pm.018}$ & 106 & 0.9 & 0.728$_{\pm.068}$ & 106 & 0.9 & 0.857$_{\pm.016}$ & 106 & 0.9 & 0.8915 \\
\quad SASHA~\cite{sasha2025}, 10\%$^\dagger$ & 0.952$_{\pm.019}$ & 17 & 5.4 & 0.945$_{\pm.040}$ & 17 & 5.0 & 0.973$_{\pm.015}$ & 16 & 4.7 & 0.976$_{\pm.016}$ & 17 & 4.3 & 0.761$_{\pm.075}$ & 15 & 4.3 & 0.873$_{\pm.014}$ & 17 & 4.6 & 0.9133 \\
\quad SASHA~\cite{sasha2025}, 20\%$^\dagger$ & 0.953$_{\pm.019}$ & 26 & 3.3 & 0.953$_{\pm.024}$ & 27 & 3.0 & 0.973$_{\pm.012}$ & 26 & 2.8 & 0.976$_{\pm.019}$ & 27 & 2.6 & 0.760$_{\pm.068}$ & 25 & 2.6 & 0.878$_{\pm.012}$ & 26 & 2.7 & 0.9157 \\
\midrule\multicolumn{20}{@{}l}{\emph{Frozen gated ABMIL in our pipeline}} \\
\quad full bag & 0.988$_{\pm.009}$ & 100 & 1.0 & 0.974$_{\pm.020}$ & 100 & 1.0 & 0.974$_{\pm.021}$ & 100 & 1.0 & 0.979$_{\pm.020}$ & 100 & 1.0 & 0.768$_{\pm.056}$ & 100 & 1.0 & 0.913$_{\pm.014}$ & 100 & 1.0 & 0.9328 \\
\quad our selection, $D$ & 0.985$_{\pm.011}$ & 4.4 & 21.1 & 0.962$_{\pm.033}$ & 4.2 & 21.8 & 0.969$_{\pm.016}$ & 4.1 & 21.2 & 0.981$_{\pm.016}$ & 4.7 & 13.3 & 0.774$_{\pm.039}$ & 4.3 & 13.3 & 0.919$_{\pm.012}$ & 4.5 & 17.8 & 0.9318 \\
\bottomrule
\end{tabular}}
\end{table*}

\begin{table*}[!htbp]\centering\scriptsize\setlength\tabcolsep{2.4pt}\renewcommand\arraystretch{0.85}
\caption{\textbf{Accuracy and cost with UNI2-h}, on five design tasks (LUNG, RCC staging, BRACS, CAMELYON16, CAMELYON17). AUROC is the test AUROC, mean over 25 runs (5 on BRACS, a superscript gives fewer). F\% and $\times$ as in Table~\ref{tab:main}, measured with Virchow2 and assumed equal for UNI2-h, also for the baselines whose selection depends on the features (same tiles, encoder of the same size), except ADMIL, whose F\% follows the $K$ it chose on the UNI2-h validation slides. Our selection uses the scorers fitted to the frozen UNI2-h gated ABMIL's own attention, no new labels, its speed-up the Virchow2 measurement of the same reads. Random tiles are as many $20\times$ tiles as our selection reads, drawn uniformly, mean over 5 draws, so they cost less than our selection (no coarse tiles, no thumbnail) and their F\% is not given. Table~\ref{tab:main} instead matches random tiles to the FLOPs of $D$. Bold marks the best per task. The two gated ABMIL full-bag rows are different trained models, the baseline's code on raw features, and the model in our pipeline (with an input projection) that our selection is read into. ADMIL uses seed $0$ and is not timed. n/a means not run or not timed. Other marks as in Table~\ref{tab:mainsix}, except that ADMIL's $K$ is chosen on the UNI2-h validation slides and ADMIL is not timed here. Mean is over the five tasks shown.}\label{tab:mainuni}
\adjustbox{max width=\linewidth}{\begin{tabular}{@{}l@{\hspace{4pt}}crr@{\hspace{4pt}}crr@{\hspace{4pt}}crr@{\hspace{4pt}}crr@{\hspace{4pt}}crr@{\hspace{4pt}}c@{}}
\toprule
 & \multicolumn{3}{c}{LUNG} & \multicolumn{3}{c}{RCC staging} & \multicolumn{3}{c}{BRACS} & \multicolumn{3}{c}{CAMELYON16} & \multicolumn{3}{c}{CAMELYON17} & \\ \cmidrule(lr){2-4}\cmidrule(lr){5-7}\cmidrule(lr){8-10}\cmidrule(lr){11-13}\cmidrule(lr){14-16}
Method & AUROC & F\% & $\times$ & AUROC & F\% & $\times$ & AUROC & F\% & $\times$ & AUROC & F\% & $\times$ & AUROC & F\% & $\times$ & Mean \\
\midrule
gated ABMIL (baseline code) & 0.978 & 100 & 1.0 & \textbf{0.791} & 100 & 1.0 & 0.899 & 100 & 1.0 & 0.992 & 100 & 1.0 & 0.976 & 100 & 1.0 & 0.9271 \\
DTFD-MIL~\cite{zhang2022dtfd}$^\ddagger$ & \textbf{0.982} & 100 & 1.0 & 0.783 & 100 & 1.0 & 0.900 & 100 & 1.0 & \textbf{0.995} & 100 & 1.0 & \textbf{0.977} & 100 & 1.0 & 0.9273 \\
\midrule
ZoomMIL~\cite{thandiackal2022zoommil} & 0.981 & 8.0 & 10.9 & 0.756 & 7.3 & 11.0 & \textbf{0.911} & 7.8 & 11.8 & 0.970 & 7.7 & 13.0 & 0.913 & 7.7 & 12.5 & 0.9059 \\
PathCTM~\cite{pathctm} & 0.976$^{(5)}$ & 7.8 & 10.2 & 0.753$^{(5)}$ & 7.4 & 10.3 & 0.892 & 7.6 & 11.6 & 0.960$^{(5)}$ & 7.6 & 12.5 & 0.924$^{(5)}$ & 7.6 & 12.5 & 0.9010 \\
HDMIL~\cite{dong2025hdmil} & 0.975 & 81 & 1.2 & 0.736 & 73 & 1.4 & 0.864 & 94 & 1.1 & 0.980 & 64 & 1.5 & 0.966 & 63 & 1.6 & 0.9042 \\
ADMIL~\cite{stothers2026admil}$^\P$ & 0.967$^{(5)}$ & 0.3 & n/a & 0.761$^{(5)}$ & 2.5 & n/a & 0.866 & 2.5 & n/a & 0.968 & 1.1 & n/a & 0.954$^{(5)}$ & 4.5 & n/a & 0.9032 \\
HAG-MIL~\cite{xiong2023hagmil} & 0.969 & 106 & 0.9 & 0.782 & 106 & 0.9 & 0.882 & 106 & 0.9 & 0.980 & 106 & 0.9 & 0.920 & 106 & 0.9 & 0.9068 \\
IPS~\cite{bergner2023ips} & 0.980 & 99 & 1.0 & 0.741 & 99 & 1.0 & 0.863 & 99 & 1.0 & 0.980 & 99 & 1.0 & 0.968 & 99 & 1.0 & 0.9063 \\
GFR-MIL~\cite{gfrmil}$^\S$ & 0.973 & 34 & 2.7 & 0.764 & 34 & 2.6 & 0.808 & 34 & 2.7 & 0.746 & 34 & 2.9 & 0.840 & 34 & 3.1 & 0.8263 \\
DiagMIL~\cite{diagmil}$^\S$ & 0.978 & 106 & 0.9 & 0.761 & 106 & 0.9 & 0.878 & 106 & 0.9 & 0.930 & 106 & 0.9 & 0.947 & 106 & 0.9 & 0.8987 \\
SASHA~\cite{sasha2025}, 10\%$^\dagger$ & 0.978 & 17 & 4.3 & 0.774 & 15 & 4.3 & 0.889 & 17 & 4.6 & 0.930 & 17 & 5.4 & 0.894 & 17 & 5.0 & 0.8929 \\
SASHA~\cite{sasha2025}, 20\%$^\dagger$ & 0.979 & 27 & 2.6 & 0.766 & 25 & 2.6 & 0.891 & 26 & 2.7 & 0.954 & 26 & 3.3 & 0.930 & 27 & 3.0 & 0.9037 \\
\midrule
gated ABMIL in our pipeline (frozen), full bag & 0.978 & 100 & 1.0 & 0.781 & 100 & 1.0 & 0.899 & 100 & 1.0 & 0.992 & 100 & 1.0 & \textbf{0.977} & 100 & 1.0 & 0.9253 \\
frozen gated ABMIL, our selection, $D$ & 0.981 & 4.7 & 13.3 & 0.757 & 4.3 & 13.3 & 0.896 & 4.5 & 17.8 & 0.982 & 4.4 & 21.1 & 0.966 & 4.2 & 21.8 & 0.9162 \\
frozen gated ABMIL, random tiles, same count & 0.976 & n/a & n/a & 0.781 & n/a & n/a & 0.874 & n/a & n/a & 0.794 & n/a & n/a & 0.802 & n/a & n/a & 0.8451 \\
\bottomrule
\end{tabular}}
\end{table*}

\section{Ablation of what the selector is fitted to}\label{app:ablation}
Our two scorers are fitted to one frozen attention model's attention (Alg.~\ref{alg:fit}). Here the same scorers, or the same reading, use a different target, and each is read by the same frozen full-bag models at $D$ (Virchow2, test AUROC, mean over 25 runs, BRACS 5). Differences are from the within-run patient bootstrap with the models fixed, 10{,}000 draws, $95\%$ interval. The alternatives are the following.
\begin{itemize}\setlength\itemsep{1pt}
\item \emph{Fitted to slide labels.} The same attention $a_r$, optimiser, epochs and standardisation (Sec.~\ref{app:impl}), but each scorer adds a logistic output $\hat y=\operatorname{sigm}(w^\top\sum_r a_r v_r+w_0)$ and is trained on the binary cross-entropy between $\hat y$ and a slide target in $[0,1]$, the label, or for more than two classes a scalar code fixed in advance ($0,\tfrac12,1$ for three classes). The gate's logit gradient adds $\lambda_{\mathrm g}\theta_r(a^{\mathrm{th}}_r-a^{\mathrm c}_r)$ with $\lambda_{\mathrm g}=1$ and $\theta_r=a^{\mathrm c}_r/\max_{r'}a^{\mathrm c}_{r'}$, which pulls its attention toward the coarse scorer's. This needs the slide labels of the training slides.
\item \emph{Fitted to each aggregator's own importance.} Our KL fit with the target taken from the reading model itself instead of the gated ABMIL, ABMIL its attention, DSMIL its bag attention, DTFD-MIL its first-tier attention, TransMIL, mean and max pooling the absolute first-order contribution $|\sum_d g_d h_d|$ of each tile to the predicted logit (CAMELYON16 and CAMELYON17, registered before its results).
\item \emph{The model's own coarse attention, behind the label-fitted gate.} The zoom order inside the gate's top $K$ is the frozen model's own attention on its projected coarse tokens (gated ABMIL, ABMIL, DSMIL bag attention, DTFD-MIL first-tier attention). This reader keeps the label-fitted gate, so it differs from ours in both stages.
\item \emph{Random tiles} of the same number as our selection reads, and the model's \emph{full bag}.
\end{itemize}
\begin{table}[H]\centering\scriptsize\setlength\tabcolsep{2.5pt}\adjustbox{max width=\linewidth}{\begin{tabular}{@{}lccccccc@{}}\toprule
Task & ours & to labels & own coarse attention & random & full bag & ours $-$ to labels & ours $-$ own coarse attention \\ \midrule
CAMELYON16 & 0.985 & 0.989 & 0.976 & 0.800 & 0.988 & $-0.003$ [$-0.017$, $+0.012$] & $+0.010$ [$-0.008$, $+0.034$] \\
CAMELYON17 & 0.962 & 0.944 & 0.941 & 0.801 & 0.974 & $+0.018$ [$+0.002$, $+0.044$] & $+0.021$ [$0.000$, $+0.058$] \\
BRCA & 0.969 & 0.973 & 0.970 & 0.969 & 0.974 & $-0.003$ [$-0.012$, $+0.004$] & $-0.001$ [$-0.008$, $+0.006$] \\
LUNG & 0.981 & 0.980 & 0.979 & 0.978 & 0.979 & $+0.001$ [$-0.003$, $+0.005$] & $+0.002$ [$-0.001$, $+0.005$] \\
RCC staging & 0.774 & 0.776 & 0.760 & 0.765 & 0.768 & $-0.002$ [$-0.036$, $+0.033$] & $+0.014$ [$-0.014$, $+0.041$] \\
BRACS & 0.919 & 0.919 & 0.911 & 0.897 & 0.913 & $0.000$ [$-0.020$, $+0.017$] & $+0.008$ [$-0.006$, $+0.022$] \\
\bottomrule\end{tabular}}\caption{What the selector is fitted to, frozen gated ABMIL at $D$. For gated ABMIL our target is its own importance. Brackets give the $95\%$ interval.}\label{tab:ablgated}\end{table}
\begin{table}[H]\centering\scriptsize\setlength\tabcolsep{2.5pt}\adjustbox{max width=\linewidth}{\begin{tabular}{@{}llccccccc@{}}\toprule
Task & Frozen model & ours & to labels & own importance & own coarse attention & random & full bag & ours $-$ own importance \\ \midrule
CAMELYON16 & ABMIL & 0.988 & 0.989 & 0.982 & 0.980 & 0.830 & 0.990 & $+0.006$ [$0.000$, $+0.014$] \\
CAMELYON16 & DSMIL & 0.975 & 0.980 & 0.968 & 0.931 & 0.791 & 0.973 & $+0.007$ [$0.000$, $+0.018$] \\
CAMELYON16 & DTFD-MIL & 0.984 & 0.986 & 0.980 & 0.980 & 0.811 & 0.988 & $+0.004$ [$0.000$, $+0.009$] \\
CAMELYON16 & TransMIL & 0.976 & 0.978 & 0.661 & n/a & 0.758 & 0.985 & $+0.315$ [$+0.269$, $+0.361$] \\
CAMELYON16 & max pooling & 0.978 & 0.978 & 0.962 & n/a & 0.724 & 0.990 & $+0.016$ [$0.000$, $+0.036$] \\
CAMELYON16 & mean pooling & 0.843 & 0.868 & 0.759 & n/a & 0.588 & 0.587 & $+0.084$ [$+0.039$, $+0.135$] \\
\midrule
CAMELYON17 & ABMIL & 0.963 & 0.935 & 0.957 & 0.937 & 0.796 & 0.977 & $+0.006$ [$0.000$, $+0.016$] \\
CAMELYON17 & DSMIL & 0.965 & 0.938 & 0.957 & 0.907 & 0.775 & 0.973 & $+0.008$ [$-0.004$, $+0.026$] \\
CAMELYON17 & DTFD-MIL & 0.961 & 0.932 & 0.969 & 0.933 & 0.767 & 0.973 & $-0.008$ [$-0.027$, $+0.001$] \\
CAMELYON17 & TransMIL & 0.940 & 0.915 & 0.714 & n/a & 0.771 & 0.975 & $+0.226$ [$+0.168$, $+0.288$] \\
CAMELYON17 & max pooling & 0.962 & 0.942 & 0.959 & n/a & 0.831 & 0.976 & $+0.003$ [$-0.016$, $+0.038$] \\
CAMELYON17 & mean pooling & 0.898 & 0.886 & 0.848 & n/a & 0.780 & 0.785 & $+0.050$ [$+0.014$, $+0.103$] \\
\bottomrule\end{tabular}}\caption{What the selector is fitted to, the other frozen aggregators on the lymph-node tasks at $D$. Own coarse attention exists for aggregators with attention only. Brackets give the $95\%$ interval.}\label{tab:ablagg}\end{table}
\begin{table}[H]\centering\scriptsize\setlength\tabcolsep{2pt}\adjustbox{max width=\linewidth}{\begin{tabular}{@{}lcccccc@{}}\toprule Frozen model & CAMELYON16 & CAMELYON17 & BRCA & LUNG & RCC staging & BRACS \\ \midrule
ABMIL & $-0.001$ [$-0.013$, $+0.012$] & $+0.028$ [$+0.005$, $+0.060$] & $-0.004$ [$-0.012$, $+0.002$] & $+0.001$ [$-0.003$, $+0.005$] & $-0.007$ [$-0.039$, $+0.025$] & $+0.002$ [$-0.018$, $+0.020$] \\
DSMIL & $-0.005$ [$-0.016$, $+0.004$] & $+0.027$ [$+0.006$, $+0.059$] & $-0.004$ [$-0.011$, $+0.003$] & $+0.001$ [$-0.003$, $+0.005$] & $-0.004$ [$-0.036$, $+0.029$] & $-0.011$ [$-0.027$, $+0.006$] \\
DTFD-MIL & $-0.002$ [$-0.013$, $+0.007$] & $+0.029$ [$+0.002$, $+0.068$] & $-0.005$ [$-0.013$, $+0.002$] & $+0.001$ [$-0.002$, $+0.005$] & $-0.003$ [$-0.038$, $+0.032$] & $-0.017$ [$-0.039$, $+0.004$] \\
TransMIL & $-0.002$ [$-0.015$, $+0.010$] & $+0.026$ [$-0.012$, $+0.068$] & $-0.003$ [$-0.013$, $+0.005$] & $0.000$ [$-0.004$, $+0.004$] & $+0.022$ [$-0.010$, $+0.054$] & $-0.011$ [$-0.031$, $+0.009$] \\
max pooling & $0.000$ [$-0.023$, $+0.032$] & $+0.020$ [$-0.003$, $+0.048$] & $-0.006$ [$-0.016$, $+0.001$] & $-0.002$ [$-0.005$, $+0.002$] & $+0.012$ [$-0.020$, $+0.046$] & $+0.007$ [$-0.010$, $+0.021$] \\
mean pooling & $-0.025$ [$-0.063$, $+0.011$] & $+0.012$ [$-0.017$, $+0.038$] & $-0.004$ [$-0.013$, $+0.004$] & $+0.003$ [$-0.001$, $+0.007$] & $-0.019$ [$-0.058$, $+0.020$] & $-0.008$ [$-0.033$, $+0.020$] \\
\bottomrule\end{tabular}}\caption{Ours minus the selector fitted to slide labels, each frozen aggregator at $D$ (test AUROC). Brackets give the $95\%$ interval. Gated ABMIL is in Table~\ref{tab:ablgated}.}\label{tab:abllabels}\end{table}
\paragraph{Labels did not improve the ranking here.} For gated ABMIL the selector fitted to labels is within 0.003 of ours on five tasks, and ours is ahead on CAMELYON17 by 0.018 [$+0.002$, $+0.044$], and by 0.012 to 0.029 for the other aggregators there (Tables~\ref{tab:ablgated} and~\ref{tab:abllabels}). Fitting to each aggregator's own importance is never significantly better than fitting to the gated ABMIL (at most 0.008 ahead), and TransMIL's gradient importance falls below random tiles on both lymph-node tasks (Table~\ref{tab:ablagg}), so on the lymph-node tasks one attention model serves as the teacher for every aggregator. The model's own coarse attention behind the label-fitted gate is 0.021 to 0.058 below ours on CAMELYON17 for the four attention aggregators. This gap mixes the zoom target with the gate, since behind our attention-fitted gate the same frozen model's own coarse attention comes within $0.004$ of ours on CAMELYON17 (Sec.~\ref{app:c1}).

\section{Other selectors in front of the same frozen model}\label{app:c1}
Registered before any of these reads (with the analysis code frozen), this comparison keeps the frozen gated ABMIL and its weights and changes only which tiles it reads, at matched encoder cost. Rules behind our gate take the same thumbnail and the same $500$ coarse encodings as $D$ and differ only in which $10$ regions are zoomed, namely by the frozen model's own attention on its $5\times$ tokens (training-free), ZoomMIL's zoom scores and HAG-MIL's coarse attention (each from that method's trained model of the split). The same three rules without a gate are compared with ours without a gate. Tile screens, ADMIL's PriorNet and HDMIL's screen, score every tile and keep, per slide, as many $20\times$ tiles as $D$'s cost minus the screen's own cost. The BRACS tile-screen rows were added after the first analysis, under the same registered rules, once the coordinates needed to map those screens onto our tiles could be read. BRACS was tiled on grids not aligned to the patch size, so its map uses the exact parent rule of its cache. Two reads failed to start (a GPU initialisation error) and were relaunched unchanged before any analysis, so every registered read entered the analysis, although the registered condition of no failed read was met only after relaunch.

\begin{table}[H]\centering\scriptsize\setlength\tabcolsep{2pt}\adjustbox{max width=\linewidth}{\begin{tabular}{@{}l@{\hspace{6pt}}cccccc@{\hspace{8pt}}cccccc@{}}\toprule
Selector & \multicolumn{6}{c}{\emph{CAMELYON16} (5 runs)} & \multicolumn{6}{c}{\emph{CAMELYON17} (5 runs)} \\ \cmidrule(lr){2-7}\cmidrule(lr){8-13}
 & AUROC & F\% & $-$ours [95\% CI] & $p$ & $-$full & $-$random & AUROC & F\% & $-$ours [95\% CI] & $p$ & $-$full & $-$random \\ \midrule
ours, $D$ (ref.) & 0.985 & 4.4 &  &  & $+0.002$ & $+0.098$ & 0.963 & 4.2 &  &  & $-0.010$ & $+0.092$ \\
ours, no gate, $b{=}10$ (ref.) & 0.988 & 7.8 &  &  & $+0.005$ & $+0.067$ & 0.981 & 8.0 &  &  & $+0.008$ & $+0.088$ \\
own $5\times$ attention, our gate & 0.968 & 4.4 & $-0.017$ $[-0.041,-0.000]$ & $0.093$ / $1.000$ & $-0.015$ & $+0.082$ & 0.959 & 4.2 & $-0.004$ $[-0.024,+0.011]$ & $0.983$ / $1.000$ & $-0.014$ & $+0.088$ \\
ZoomMIL zoom scores, our gate & 0.966 & 4.3 & $-0.019$ $[-0.043,+0.001]$ & $0.099$ / $1.000$ & $-0.018$ & $+0.084$ & 0.866 & 4.1 & $-0.097$ $[-0.161,-0.056]$ & $<\!0.001$ / $1.000$ & $-0.107$ & $-0.003$ \\
HAG-MIL coarse attention, our gate & 0.744 & 4.4 & $-0.241$ $[-0.294,-0.191]$ & $<\!0.001$ / $1.000$ & $-0.240$ & $-0.144$ & 0.710 & 4.1 & $-0.253$ $[-0.328,-0.178]$ & $<\!0.001$ / $1.000$ & $-0.263$ & $-0.157$ \\
ADMIL PriorNet, top $n_i$ tiles & 0.978 & 4.4 & $-0.007$ $[-0.023,+0.008]$ & $0.238$ / $1.000$ & $-0.005$ & $+0.091$ & 0.961 & 4.2 & $-0.002$ $[-0.041,+0.039]$ & $0.983$ / $1.000$ & $-0.012$ & $+0.090$ \\
HDMIL screen, top $n_i$ tiles & 0.884 & 4.4 & $-0.100$ $[-0.151,-0.056]$ & $<\!0.001$ / $1.000$ & $-0.099$ & $-0.003$ & 0.896 & 4.2 & $-0.067$ $[-0.124,-0.024]$ & $<\!0.001$ / $1.000$ & $-0.077$ & $+0.025$ \\
own $5\times$ attention, no gate & 0.972 & 7.7 & $-0.017$ $[-0.044,+0.001]$ & $0.238$ / $1.000$ & $-0.012$ & $+0.053$ & 0.979 & 8.0 & $-0.002$ $[-0.021,+0.014]$ & $0.983$ / $1.000$ & $+0.006$ & $+0.087$ \\
ZoomMIL zoom scores, no gate & 0.970 & 7.6 & $-0.018$ $[-0.041,-0.001]$ & $0.041$ / $1.000$ & $-0.013$ & $+0.057$ & 0.851 & 7.9 & $-0.130$ $[-0.199,-0.083]$ & $<\!0.001$ / $1.000$ & $-0.122$ & $-0.046$ \\
HAG-MIL coarse attention, no gate & 0.723 & 7.7 & $-0.266$ $[-0.317,-0.216]$ & $<\!0.001$ / $1.000$ & $-0.261$ & $-0.195$ & 0.654 & 7.9 & $-0.327$ $[-0.394,-0.253]$ & $<\!0.001$ / $1.000$ & $-0.319$ & $-0.238$ \\
\cmidrule(l){1-13}
ZoomMIL zoom scores, no gate, $b{=}12$ & 0.972 & 7.8 & $-0.017$ $[-0.038,-0.001]$ & ($0.014$ / $0.987$) & $-0.012$ & $+0.054$ & 0.851 & 8.1 & $-0.130$ $[-0.198,-0.081]$ & ($<\!0.001$ / $1.000$) & $-0.122$ & $-0.046$ \\
ADMIL PriorNet, top 128 & 0.970 & 1.1 & $-0.015$ $[-0.034,+0.002]$ & ($0.041$ / $0.959$) & $-0.014$ & $+0.082$ & 0.957 & 1.1 & $-0.005$ $[-0.038,+0.022]$ & ($0.335$ / $0.665$) & $-0.016$ & $+0.086$ \\
ADMIL PriorNet, top 330 & 0.978 & 2.5 & $-0.007$ $[-0.024,+0.007]$ & ($0.180$ / $0.821$) & $-0.006$ & $+0.090$ & 0.965 & 2.4 & $+0.003$ $[-0.033,+0.043]$ & ($0.525$ / $0.475$) & $-0.008$ & $+0.094$ \\
ADMIL PriorNet, top 660 & 0.977 & 4.8 & $-0.007$ $[-0.024,+0.008]$ & ($0.174$ / $0.827$) & $-0.006$ & $+0.090$ & 0.959 & 4.5 & $-0.004$ $[-0.048,+0.038]$ & ($0.392$ / $0.608$) & $-0.015$ & $+0.087$ \\
HDMIL's own keep set (cost not matched) & 0.982 & 76.2 & $-0.003$ $[-0.015,+0.007]$ & ($0.261$ / $0.742$) & $-0.002$ & $+0.094$ & 0.969 & 48.7 & $+0.006$ $[-0.026,+0.044]$ & ($0.601$ / $0.400$) & $-0.004$ & $+0.098$ \\
\bottomrule\end{tabular}}\caption{\textbf{Lymph-node tasks.} Other selection rules in front of the same frozen gated ABMIL (Virchow2, seed 0 of each split, test AUROC). Rows behind our gate and the tile screens are compared with ours at $D$, the no-gate rows with ours with no gate at $b{=}10$. F\% counts encoder evaluations as a share of the full bag's (screen cost included for the tile screens, thumbnail for the gated rows). $p$ is one-sided, within-run patient bootstrap (10,000 draws), Holm-corrected over the eight primary rows of the task, uncorrected in parentheses for the secondary rows below the thin rule. $-$full and $-$random compare with the same model on its full bag and on random $20\times$ tiles at the selector's own cost, at $D$'s cost for every tile-screen row. Own $5\times$ attention is the frozen model's own attention on its $5\times$ tokens, ref.\ marks the two reference rows, $-$ours is the row minus ours, and $p$ is given as ours better / selector better.}\label{tab:c1lymph}\end{table}
\begin{table}[H]\centering\scriptsize\setlength\tabcolsep{2pt}\adjustbox{max width=\linewidth}{\begin{tabular}{@{}l@{\hspace{6pt}}cccccc@{\hspace{8pt}}cccccc@{}}\toprule
Selector & \multicolumn{6}{c}{\emph{BRCA} (5 runs)} & \multicolumn{6}{c}{\emph{LUNG} (5 runs)} \\ \cmidrule(lr){2-7}\cmidrule(lr){8-13}
 & AUROC & F\% & $-$ours [95\% CI] & $p$ & $-$full & $-$random & AUROC & F\% & $-$ours [95\% CI] & $p$ & $-$full & $-$random \\ \midrule
ours, $D$ (ref.) & 0.963 & 4.1 &  &  & $-0.008$ & $-0.008$ & 0.983 & 4.7 &  &  & $+0.002$ & $+0.003$ \\
ours, no gate, $b{=}10$ (ref.) & 0.963 & 7.6 &  &  & $-0.008$ & $-0.006$ & 0.983 & 7.9 &  &  & $+0.002$ & $+0.002$ \\
own $5\times$ attention, our gate & 0.961 & 4.1 & $-0.002$ $[-0.012,+0.009]$ & $1.000$ / $1.000$ & $-0.010$ & $-0.010$ & 0.982 & 4.7 & $-0.001$ $[-0.004,+0.002]$ & $0.601$ / $1.000$ & $+0.001$ & $+0.002$ \\
ZoomMIL zoom scores, our gate & 0.940 & 4.1 & $-0.023$ $[-0.043,-0.003]$ & $0.041$ / $1.000$ & $-0.031$ & $-0.031$ & 0.958 & 4.6 & $-0.025$ $[-0.040,-0.012]$ & $<\!0.001$ / $1.000$ & $-0.024$ & $-0.023$ \\
HAG-MIL coarse attention, our gate & 0.919 & 4.1 & $-0.044$ $[-0.082,-0.009]$ & $0.022$ / $1.000$ & $-0.053$ & $-0.053$ & 0.960 & 4.5 & $-0.023$ $[-0.035,-0.012]$ & $<\!0.001$ / $1.000$ & $-0.022$ & $-0.021$ \\
ADMIL PriorNet, top $n_i$ tiles & 0.970 & 4.1 & $+0.007$ $[-0.003,+0.019]$ & $1.000$ / $0.701$ & $-0.001$ & $-0.001$ & 0.977 & 4.7 & $-0.006$ $[-0.016,+0.001]$ & $0.183$ / $1.000$ & $-0.005$ & $-0.004$ \\
HDMIL screen, top $n_i$ tiles & 0.944 & 4.1 & $-0.019$ $[-0.036,-0.003]$ & $0.041$ / $1.000$ & $-0.027$ & $-0.027$ & 0.959 & 4.7 & $-0.024$ $[-0.040,-0.009]$ & $0.001$ / $1.000$ & $-0.023$ & $-0.022$ \\
own $5\times$ attention, no gate & 0.962 & 7.6 & $-0.002$ $[-0.012,+0.009]$ & $1.000$ / $1.000$ & $-0.010$ & $-0.008$ & 0.982 & 7.9 & $-0.001$ $[-0.004,+0.002]$ & $0.601$ / $1.000$ & $+0.001$ & $+0.002$ \\
ZoomMIL zoom scores, no gate & 0.934 & 7.6 & $-0.030$ $[-0.053,-0.007]$ & $0.022$ / $1.000$ & $-0.038$ & $-0.036$ & 0.944 & 7.7 & $-0.039$ $[-0.059,-0.021]$ & $<\!0.001$ / $1.000$ & $-0.037$ & $-0.036$ \\
HAG-MIL coarse attention, no gate & 0.881 & 7.4 & $-0.083$ $[-0.131,-0.036]$ & $<\!0.001$ / $1.000$ & $-0.091$ & $-0.089$ & 0.930 & 7.6 & $-0.053$ $[-0.072,-0.034]$ & $<\!0.001$ / $1.000$ & $-0.051$ & $-0.050$ \\
\cmidrule(l){1-13}
ZoomMIL zoom scores, no gate, $b{=}12$ & 0.936 & 7.8 & $-0.028$ $[-0.050,-0.006]$ & ($0.003$ / $0.997$) & $-0.036$ & $-0.035$ & 0.949 & 8.0 & $-0.034$ $[-0.053,-0.018]$ & ($<\!0.001$ / $1.000$) & $-0.032$ & $-0.032$ \\
ADMIL PriorNet, top 128 & 0.969 & 1.1 & $+0.006$ $[-0.005,+0.020]$ & ($0.860$ / $0.140$) & $-0.002$ & $-0.002$ & 0.974 & 1.3 & $-0.009$ $[-0.020,-0.000]$ & ($0.017$ / $0.983$) & $-0.008$ & $-0.007$ \\
ADMIL PriorNet, top 330 & 0.971 & 2.4 & $+0.008$ $[-0.003,+0.020]$ & ($0.921$ / $0.079$) & $-0.001$ & $-0.001$ & 0.976 & 3.0 & $-0.007$ $[-0.018,+0.000]$ & ($0.034$ / $0.966$) & $-0.006$ & $-0.005$ \\
ADMIL PriorNet, top 660 & 0.971 & 4.6 & $+0.008$ $[-0.003,+0.020]$ & ($0.921$ / $0.079$) & $-0.001$ & $-0.001$ & 0.977 & 5.7 & $-0.006$ $[-0.016,+0.001]$ & ($0.076$ / $0.925$) & $-0.004$ & $-0.003$ \\
HDMIL's own keep set (cost not matched) & 0.971 & 81.0 & $+0.008$ $[-0.002,+0.020]$ & ($0.942$ / $0.058$) & $0.000$ & $0.000$ & 0.982 & 88.0 & $-0.001$ $[-0.005,+0.004]$ & ($0.386$ / $0.614$) & $+0.001$ & $+0.002$ \\
\midrule Selector & \multicolumn{6}{c}{\emph{RCC staging} (5 runs)} & \multicolumn{6}{c}{\emph{BRACS} (5 runs)} \\ \cmidrule(lr){2-7}\cmidrule(lr){8-13}
 & AUROC & F\% & $-$ours [95\% CI] & $p$ & $-$full & $-$random & AUROC & F\% & $-$ours [95\% CI] & $p$ & $-$full & $-$random \\ \midrule
ours, $D$ (ref.) & 0.764 & 4.3 &  &  & $+0.002$ & $+0.003$ & 0.919 & 4.5 &  &  & $+0.007$ & $+0.011$ \\
ours, no gate, $b{=}10$ (ref.) & 0.763 & 7.7 &  &  & $+0.001$ & $+0.002$ & 0.913 & 7.7 &  &  & $+0.001$ & $+0.003$ \\
own $5\times$ attention, our gate & 0.770 & 4.3 & $+0.007$ $[-0.015,+0.028]$ & $1.000$ / $1.000$ & $+0.009$ & $+0.010$ & 0.907 & 4.5 & $-0.012$ $[-0.026,+0.003]$ & $0.526$ / $1.000$ & $-0.006$ & $-0.002$ \\
ZoomMIL zoom scores, our gate & 0.711 & 4.1 & $-0.053$ $[-0.090,-0.017]$ & $0.016$ / $1.000$ & $-0.050$ & $-0.050$ & 0.919 & 4.5 & $0.000$ $[-0.034,+0.023]$ & $1.000$ / $1.000$ & $+0.007$ & $+0.011$ \\
HAG-MIL coarse attention, our gate & 0.740 & 4.2 & $-0.024$ $[-0.064,+0.015]$ & $0.569$ / $1.000$ & $-0.022$ & $-0.021$ & 0.909 & 4.5 & $-0.010$ $[-0.028,+0.009]$ & $0.955$ / $1.000$ & $-0.003$ & $+0.001$ \\
ADMIL PriorNet, top $n_i$ tiles & 0.760 & 4.3 & $-0.004$ $[-0.039,+0.031]$ & $1.000$ / $1.000$ & $-0.002$ & $-0.001$ & 0.928 & 4.5 & $+0.009$ $[-0.008,+0.027]$ & $1.000$ / $1.000$ & $+0.016$ & $+0.020$ \\
HDMIL screen, top $n_i$ tiles & 0.720 & 4.3 & $-0.044$ $[-0.081,-0.006]$ & $0.076$ / $1.000$ & $-0.041$ & $-0.040$ & 0.914 & 4.5 & $-0.005$ $[-0.027,+0.009]$ & $1.000$ / $1.000$ & $+0.001$ & $+0.006$ \\
own $5\times$ attention, no gate & 0.780 & 7.6 & $+0.017$ $[-0.005,+0.040]$ & $1.000$ / $0.542$ & $+0.018$ & $+0.019$ & 0.908 & 7.7 & $-0.005$ $[-0.015,+0.008]$ & $1.000$ / $1.000$ & $-0.005$ & $-0.003$ \\
ZoomMIL zoom scores, no gate & 0.681 & 7.2 & $-0.082$ $[-0.131,-0.036]$ & $0.002$ / $1.000$ & $-0.081$ & $-0.080$ & 0.924 & 7.7 & $+0.010$ $[-0.006,+0.025]$ & $1.000$ / $0.778$ & $+0.011$ & $+0.013$ \\
HAG-MIL coarse attention, no gate & 0.738 & 7.6 & $-0.025$ $[-0.070,+0.018]$ & $0.569$ / $1.000$ & $-0.024$ & $-0.023$ & 0.907 & 7.7 & $-0.006$ $[-0.023,+0.013]$ & $1.000$ / $1.000$ & $-0.005$ & $-0.003$ \\
\cmidrule(l){1-13}
ZoomMIL zoom scores, no gate, $b{=}12$ & 0.687 & 7.3 & $-0.076$ $[-0.122,-0.031]$ & ($<\!0.001$ / $1.000$) & $-0.074$ & $-0.073$ & 0.923 & 7.9 & $+0.010$ $[-0.007,+0.028]$ & ($0.889$ / $0.111$) & $+0.010$ & $+0.014$ \\
ADMIL PriorNet, top 128 & 0.759 & 1.1 & $-0.005$ $[-0.045,+0.034]$ & ($0.395$ / $0.605$) & $-0.003$ & $-0.002$ & 0.919 & 1.1 & $0.000$ $[-0.013,+0.018]$ & ($0.564$ / $0.436$) & $+0.007$ & $+0.011$ \\
ADMIL PriorNet, top 330 & 0.759 & 2.5 & $-0.005$ $[-0.042,+0.032]$ & ($0.399$ / $0.601$) & $-0.002$ & $-0.001$ & 0.925 & 2.5 & $+0.006$ $[-0.008,+0.021]$ & ($0.804$ / $0.196$) & $+0.013$ & $+0.017$ \\
ADMIL PriorNet, top 660 & 0.760 & 4.8 & $-0.004$ $[-0.039,+0.030]$ & ($0.414$ / $0.586$) & $-0.002$ & $0.000$ & 0.928 & 4.8 & $+0.009$ $[-0.009,+0.027]$ & ($0.834$ / $0.167$) & $+0.015$ & $+0.020$ \\
HDMIL's own keep set (cost not matched) & 0.761 & 54.8 & $-0.003$ $[-0.031,+0.026]$ & ($0.423$ / $0.577$) & $0.000$ & $+0.001$ & 0.907 & 92.8 & $-0.012$ $[-0.047,+0.015]$ & ($0.192$ / $0.809$) & $-0.005$ & $-0.001$ \\
\bottomrule\end{tabular}}\caption{\textbf{Diffuse tasks.} As Table~\ref{tab:c1lymph}, on BRCA, LUNG, RCC staging and BRACS (Virchow2, seed 0 of each split, BRACS 5 seeds, test AUROC).}\label{tab:c1diffuse}\end{table}
\paragraph{Registered decisions.} The registered lymph-node decision holds, as on CAMELYON16 and CAMELYON17 no rule behind our gate or tile screen is ahead of ours at $D$, and no rule without a gate is ahead of ours without a gate, at Holm-corrected one-sided $p<0.05$. No rule is significantly ahead of ours on a lymph-node task, the registered warning case. The point on the frozen model's own attention is unresolved either way. Ours is not significantly ahead of it on either lymph-node task after Holm correction (ours minus it $+0.017$ on CAMELYON16, Holm $p=0.093$, and $+0.004$ on CAMELYON17, Holm $p=0.983$), and the registered condition for calling the fit useless, both within $0.005$ on all six tasks, also fails (CAMELYON16 $-0.017$, RCC staging $+0.007$, BRACS $-0.012$, the frozen model's attention minus ours). So on these tasks neither the fitted selector nor the frozen model's own coarse attention is significantly ahead of the other. The model's own coarse attention needs no fitting but exists only for a model with attention on its coarse tokens, while the fitted selector is also read by TransMIL, mean and max pooling (Table~\ref{tab:frozen}), which have none.
\paragraph{Where the other rules fall behind.} HAG-MIL's coarse attention falls $0.241$ and $0.253$ behind ours on CAMELYON16 and CAMELYON17, below random tiles at the same cost, and HDMIL's screen $0.100$ and $0.067$. HDMIL's screen varies little within a slide (median standard deviation of its scores $0.001$ to $0.008$ over the CAMELYON17 splits, $0.001$ to $0.019$ on CAMELYON16), and on CAMELYON17 its own keep set, which reads about half of the tiles there, stays within $0.004$ of the full bag. ZoomMIL's zoom scores lose $0.019$ on CAMELYON16 and $0.097$ on CAMELYON17. ADMIL's screen is the closest rule on both lymph-node tasks, closer than the frozen model's own attention ($-0.007$ and $-0.002$), and on BRCA and BRACS it is $0.007$ and $0.009$ ahead of ours, not significantly. On the diffuse tasks the frozen model's own attention and ADMIL's screen are within $0.017$ of ours, while ZoomMIL's and HAG-MIL's scores and HDMIL's screen lose up to $0.083$ on BRCA, LUNG and RCC staging.
\paragraph{Why HAG-MIL's ranking falls below random tiles} (split 0). The transplant is faithful. HAG-MIL's low model, fed the same $5\times$ rows, classifies the 128 CAMELYON16 test slides with AUROC $0.970$, and its attention equals the saved ranking (largest difference $1.7\cdot10^{-6}$). Yet its top $10$ regions behind our gate touch a lesion on 23\% of the 48 annotated tumour slides, against 38\% for $10$ random regions of the same gate, 85\% for ZoomMIL's ranking on the same rows and mapping, and 96\% for ours. Its own pipeline zooms into nearly every region (up to $2{,}500$ per slide), so its ranking is never put under pressure, and its attention is computed after a transformer layer that mixes the tokens, so the model can classify well while its pooling weight is not on the lesion.

\section{The performance-first rule}\label{app:rule}\label{app:accfirst}\label{sec:ladder}
\begin{algorithm}[t]
\caption{Performance-first rule, one decision per trained model}\label{alg:ruleF}
\begin{algorithmic}[1]
\State evidence $E\gets$ training and validation slides, or validation slides alone if training AUROC of $T$ is at least $0.999$
\State $\underline\Delta(X)\gets$ $5\%$ quantile of $\mathrm{AUROC}_E(X)-\mathrm{AUROC}_E(T)$ over $2000$ class-stratified bootstrap resamples of $E$
\State ladder $\gets$ the 32 budgets with $b>0$ in decreasing encoder cost on $E$
\State $X\gets T$
\For{each budget $Y$ on the ladder}
  \If{$\underline\Delta(Y)>-\xi$} $X\gets Y$ \Else\ \textbf{break} \EndIf
\EndFor
\State \Return $X$
\end{algorithmic}
\end{algorithm}
The margin is a share of the full bag's remaining error,
\begin{equation}\label{eq:margin}
  \xi=\min\!\big(0.005,\ \max\big(0.001,\ 0.10\,(1-\mathrm{AUROC}_E(T))\big)\big).
\end{equation}
The rule tests the budgets in a fixed order, as fixed-sequence testing does, but without its error-rate guarantee, since the bound is a bootstrap quantile and the same evidence slides are reused at every step. It walks down from the full bag and stops at the first budget that fails, so a cheap budget that passes by chance below a failing one is not chosen. It is decided once per trained model and scored on that model's test slides. The ladder holds the full grid $K\in\{100,250,500,1000,N\}$, $b\in\{2,5,10,20,40,100\}$, with $(N,250)$ and $(N,500)$ added so that it also covers $20$ to $100\%$ of the slide. The rule, the ladder and the validation-only guard of Alg.~\ref{alg:ruleF} were fixed in writing before any of its decisions was seen. It is the third rule we tried on the design tasks, so its clean test is the registered out-of-sample run of Sec.~\ref{app:oosmore} (Table~\ref{tab:ruleoos}), on models and slides that took no part in building it.

\subsection{The performance-first rule across cohorts}\label{app:e8}
This analysis runs the performance-first rule unchanged on the two cross-cohort moves of Table~\ref{tab:general}, the selector fitted on one lymph-node cohort and read by the 25 frozen gated ABMIL models of the other, each model's budget chosen on its own training and validation slides of the target cohort.
\begin{table}[h]\centering\scriptsize\setlength\tabcolsep{3pt}\adjustbox{max width=\linewidth}{\begin{tabular}{@{}lcccccc@{}}\toprule
 & rule $-$ full & runs within & cost & keeps full & \multicolumn{2}{c}{chosen budget $-$ full} \\ \cmidrule(l){6-7}
Move & $[95\%$ CI$]$ & $0.005$ (\%) & (\%) & bag (\%) & train & val. \\ \midrule
C16 selector, C17 models & $+0.001$ $[0.000,+0.003]$ & 100 & 91 & 88 & $+0.001$ & $+0.002$ \\
C17 selector, C16 models & $-0.017$ $[-0.045,+0.004]$ & 20 & 13 & 0 & $-0.005$ & $0.000$ \\
\bottomrule\end{tabular}}\caption{The performance-first rule under cross-cohort transfer, frozen gated ABMIL (Virchow2, 25 runs each). The last two columns are the mean AUROC of each run's chosen budget minus its full bag on training and validation slides; on test slides it is the first column. On the CAMELYON16 targets the rule used the validation slides only, since the training AUROC of the full bag was $1.000$. C16 and C17 are CAMELYON16 and CAMELYON17.}\label{tab:e8}\end{table}
With the CAMELYON16 selector the CAMELYON17 models keep the full bag in $88\%$ of runs and the rule holds. With the CAMELYON17 selector the CAMELYON16 models never keep it, choose budgets of about $13\%$ of the cost, and lose $0.017$ on test on average, so the rule does not stay within its margin there. The frozen models fit their CAMELYON16 training slides perfectly (AUROC $1.000$), so in all 25 runs the rule reads only the validation slides, as Alg.~\ref{alg:ruleF} prescribes, and sees a loss of $0.000$ there at the budgets it chose, against $-0.017$ on the test slides ($-0.005$ on the training slides it did not use). The few validation slides understate a loss that appears on the official test set. The rule can report a cheaper reading as safe when its evidence slides are easier than the slides it will read.

\section{The $(1000,10)$ point}\label{app:k1000}
The budget $(1000,10)$ keeps $1000$ regions at the coarse level and zooms into $10$. We registered its test on the other five aggregators and on every out-of-sample setting before reading them, and the gated ABMIL design rows below are not counted as confirmation. Its registration allowed calling it close to the full bag only if its mean AUROC is at most $0.005$ below the full bag in at least 30 of the 36 design pairs. It is in 33, against 22 for $D$ and 32 without the gate, counted on unrounded differences (three pairs lie between $-0.0055$ and $-0.005$ and print as $-0.005$, gated ABMIL on BRCA at $D$, DTFD-MIL on BRCA at $D$, TransMIL on CAMELYON16 without the gate), at $6.3$ to $6.8\%$ of the encoder cost against $4.1$ to $4.7\%$ for $D$ and $7.6$ to $8.1\%$ without the gate. The exceptions are DTFD-MIL on BRACS ($-0.006$), TransMIL on CAMELYON17 ($-0.018$), TransMIL on BRACS ($-0.034$). The one-sided $95\%$ lower bound of a patient bootstrap within runs is above $-0.010$ in 17 of 36 pairs (10 at $D$), so single pairs are mostly not shown non-inferior, and a single run is within $0.005$ in 70\% of runs on average over the pairs. Of the 30 pairs of the other five aggregators, read only after the budget was fixed, 27 are within $0.005$. Out of sample the bound is above $-0.010$ in 10 of 11 settings (9 at $D$), where the registered confirmation was all 11. On the PANDA biopsies every slide has fewer than $500$ regions, so the three readings coincide.
\begin{table*}[h]\centering\scriptsize\setlength\tabcolsep{3pt}\adjustbox{max width=\linewidth}{\begin{tabular}{@{}lcccccccccccccccccc@{}}\toprule
 & \multicolumn{3}{c}{CAMELYON16} & \multicolumn{3}{c}{CAMELYON17} & \multicolumn{3}{c}{BRCA} & \multicolumn{3}{c}{LUNG} & \multicolumn{3}{c}{RCC staging} & \multicolumn{3}{c}{BRACS} \\ \cmidrule(lr){2-4}\cmidrule(lr){5-7}\cmidrule(lr){8-10}\cmidrule(lr){11-13}\cmidrule(lr){14-16}\cmidrule(lr){17-19}
Frozen model & $D$ & $(1000,10)$ & no gate & $D$ & $(1000,10)$ & no gate & $D$ & $(1000,10)$ & no gate & $D$ & $(1000,10)$ & no gate & $D$ & $(1000,10)$ & no gate & $D$ & $(1000,10)$ & no gate \\ \midrule
gated ABMIL & $-0.003$ ($-0.011$) & $+0.002$ ($-0.003$) & $0.000$ ($-0.007$) & $-0.012$ ($-0.039$) & $+0.003$ ($-0.006$) & $+0.004$ ($-0.004$) & $-0.005$ ($-0.012$) & $-0.005$ ($-0.011$) & $-0.004$ ($-0.011$) & $+0.002$ ($-0.001$) & $+0.002$ ($-0.001$) & $+0.002$ ($-0.001$) & $+0.006$ ($-0.016$) & $+0.002$ ($-0.019$) & $+0.003$ ($-0.019$) & $+0.007$ ($-0.015$) & $+0.005$ ($-0.015$) & $+0.001$ ($-0.024$) \\
ABMIL & $-0.002$ ($-0.010$) & $+0.001$ ($-0.003$) & $-0.001$ ($-0.007$) & $-0.014$ ($-0.043$) & $+0.001$ ($-0.005$) & $+0.002$ ($-0.003$) & $-0.005$ ($-0.012$) & $-0.003$ ($-0.008$) & $-0.003$ ($-0.008$) & $+0.002$ ($-0.000$) & $+0.002$ ($-0.001$) & $+0.002$ ($-0.001$) & $+0.002$ ($-0.017$) & $-0.003$ ($-0.023$) & $-0.002$ ($-0.023$) & $+0.012$ ($-0.008$) & $+0.009$ ($-0.011$) & $+0.005$ ($-0.020$) \\
DSMIL & $+0.003$ ($-0.009$) & $+0.006$ ($-0.005$) & $+0.006$ ($-0.003$) & $-0.008$ ($-0.035$) & $+0.008$ ($+0.002$) & $+0.009$ ($+0.003$) & $-0.006$ ($-0.015$) & $-0.004$ ($-0.010$) & $-0.004$ ($-0.010$) & $+0.002$ ($-0.001$) & $+0.002$ ($-0.002$) & $+0.002$ ($-0.002$) & $+0.006$ ($-0.016$) & $+0.005$ ($-0.015$) & $+0.004$ ($-0.016$) & $+0.007$ ($-0.017$) & $+0.009$ ($-0.017$) & $+0.002$ ($-0.025$) \\
DTFD-MIL & $-0.004$ ($-0.012$) & $-0.003$ ($-0.011$) & $-0.003$ ($-0.012$) & $-0.012$ ($-0.042$) & $+0.004$ ($-0.005$) & $+0.006$ ($-0.000$) & $-0.005$ ($-0.011$) & $-0.004$ ($-0.010$) & $-0.004$ ($-0.009$) & $+0.002$ ($-0.001$) & $+0.002$ ($-0.001$) & $+0.002$ ($-0.001$) & $+0.011$ ($-0.012$) & $+0.008$ ($-0.016$) & $+0.007$ ($-0.017$) & $-0.006$ ($-0.028$) & $-0.006$ ($-0.028$) & $-0.013$ ($-0.044$) \\
TransMIL & $-0.010$ ($-0.021$) & $-0.005$ ($-0.015$) & $-0.005$ ($-0.014$) & $-0.035$ ($-0.071$) & $-0.018$ ($-0.046$) & $-0.012$ ($-0.029$) & $0.000$ ($-0.007$) & $+0.002$ ($-0.004$) & $+0.002$ ($-0.004$) & $+0.001$ ($-0.002$) & $+0.001$ ($-0.003$) & $+0.001$ ($-0.003$) & $+0.002$ ($-0.020$) & $+0.006$ ($-0.017$) & $+0.004$ ($-0.019$) & $-0.034$ ($-0.055$) & $-0.034$ ($-0.055$) & $-0.040$ ($-0.062$) \\
max pooling & $-0.012$ ($-0.025$) & $-0.004$ ($-0.014$) & $-0.003$ ($-0.011$) & $-0.013$ ($-0.036$) & $0.000$ ($-0.012$) & $+0.002$ ($-0.003$) & $-0.007$ ($-0.015$) & $-0.004$ ($-0.010$) & $-0.004$ ($-0.010$) & $+0.002$ ($-0.001$) & $+0.001$ ($-0.002$) & $+0.001$ ($-0.002$) & $-0.001$ ($-0.024$) & $+0.002$ ($-0.019$) & $+0.003$ ($-0.018$) & $-0.001$ ($-0.019$) & $-0.001$ ($-0.019$) & $-0.004$ ($-0.027$) \\
\bottomrule\end{tabular}}\caption{Design tasks, test AUROC minus the full bag of the same model, mean over runs (25, BRACS 5), with the one-sided $95\%$ lower bound in brackets (patient bootstrap within runs, models fixed). A bound printed as $-0.000$ is below zero.}\label{tab:k1000design}\end{table*}
\begin{table}[h]\centering\scriptsize\setlength\tabcolsep{2pt}\adjustbox{max width=\linewidth}{\begin{tabular}{@{}lccc@{}}\toprule
Setting & $D$ & $(1000,10)$ & no gate \\ \midrule
PANDA cancer, gated ABMIL & $-0.001$ ($-0.001$) 37.4\% & $-0.001$ ($-0.001$) 37.4\% & $-0.001$ ($-0.001$) 37.4\% \\
PANDA cancer, DSMIL & $-0.001$ ($-0.001$) 37.4\% & $-0.001$ ($-0.001$) 37.4\% & $-0.001$ ($-0.001$) 37.4\% \\
PANDA cancer, TransMIL & $0.000$ ($-0.001$) 37.4\% & $0.000$ ($-0.001$) 37.4\% & $0.000$ ($-0.001$) 37.4\% \\
PANDA ISUP, gated ABMIL & $-0.003$ ($-0.003$) 37.2\% & $-0.003$ ($-0.003$) 37.2\% & $-0.003$ ($-0.003$) 37.2\% \\
PANDA ISUP, DSMIL & $-0.003$ ($-0.004$) 37.2\% & $-0.003$ ($-0.004$) 37.2\% & $-0.003$ ($-0.004$) 37.2\% \\
PANDA ISUP, TransMIL & $-0.008$ ($-0.009$) 37.2\% & $-0.008$ ($-0.009$) 37.2\% & $-0.008$ ($-0.009$) 37.2\% \\
KIRC grade, gated ABMIL & $+0.022$ ($+0.004$) 4.1\% & $+0.020$ ($+0.002$) 6.4\% & $+0.020$ ($+0.002$) 7.6\% \\
KIRC grade, DSMIL & $+0.033$ ($+0.014$) 4.1\% & $+0.030$ ($+0.011$) 6.4\% & $+0.030$ ($+0.011$) 7.6\% \\
KIRC grade, TransMIL & $+0.069$ ($+0.043$) 4.1\% & $+0.067$ ($+0.042$) 6.4\% & $+0.066$ ($+0.040$) 7.6\% \\
CAMELYON17, centre left out & $+0.003$ ($-0.015$) 4.7\% & $+0.013$ ($-0.001$) 6.9\% & $+0.012$ ($-0.001$) 8.2\% \\
SLN-Breast & $-0.007$ ($-0.016$) 5.9\% & $-0.003$ ($-0.011$) 8.0\% & $-0.003$ ($-0.011$) 8.8\% \\
\midrule
CAMELYON17, seen centre (control) & $-0.018$ ($-0.039$) 4.4\% & $-0.010$ ($-0.031$) 6.6\% & $-0.010$ ($-0.030$) 8.1\% \\
CAMELYON16 selector on CAMELYON17 & $-0.025$ ($-0.048$) 4.2\% & $-0.004$ ($-0.022$) 6.5\% & $+0.003$ ($-0.013$) 8.0\% \\
CAMELYON17 selector on CAMELYON16 & $-0.006$ ($-0.026$) 4.4\% & $+0.008$ ($-0.000$) 6.6\% & $+0.008$ ($0.000$) 7.8\% \\
\bottomrule\end{tabular}}\caption{Out of sample, the seen-centre control and the cohort transfers, as Table~\ref{tab:k1000design}, with the encoder cost. The same settings against random tiles are in Table~\ref{tab:k1000rand}, and under the performance-first rule in Table~\ref{tab:ruleoos}.}\label{tab:k1000oos}\end{table}

\section{Further analyses of the frozen reading}\label{app:review}
All analyses here were fixed in writing before their results and are exploratory. They use saved predictions of the same frozen full-bag models.
\paragraph{Operating points at a validation threshold.} The threshold is set at $95\%$ specificity on each run's validation negatives (30 to 34 on these tasks, at the interpolated $95$th percentile of their scores) and applied to its test slides, for $D$ and the full bag of the same model. Table~\ref{tab:valops} gives means over 25 runs, with $95\%$ intervals resampling runs, and counts misses by the slide's largest lesion (CAMELYON16 annotations, CAMELYON17 stage labels).
\begin{table}[H]\centering\scriptsize\setlength\tabcolsep{2pt}\adjustbox{max width=\linewidth}{\begin{tabular}{@{}l@{\hspace{6pt}}ccccc@{\hspace{6pt}}ccccc@{}}\toprule
 & \multicolumn{5}{c}{CAMELYON16} & \multicolumn{5}{c}{CAMELYON17} \\ \cmidrule(lr){2-6}\cmidrule(lr){7-11}
Frozen model & spec.\ $D$ / full & sens.\ $D$ / full & sens.\ $D-$full & ITC missed $D$ / full & micro missed $D$ / full & spec.\ $D$ / full & sens.\ $D$ / full & sens.\ $D-$full & ITC missed $D$ / full & micro missed $D$ / full \\ \midrule
gated ABMIL & 0.92 / 0.92 & 0.964 / 0.971 & $-0.007$ $[-0.016,+0.002]$ & 0.40 $[0.36,0.45]$ / 0.31 & 0.02 / 0.02 & 0.90 / 0.92 & 0.928 / 0.943 & $-0.015$ $[-0.032,+0.001]$ & 0.36 $[0.22,0.52]$ / 0.20 & 0.07 / 0.07 \\
ABMIL & 0.91 / 0.93 & 0.963 / 0.969 & $-0.006$ $[-0.017,+0.005]$ & 0.39 $[0.33,0.44]$ / 0.32 & 0.02 / 0.02 & 0.91 / 0.90 & 0.924 / 0.945 & $-0.022$ $[-0.035,-0.006]$ & 0.38 $[0.25,0.53]$ / 0.19 & 0.07 / 0.07 \\
DSMIL & 0.92 / 0.92 & 0.919 / 0.941 & $-0.022$ $[-0.041,-0.007]$ & 0.51 $[0.43,0.57]$ / 0.57 & 0.08 / 0.04 & 0.92 / 0.89 & 0.923 / 0.944 & $-0.021$ $[-0.041,-0.002]$ & 0.35 $[0.21,0.50]$ / 0.19 & 0.09 / 0.08 \\
DTFD-MIL & 0.88 / 0.90 & 0.966 / 0.976 & $-0.010$ $[-0.017,-0.002]$ & 0.40 $[0.35,0.45]$ / 0.29 & 0.02 / 0.01 & 0.90 / 0.91 & 0.930 / 0.950 & $-0.020$ $[-0.039,-0.000]$ & 0.35 $[0.20,0.49]$ / 0.18 & 0.07 / 0.06 \\
TransMIL & 0.90 / 0.90 & 0.951 / 0.971 & $-0.020$ $[-0.031,-0.010]$ & 0.43 $[0.37,0.48]$ / 0.28 & 0.04 / 0.02 & 0.94 / 0.93 & 0.906 / 0.926 & $-0.020$ $[-0.050,+0.008]$ & 0.38 $[0.25,0.52]$ / 0.24 & 0.08 / 0.10 \\
max pooling & 0.87 / 0.91 & 0.966 / 0.967 & $-0.002$ $[-0.009,+0.006]$ & 0.40 $[0.36,0.45]$ / 0.36 & 0.02 / 0.02 & 0.92 / 0.92 & 0.926 / 0.946 & $-0.021$ $[-0.037,-0.006]$ & 0.36 $[0.22,0.51]$ / 0.18 & 0.07 / 0.07 \\
\bottomrule\end{tabular}}\caption{Operating points with the threshold set on validation slides. ITC means isolated tumour cells. ITC and micro missed give the share of each run's test slides with only isolated tumour cells or micrometastases that the reader misses, mean over runs. A test set holds 3 ITC and 27 micrometastasis slides on CAMELYON16, and about 3 and 7.2 on CAMELYON17.}\label{tab:valops}\end{table}
With 3 ITC slides per test set (the same 3 in every CAMELYON16 run, 2 to 4 on CAMELYON17), the ITC rates rest on very few slides. 
\paragraph{Cross-cohort transfer of our selector} (registered before its results). The selector is fitted on one lymph-node cohort's training slides and read by the 25 frozen models of the other, and random tiles are matched to its own count.
\begin{table}[H]\centering\scriptsize\setlength\tabcolsep{2pt}\adjustbox{max width=\linewidth}{\begin{tabular}{@{}lcccccc@{}}\toprule Direction & transferred & in-domain & full & random & $-$ in-domain & $-$ random \\ \midrule
CAMELYON16 selector, CAMELYON17 models & 0.949 & 0.962 & 0.974 & 0.803 & $-0.013$ $[-0.053,+0.028]$ & $+0.146$ $[+0.095,+0.208]$ \\
CAMELYON17 selector, CAMELYON16 models & 0.982 & 0.985 & 0.988 & 0.812 & $-0.003$ $[-0.028,+0.014]$ & $+0.170$ $[+0.115,+0.232]$ \\
\bottomrule\end{tabular}}\caption{The attention-fitted selector moved between the lymph-node cohorts (Virchow2, 25 runs, within-run bootstrap). Differences are the transferred selector minus the column named.}\label{tab:lfcross}\end{table}
The transferred selector minus the full bag is $-0.025$ $[-0.055,-0.002]$ with the CAMELYON16 selector read by the CAMELYON17 models, and $-0.006$ $[-0.030,+0.009]$ with the CAMELYON17 selector read by the CAMELYON16 models (same bootstrap), the $D$ minus full bag values of Table~\ref{tab:general} of the main paper.
\paragraph{The sensitivity side of the operating point.} The threshold is set on each run's validation positives for $98\%$ sensitivity and applied to its test slides, for $D$ and the full bag of the same model (25 runs). On CAMELYON16 the few validation positives make the threshold itself unstable, so test sensitivity lands near $0.9$ for both readers. At this threshold the selection at $D$ gives up $0.05$ to $0.19$ of specificity on CAMELYON16 and $0.02$ to $0.19$ of specificity on CAMELYON17 for about the same sensitivity, as differences of the unrounded means, which can differ by $0.01$ from the difference of the printed entries.
\begin{table}[H]\centering\scriptsize\setlength\tabcolsep{2pt}\adjustbox{max width=\linewidth}{\begin{tabular}{@{}l@{\hspace{6pt}}cccc@{\hspace{6pt}}cccc@{}}\toprule
 & \multicolumn{4}{c}{CAMELYON16} & \multicolumn{4}{c}{CAMELYON17} \\ \cmidrule(lr){2-5}\cmidrule(lr){6-9}
Frozen model & spec.\ $D$ & spec.\ full & sens.\ $D$ & sens.\ full & spec.\ $D$ & spec.\ full & sens.\ $D$ & sens.\ full \\ \midrule
gated ABMIL & 0.82 & 1.00 & 0.891 & 0.872 & 0.59 & 0.65 & 0.944 & 0.951 \\
ABMIL & 0.82 & 1.00 & 0.887 & 0.874 & 0.58 & 0.66 & 0.949 & 0.962 \\
DSMIL & 0.81 & 1.00 & 0.909 & 0.876 & 0.54 & 0.71 & 0.955 & 0.960 \\
DTFD-MIL & 0.87 & 1.00 & 0.918 & 0.891 & 0.59 & 0.77 & 0.950 & 0.949 \\
TransMIL & 0.83 & 0.99 & 0.923 & 0.908 & 0.59 & 0.69 & 0.944 & 0.958 \\
max pooling & 0.95 & 1.00 & 0.903 & 0.869 & 0.66 & 0.68 & 0.937 & 0.955 \\
\bottomrule\end{tabular}}\caption{Test specificity and sensitivity at a validation threshold set for $98\%$ sensitivity. Misses by lesion size are in Tables~\ref{tab:valops} and~\ref{tab:smalllesion}.}\label{tab:sens98}\end{table}
\paragraph{External lymph-node test of the frozen reading} (registered before its result). The 25 frozen CAMELYON17 gated ABMIL models of Table~\ref{tab:frozen} are read on SLN-Breast (130 slides, 36 metastatic) through the CAMELYON17-fitted selector of their split, at $D$, on their full bag and on count-matched random tiles. Intervals use a slide bootstrap (no patient identifiers) with 10{,}000 draws shared by the runs. Slides of the institution that provides SLN-Breast were used to pretrain Virchow2.
\begin{table}[H]\centering\scriptsize\setlength\tabcolsep{4pt}\begin{tabular}{@{}lccc@{}}\toprule & $D$ & full bag & random tiles \\ \midrule
AUROC & 0.983 & 0.990 & 0.932 \\
$D$ minus & & $-0.007$ $[-0.019,+0.001]$ & $+0.052$ $[+0.013,+0.102]$ \\
specificity, carried threshold & 0.894 & 0.886 & \\
sensitivity, carried threshold & 0.971 & 0.992 & \\
sensitivity, threshold re-set on SLN negatives & 0.963 & 0.984 & \\
metastatic slides missed at $D$ and found by the full bag, per run & 0.88 (carried), 1.12 (re-set) & & \\
\bottomrule\end{tabular}\caption{The frozen CAMELYON17 gated ABMIL on SLN-Breast. The carried threshold is $95\%$ specificity on the run's CAMELYON17 validation negatives. The missed slides are not broken down by lesion size because the cohort carries no lesion annotations.}\label{tab:slnfrozen}\end{table}
\paragraph{The fast budget is not tuned.} Table~\ref{tab:neighD} gives the test AUROC minus the full bag at $D$ and at the eight budgets around it, for every frozen aggregator and task (mean over runs). The rank is $D$'s place among the nine (1 best).
\begin{table}[H]\centering\scriptsize\setlength\tabcolsep{2pt}\adjustbox{max width=\linewidth}{\begin{tabular}{@{}l@{\hspace{6pt}}cccccccccc@{\hspace{6pt}}cccccccccc@{}}\toprule
Frozen model & $(250,5)$ & $(250,10)$ & $(250,20)$ & $(500,5)$ & $(500,10)$ & $(500,20)$ & $(1000,5)$ & $(1000,10)$ & $(1000,20)$ & rank & $(250,5)$ & $(250,10)$ & $(250,20)$ & $(500,5)$ & $(500,10)$ & $(500,20)$ & $(1000,5)$ & $(1000,10)$ & $(1000,20)$ & rank \\
\midrule & \multicolumn{10}{c}{\emph{CAMELYON16}} & \multicolumn{10}{c}{\emph{CAMELYON17}} \\ \cmidrule(lr){2-11}\cmidrule(lr){12-21}
gated ABMIL & $-0.019$ & $-0.018$ & $-0.018$ & $-0.003$ & $-0.003$ & $-0.004$ & $-0.001$ & $+0.002$ & $+0.001$ & 5 & $-0.036$ & $-0.037$ & $-0.035$ & $-0.013$ & $-0.012$ & $-0.012$ & $+0.003$ & $+0.003$ & $+0.004$ & 5 \\
ABMIL & $-0.019$ & $-0.018$ & $-0.018$ & $-0.003$ & $-0.002$ & $-0.003$ & $-0.002$ & $+0.001$ & $+0.001$ & 4 & $-0.036$ & $-0.037$ & $-0.037$ & $-0.016$ & $-0.014$ & $-0.015$ & $+0.000$ & $+0.001$ & $+0.001$ & 4 \\
DSMIL & $-0.011$ & $-0.015$ & $-0.013$ & $+0.003$ & $+0.003$ & $+0.002$ & $+0.003$ & $+0.006$ & $+0.004$ & 5 & $-0.030$ & $-0.033$ & $-0.032$ & $-0.010$ & $-0.008$ & $-0.007$ & $+0.005$ & $+0.008$ & $+0.005$ & 5 \\
DTFD-MIL & $-0.014$ & $-0.016$ & $-0.015$ & $-0.004$ & $-0.004$ & $-0.006$ & $-0.005$ & $-0.003$ & $-0.004$ & 4 & $-0.035$ & $-0.038$ & $-0.036$ & $-0.010$ & $-0.012$ & $-0.009$ & $+0.006$ & $+0.004$ & $+0.006$ & 6 \\
TransMIL & $-0.036$ & $-0.033$ & $-0.034$ & $-0.011$ & $-0.010$ & $-0.010$ & $-0.009$ & $-0.005$ & $-0.006$ & 4 & $-0.062$ & $-0.056$ & $-0.049$ & $-0.031$ & $-0.035$ & $-0.030$ & $-0.011$ & $-0.018$ & $-0.013$ & 6 \\
max pooling & $-0.038$ & $-0.038$ & $-0.039$ & $-0.009$ & $-0.012$ & $-0.012$ & $-0.006$ & $-0.004$ & $-0.004$ & 5 & $-0.033$ & $-0.035$ & $-0.035$ & $-0.014$ & $-0.013$ & $-0.014$ & $+0.001$ & $-0.000$ & $-0.001$ & 4 \\
\midrule & \multicolumn{10}{c}{\emph{BRCA}} & \multicolumn{10}{c}{\emph{LUNG}} \\ \cmidrule(lr){2-11}\cmidrule(lr){12-21}
gated ABMIL & $-0.009$ & $-0.006$ & $-0.004$ & $-0.010$ & $-0.005$ & $-0.003$ & $-0.008$ & $-0.005$ & $-0.003$ & 5 & $+0.002$ & $+0.002$ & $+0.002$ & $+0.001$ & $+0.002$ & $+0.002$ & $+0.001$ & $+0.002$ & $+0.002$ & 1 \\
ABMIL & $-0.008$ & $-0.005$ & $-0.003$ & $-0.009$ & $-0.005$ & $-0.002$ & $-0.006$ & $-0.003$ & $-0.002$ & 5 & $+0.002$ & $+0.002$ & $+0.001$ & $+0.001$ & $+0.002$ & $+0.002$ & $+0.001$ & $+0.002$ & $+0.002$ & 1 \\
DSMIL & $-0.011$ & $-0.007$ & $-0.003$ & $-0.012$ & $-0.006$ & $-0.002$ & $-0.008$ & $-0.004$ & $-0.003$ & 5 & $+0.003$ & $+0.002$ & $+0.002$ & $+0.002$ & $+0.002$ & $+0.002$ & $+0.002$ & $+0.002$ & $+0.002$ & 6 \\
DTFD-MIL & $-0.008$ & $-0.006$ & $-0.004$ & $-0.008$ & $-0.005$ & $-0.004$ & $-0.006$ & $-0.004$ & $-0.003$ & 5 & $+0.001$ & $+0.001$ & $+0.001$ & $+0.001$ & $+0.002$ & $+0.001$ & $+0.001$ & $+0.002$ & $+0.001$ & 2 \\
TransMIL & $-0.002$ & $-0.001$ & $+0.001$ & $-0.003$ & $-0.000$ & $+0.003$ & $-0.001$ & $+0.002$ & $+0.002$ & 5 & $+0.001$ & $+0.000$ & $+0.001$ & $+0.001$ & $+0.001$ & $+0.001$ & $+0.001$ & $+0.001$ & $+0.001$ & 2 \\
max pooling & $-0.011$ & $-0.008$ & $-0.003$ & $-0.011$ & $-0.007$ & $-0.003$ & $-0.008$ & $-0.004$ & $-0.003$ & 5 & $+0.001$ & $+0.002$ & $+0.001$ & $+0.001$ & $+0.002$ & $+0.001$ & $+0.001$ & $+0.001$ & $+0.001$ & 1 \\
\midrule & \multicolumn{10}{c}{\emph{RCC staging}} & \multicolumn{10}{c}{\emph{BRACS}} \\ \cmidrule(lr){2-11}\cmidrule(lr){12-21}
gated ABMIL & $-0.006$ & $+0.003$ & $+0.006$ & $-0.005$ & $+0.006$ & $+0.007$ & $-0.001$ & $+0.002$ & $+0.005$ & 3 & $+0.001$ & $+0.008$ & $+0.010$ & $-0.001$ & $+0.007$ & $+0.009$ & $-0.004$ & $+0.005$ & $+0.008$ & 5 \\
ABMIL & $-0.009$ & $+0.000$ & $+0.004$ & $-0.008$ & $+0.002$ & $+0.005$ & $-0.004$ & $-0.003$ & $+0.001$ & 3 & $+0.007$ & $+0.014$ & $+0.016$ & $+0.005$ & $+0.012$ & $+0.013$ & $-0.000$ & $+0.009$ & $+0.011$ & 4 \\
DSMIL & $-0.005$ & $+0.005$ & $+0.010$ & $-0.005$ & $+0.006$ & $+0.009$ & $-0.003$ & $+0.005$ & $+0.006$ & 4 & $+0.007$ & $+0.008$ & $+0.010$ & $+0.004$ & $+0.007$ & $+0.009$ & $+0.005$ & $+0.009$ & $+0.008$ & 7 \\
DTFD-MIL & $-0.002$ & $+0.008$ & $+0.009$ & $+0.002$ & $+0.011$ & $+0.012$ & $+0.003$ & $+0.008$ & $+0.010$ & 2 & $-0.011$ & $-0.003$ & $+0.000$ & $-0.012$ & $-0.006$ & $-0.002$ & $-0.012$ & $-0.006$ & $-0.001$ & 5 \\
TransMIL & $-0.038$ & $+0.004$ & $+0.002$ & $-0.042$ & $+0.002$ & $-0.001$ & $-0.042$ & $+0.006$ & $-0.007$ & 3 & $-0.028$ & $-0.037$ & $-0.023$ & $-0.026$ & $-0.034$ & $-0.023$ & $-0.026$ & $-0.034$ & $-0.023$ & 8 \\
max pooling & $-0.011$ & $+0.001$ & $+0.006$ & $-0.014$ & $-0.001$ & $+0.011$ & $-0.008$ & $+0.002$ & $+0.004$ & 6 & $-0.000$ & $-0.003$ & $-0.007$ & $-0.003$ & $-0.001$ & $-0.004$ & $-0.004$ & $-0.001$ & $-0.003$ & 3 \\
\bottomrule\end{tabular}}\caption{$D=(500,10)$ against its neighbours, test AUROC minus the full bag (two tasks per block). Rank is $D$'s place among the nine budgets (1 best).}\label{tab:neighD}\end{table}

\section{Small lesions at the larger gate}\label{app:smalllesion}
The following exploratory count reads the saved predictions of the same frozen models (25 runs each) at $D$, at $(1000,10)$, without a gate (every region at the coarse level, $10$ zoomed) and on the full bag, each with its threshold set at $95\%$ specificity on the run's validation negatives at that budget. A test set holds about 3 slides with only isolated tumour cells (ITC) and 7.2 with micrometastases on CAMELYON17, and 3 and 27 on CAMELYON16, so the counts rest on few slides.
\begin{table}[H]\centering\scriptsize\setlength\tabcolsep{2pt}\adjustbox{max width=\linewidth}{\begin{tabular}{@{}l@{\hspace{6pt}}cccccccc@{\hspace{6pt}}cccccccc@{}}\toprule
 & \multicolumn{8}{c}{CAMELYON16} & \multicolumn{8}{c}{CAMELYON17} \\ \cmidrule(lr){2-9}\cmidrule(lr){10-17}
 & \multicolumn{4}{c}{ITC} & \multicolumn{4}{c}{micrometastasis} & \multicolumn{4}{c}{ITC} & \multicolumn{4}{c}{micrometastasis} \\ \cmidrule(lr){2-5}\cmidrule(lr){6-9}\cmidrule(lr){10-13}\cmidrule(lr){14-17}
Frozen model & full & $D$ & $(1000,10)$ & no gate & full & $D$ & $(1000,10)$ & no gate & full & $D$ & $(1000,10)$ & no gate & full & $D$ & $(1000,10)$ & no gate \\ \midrule
gated ABMIL & 0.92 & 1.20 & 1.00 & 1.00 & 0.52 & 0.56 & 0.28 & 0.28 & 0.60 & 0.84 & 0.44 & 0.40 & 0.40 & 0.40 & 0.40 & 0.40 \\
ABMIL & 0.96 & 1.16 & 1.00 & 1.00 & 0.56 & 0.64 & 0.28 & 0.24 & 0.56 & 0.92 & 0.52 & 0.44 & 0.40 & 0.40 & 0.40 & 0.40 \\
DSMIL & 1.72 & 1.52 & 1.44 & 1.48 & 1.16 & 2.28 & 2.08 & 1.64 & 0.52 & 0.80 & 0.44 & 0.40 & 0.44 & 0.56 & 0.60 & 0.60 \\
DTFD-MIL & 0.88 & 1.20 & 1.00 & 1.00 & 0.32 & 0.48 & 0.08 & 0.12 & 0.52 & 0.80 & 0.40 & 0.44 & 0.36 & 0.40 & 0.40 & 0.40 \\
TransMIL & 0.84 & 1.28 & 1.12 & 1.12 & 0.56 & 1.12 & 0.76 & 0.68 & 0.72 & 0.92 & 0.64 & 0.60 & 0.60 & 0.48 & 0.48 & 0.52 \\
max pooling & 1.08 & 1.20 & 1.00 & 1.00 & 0.52 & 0.48 & 0.12 & 0.12 & 0.48 & 0.84 & 0.44 & 0.44 & 0.44 & 0.44 & 0.44 & 0.44 \\
\bottomrule\end{tabular}}\caption{Slides with only isolated tumour cells (ITC) and with micrometastases missed per run at a validation threshold of $95\%$ specificity (Virchow2, mean over 25 runs).}\label{tab:smalllesion}\end{table}
On CAMELYON17 the larger gate misses no more ITC slides than the full bag for every aggregator, where $D$ misses more, and as many or fewer micrometastasis slides, except for DSMIL ($+0.16$) and DTFD-MIL ($+0.04$). On CAMELYON16 its ITC misses differ from the full bag by $-0.28$ to $+0.28$ slides per run, and it misses fewer micrometastasis slides for 4 aggregators and more for DSMIL ($+0.92$) and TransMIL ($+0.20$).

\section{Held-out tasks}\label{app:heldout}
Gated ABMIL, DSMIL and TransMIL, each its own frozen full-bag model (Virchow2, 5 runs per pair, 5 splits and one seed), are read through our selection at $D$ and compared with their full bags and with as many random tiles (within-run patient bootstrap, 10{,}000 draws, Table~\ref{tab:heldoutlf}). All but one KIRC grade test slide are in the RCC staging cohort, and 69 to 90 of the about 102 test slides of a split are in a given RCC staging training split. The KIRC grade selector and frozen models are fitted only on the KIRC grade training split, so no KIRC grade test slide is used to fit either, but these images were part of the cohort on which the selector's design was chosen.
\begin{table}[H]\centering\scriptsize\setlength\tabcolsep{3pt}\begin{tabular}{@{}llcccc@{}}\toprule
Task & Frozen model & runs & $D-$full & lower bound & $D-$random \\ \midrule
PANDA cancer & gated ABMIL & 5 & $-0.001$ $[-0.001,-0.000]$ & $-0.001$ & $+0.011$ $[+0.010,+0.013]$ \\
 & DSMIL & 5 & $-0.001$ $[-0.001,-0.000]$ & $-0.001$ & $+0.011$ $[+0.009,+0.012]$ \\
 & TransMIL$^\ast$ & 5 & $0.000$ $[-0.001,+0.000]$ & $-0.001$ & $+0.014$ $[+0.013,+0.016]$ \\
\midrule
PANDA ISUP & gated ABMIL & 5 & $-0.003$ $[-0.003,-0.002]$ & $-0.003$ & $+0.011$ $[+0.010,+0.012]$ \\
 & DSMIL & 5 & $-0.003$ $[-0.004,-0.002]$ & $-0.004$ & $+0.012$ $[+0.010,+0.013]$ \\
 & TransMIL$^\ast$ & 5 & $-0.008$ $[-0.009,-0.006]$ & $-0.009$ & $+0.012$ $[+0.010,+0.014]$ \\
\midrule
KIRC grade & gated ABMIL & 5 & $+0.022$ $[+0.001,+0.045]$ & $+0.004$ & $+0.026$ $[+0.005,+0.050]$ \\
 & DSMIL & 5 & $+0.033$ $[+0.011,+0.057]$ & $+0.014$ & $+0.046$ $[+0.020,+0.073]$ \\
 & TransMIL$^\ast$ & 5 & $+0.069$ $[+0.038,+0.102]$ & $+0.043$ & $+0.075$ $[+0.042,+0.110]$ \\
\bottomrule\end{tabular}\caption{Frozen models on the held-out tasks at $D$. Lower bound is the one-sided $95\%$ bound of $D$ minus the full bag, against a non-inferiority margin of $0.010$. KIRC grade reuses the RCC staging slides with a new label. $^\ast$TransMIL reads the bag in our gate order, so its full bag also depends on the ranking.}\label{tab:heldoutlf}\end{table}

\section{Implementation details}\label{app:impl}
\paragraph{Fitting the selector.} Each scorer is one weight vector $\omega$ with no bias. For a slide with regions $r=1,\dots,N$ and region features $z_r$, standardised with the mean and standard deviation of all regions of the training slides, the scorer gives $a_r=\mathrm{softmax}_r(z_r^{\top}\omega)$. The target $\pi$ is the attention of the frozen full-bag gated ABMIL of the same split (seed 0) on the slide's fully zoomed bag $T$, summed within each region and normalised to sum to one (Alg.~\ref{alg:fit} of the main paper). Both scorers are fitted to the same target by minimising $\mathrm{KL}(\pi\,\|\,a)$, whose gradient on one slide is $\sum_r(a_r-\pi_r)z_r$. We use plain stochastic gradient descent on the CPU with one slide per step, a constant learning rate of $10^{-3}$, no momentum, an $\ell_2$ penalty of $10^{-4}$ on $\omega$ and 25 epochs, reshuffling the slide order in every epoch. The weights start from $\mathcal N(0,0.01^2)$ with seed 0, and the coarse scorer is fitted before the gate on the same random stream. No validation slide is used and there is no early stopping, so the selector is the last iterate. The coarse scorer has $2{,}560$ weights on the Virchow2 features of the $5\times$ regions and the gate has 768 on the DINOv2 tokens of the thumbnail, $3{,}328$ in total, plus the stored means and standard deviations. With UNI2-h the coarse scorer has $1{,}536$ weights. Fitting one split takes a median of 4 to 26 minutes per task (Table~\ref{tab:traintime}), including one pass of the frozen model over the training slides. The label-fitted selector of Sec.~\ref{app:ablation} uses the same optimiser, learning rate, penalty, epochs, seed, initialisation and standardisation. It adds a logistic head on the slide label that is used only during fitting, and its gate is also pulled toward the coarse scorer's attention.

\paragraph{Hardware and software.} All timings use one NVIDIA RTX 6000 Ada Generation GPU (48\,GB, driver 575.57.08) with no other process on it, two Intel Xeon Gold 5318Y CPUs (24 cores each) with 1\,TB of memory, and slides stored on local NVMe SSDs. The software is Python 3.13.9, PyTorch 2.6.0 with CUDA 12.4, timm 1.0.29 and openslide-python 1.4.6 on OpenSlide 4.0.1. Virchow2 runs in half precision on batches of 64 tiles without \texttt{torch.compile}, and the DINOv2 thumbnail encoder runs in single precision. Sixteen reader processes on one CPU socket decode the tiles. FLOPs are counted with PyTorch's \texttt{FlopCounterMode} on one $224\times224$ tile, with each multiply-add counted as two operations, which gives 340 GFLOPs per Virchow2 evaluation.
Peak GPU memory allocated during a read, CUDA context excluded, is $2{,}183$\,MiB on each of the six tasks for the full bag, $D$ and $(1000,10)$ alike, $1{,}566$\,MiB of it the encoder weights, because the peak is set by one batch of 64 tiles and not by the number of tiles read.

\paragraph{Datasets and splits.} Table~\ref{tab:splitcounts} gives the slides of every split.
\begin{table*}[h]\centering\scriptsize\setlength{\tabcolsep}{3pt}
\caption{\textbf{Slides per split.} Each cell gives the number of slides, and in brackets the slides per class, in the order CAMELYON16 normal/tumour, CAMELYON17 normal/tumour, BRCA IDC/ILC, LUNG LUAD/LUSC, RCC staging pT1--2/pT3--4, BRACS benign/atypical/malignant, PANDA benign/cancer, KIRC grade low/high. CAMELYON16 uses its official test set with five train and validation splits, which hold the same counts. CAMELYON17 uses splits 1 to 5. BRACS has one official split, read with five seeds. PANDA counts biopsies, with the classes of cancer detection; ISUP grade uses the same biopsies. KIRC grade reuses the RCC staging slides with a new label, the GDC tumour grade, low G1--G2 (233 slides) and high G3--G4 (277), with GX and missing grades excluded. SLN-Breast is used only as an external test set of 130 slides, 94 negative and 36 with metastases. PANDA and KIRC grade use $70/10/20\%$ splits.}\label{tab:splitcounts}
\adjustbox{max width=\linewidth}{\begin{tabular}{@{}lllll@{\hspace{10pt}}lllll@{}}\toprule
Task & Split & Train & Validation & Test & Task & Split & Train & Validation & Test\\\midrule
CAMELYON16 & 0--4 & 214 (126/88) & 54 (32/22) & 128 (79/49) & RCC staging & 0 & 742 (527/215) & 90 (61/29) & 89 (64/25)\\
\cmidrule(r){1-5}
CAMELYON17 & 1 & 400 (256/144) & 49 (30/19) & 50 (32/18) &  & 1 & 743 (530/213) & 89 (62/27) & 89 (60/29)\\
 & 2 & 399 (255/144) & 50 (30/20) & 50 (33/17) &  & 2 & 740 (522/218) & 91 (70/21) & 90 (60/30)\\
 & 3 & 400 (251/149) & 50 (34/16) & 49 (33/16) &  & 3 & 732 (511/221) & 96 (72/24) & 93 (69/24)\\
 & 4 & 399 (255/144) & 50 (34/16) & 50 (29/21) &  & 4 & 743 (525/218) & 90 (66/24) & 88 (61/27)\\
\cmidrule(l){6-10}
 & 5 & 399 (253/146) & 50 (33/17) & 50 (32/18) & BRACS & official & 394 (202/52/140) & 64 (29/14/21) & 87 (32/23/32)\\
\midrule
BRCA & 0 & 732 (591/141) & 89 (73/16) & 92 (72/20) & PANDA & 0 & 7{,}430 (2{,}023/5{,}407) & 1{,}062 (290/772) & 2{,}123 (578/1{,}545)\\
 & 1 & 727 (587/140) & 92 (72/20) & 94 (77/17) &  & 1 & 7{,}430 (2{,}023/5{,}407) & 1{,}062 (290/772) & 2{,}123 (578/1{,}545)\\
 & 2 & 730 (590/140) & 88 (71/17) & 95 (75/20) &  & 2 & 7{,}430 (2{,}024/5{,}406) & 1{,}062 (289/773) & 2{,}123 (578/1{,}545)\\
 & 3 & 732 (590/142) & 89 (72/17) & 92 (74/18) &  & 3 & 7{,}430 (2{,}023/5{,}407) & 1{,}062 (289/773) & 2{,}123 (579/1{,}544)\\
 & 4 & 728 (583/145) & 95 (79/16) & 90 (74/16) &  & 4 & 7{,}430 (2{,}023/5{,}407) & 1{,}062 (290/772) & 2{,}123 (578/1{,}545)\\
\midrule
LUNG & 0 & 841 (434/407) & 111 (55/56) & 101 (52/49) & KIRC grade & 0 & 358 (164/194) & 51 (23/28) & 101 (46/55)\\
 & 1 & 840 (429/411) & 113 (64/49) & 100 (48/52) &  & 1 & 356 (163/193) & 52 (23/29) & 102 (47/55)\\
 & 2 & 842 (433/409) & 102 (51/51) & 109 (57/52) &  & 2 & 355 (162/193) & 53 (25/28) & 102 (46/56)\\
 & 3 & 842 (439/403) & 109 (53/56) & 102 (49/53) &  & 3 & 356 (162/194) & 51 (23/28) & 103 (48/55)\\
 & 4 & 843 (427/416) & 105 (57/48) & 105 (57/48) &  & 4 & 356 (162/194) & 51 (23/28) & 103 (48/55)\\
\bottomrule
\end{tabular}}
\end{table*}

\paragraph{Budgets of the baselines.} Table~\ref{tab:budgets} gives the setting that fixes how much of each slide a baseline reads, next to the deviations of Table~\ref{tab:dev}.
\begin{table*}[h]\centering\scriptsize\setlength{\tabcolsep}{4pt}
\caption{\textbf{Budget settings of the baselines.} The setting that fixes how much of each slide a baseline reads, the value we used, the value of the released code where it sets one, and the commit of the repository we ran. HDMIL fixes a target rate in its loss rather than a hard keep ratio, so its realised cost is the one in Table~\ref{tab:costflops}.}\label{tab:budgets}
\begin{tabular}{>{\raggedright\arraybackslash}p{1.5cm}>{\raggedright\arraybackslash}p{3.4cm}>{\raggedright\arraybackslash}p{5.6cm}>{\raggedright\arraybackslash}p{3.0cm}l}\toprule
Method & Budget setting & Value used & Released code & Commit\\\midrule
ZoomMIL & regions zoomed per level $k$ & 12 at $5\times$, 16 tiles each, at most 192 tiles at $20\times$. Three levels use $k=12$ at both steps & $k=12$ & \texttt{da7bb7f}\\
PathCTM & tokens passed to the next scale & 10, with 20 steps per scale and two scales & & \texttt{494d66d}\\
IPS & memory $M$ and patches per step $I$ & $M=I=5{,}000$ & $M=I=5{,}000$ & \texttt{4c17c1d}\\
HDMIL & target rate in its loss (keep ratio) & 0.6 on CAMELYON16, CAMELYON17 and RCC staging, 0.7 on BRCA, LUNG and BRACS & 0.6 on CAMELYON16 and RCC, 0.7 on BRCA and NSCLC & \texttt{ef3d468}\\
HAG-MIL & regions zoomed in training and test & $1{,}750$ and $2{,}500$ at $5\times$ & $7{,}000$ and $10{,}000$ at $10\times$ & \texttt{9fec1ca}\\
GFR-MIL & share of regions kept per step & 0.2 then 0.1, two steps & the same & \texttt{25c9b1e}\\
SASHA & share of regions visited & 0.1 and 0.2, threshold $\tau=0.95$ & 0.1 and 0.2, $\tau=0.95$ by default & \texttt{6385fe6}\\
DiagMIL & none, soft routing reads every tile & soft routing (AUROC), hard routing (cost only, Table~\ref{tab:req}) & & \texttt{dd04ff1}\\
ADMIL & tiles encoded $K$, chosen on validation from $\{1,2,4,8,16,32,64,128,180,330,660\}$ & 128 on CAMELYON16, 330 on CAMELYON17, 128 on BRCA, 4 on LUNG, 64 on RCC staging, 1 on BRACS & no public code & n/a\\
\bottomrule
\end{tabular}
\end{table*}
\section{Further out-of-sample results}\label{app:oosmore}
This section gives the rows that Tables~\ref{tab:main} and~\ref{tab:general} of the main paper summarise but that the tables above do not show: the performance-first rule on the settings that took no part in building it, the CAMELYON17 centre left out of training, and BRCA at every operating point of Table~\ref{tab:main}.

\paragraph{The unseen CAMELYON17 centre.}\label{app:lco} The setting ``CAMELYON17, centre left out'' of Tables~\ref{tab:k1000oos} and~\ref{tab:ruleoos} is a leave-one-centre-out design on CAMELYON17. Five patient folds are stratified by centre and tumour status, with 3 seeds per split, and each centre's slides are tested by models trained without that centre. Each split has its own full-bag gated ABMIL (Virchow2), trained without centre $c$, and its selector is fitted, without labels, to the attention of that split's seed-0 model on the training slides of the other four centres. The model is then read unchanged on the $20\times$ tiles its split's selection picks (15 runs). At $D$ on the unseen centre, the selection minus its full bag is $+0.003$ $[-0.019,+0.024]$ ($[-0.033,+0.042]$ with the two-level interval over splits), and the selection minus as many random tiles is $+0.152$ $[+0.120,+0.191]$ (within-run patient bootstrap, the same resample for both readers of a difference, 10{,}000 draws). The in-distribution control, models that saw every centre in training read on the same slides, gives $-0.018$ $[-0.043,+0.004]$ and $+0.136$ $[+0.100,+0.180]$. The selection is therefore not shown to lose more on a centre that none of its parts was fitted on than on a seen one. It loses less there as a point estimate, and both intervals include zero.

\paragraph{The performance-first rule out of sample.} Registered before any of these reads, this test runs the performance-first rule of Sec.~\ref{app:rule} unchanged, from the source of the frozen rule, on models and slides that took no part in building it. On the held-out tasks it reads the 45 frozen models of Table~\ref{tab:heldoutlf} (three tasks, three aggregators, 5 runs each), re-read at the 32 budgets with training and validation predictions saved, and each budget is chosen on that model's own training and validation slides. For the unseen centre it reads the 15 CAMELYON17 models trained with one centre left out, each budget chosen on that split's training and validation slides and scored on the left-out centre, and the 15 in-distribution controls that see every centre in training. On SLN-Breast it reads the 25 frozen CAMELYON17 gated ABMIL models, each at the budget already chosen for it on CAMELYON17, so SLN-Breast is not used to choose. The registered criterion is that of the design tasks, the rule within $0.005$ of the full bag, mean over the runs of a setting. The two transfers between CAMELYON16 and CAMELYON17 are not part of this registration (Table~\ref{tab:e8}).
\begin{table*}[h]\centering\scriptsize\setlength\tabcolsep{2.5pt}\adjustbox{max width=\linewidth}{\begin{tabular}{@{}lcccccccl@{}}\toprule
Setting & runs & full bag & performance-first rule & minus full bag [95\% CI] & worst run & runs within $0.005$ & cost & budgets chosen \\ \midrule
\multicolumn{9}{@{}l}{\emph{Held-out tasks}} \\
\quad PANDA cancer, gated ABMIL & 5 & 0.992 & 0.992 & $0.000$ $[-0.001,-0.000]$ & $-0.001$ & 5/5 & 45.7\% & (100,10)$\times$3, (100,20)$\times$2 \\
\quad PANDA cancer, DSMIL & 5 & 0.992 & 0.992 & $0.000$ $[-0.000,0.000]$ & $0.000$ & 5/5 & 56.4\% & (100,10)$\times$2, (100,20)$\times$2, (100,40) \\
\quad PANDA cancer, TransMIL & 5 & 0.990 & 0.990 & $0.000$ $[-0.000,+0.001]$ & $-0.001$ & 5/5 & 54.3\% & (100,20)$\times$4, (100,10) \\
\quad PANDA ISUP, gated ABMIL & 5 & 0.938 & 0.935 & $-0.003$ $[-0.003,-0.002]$ & $-0.004$ & 5/5 & 37.1\% & (100,10)$\times$5 \\
\quad PANDA ISUP, DSMIL & 5 & 0.939 & 0.938 & $-0.001$ $[-0.002,-0.001]$ & $-0.002$ & 5/5 & 49.8\% & (100,20)$\times$3, (100,10)$\times$2 \\
\quad PANDA ISUP, TransMIL & 5 & 0.931 & 0.929 & $-0.001$ $[-0.002,-0.000]$ & $-0.003$ & 5/5 & 58.2\% & (100,20)$\times$5 \\
\quad KIRC grade, gated ABMIL & 5 & 0.797 & 0.805 & $+0.007$ $[0.000,+0.015]$ & $0.000$ & 5/5 & 60.5\% & full bag$\times$2, (500,100), (all,250), (all,500) \\
\quad KIRC grade, DSMIL & 5 & 0.801 & 0.803 & $+0.002$ $[-0.003,+0.007]$ & $0.000$ & 5/5 & 86.6\% & full bag$\times$4, (all,250) \\
\quad KIRC grade, TransMIL & 5 & 0.752 & 0.757 & $+0.005$ $[-0.010,+0.019]$ & $-0.008$ & 4/5 & 67.5\% & full bag$\times$3, (all,5), (all,250) \\
\multicolumn{9}{@{}l}{\emph{CAMELYON17 centres} (gated ABMIL)} \\
\quad centre left out of training & 15 & 0.952 & 0.952 & $0.000$ $[-0.002,+0.003]$ & $-0.008$ & 14/15 & 82.2\% & full bag$\times$10, (all,500)$\times$3, (all,100), (all,250) \\
\quad same slides, centre seen in training & 15 & 0.954 & 0.953 & $-0.001$ $[-0.006,+0.003]$ & $-0.021$ & 14/15 & 86.9\% & full bag$\times$12, (500,100), (all,250), (all,500) \\
\multicolumn{9}{@{}l}{\emph{External cohort} (gated ABMIL, budget chosen on CAMELYON17)} \\
\quad SLN-Breast & 25 & 0.990 & 0.990 & $0.000$ $[-0.000,0.000]$ & $0.000$ & 25/25 & 86.7\% & full bag$\times$15, (all,500)$\times$9, (all,250) \\
\bottomrule\end{tabular}}
\caption{\textbf{The performance-first rule out of sample} (Virchow2, test AUROC, mean over runs). Intervals are from the within-run patient bootstrap with the models fixed, one resample shared by the runs of a setting (10{,}000 draws, CAMELYON17 stratified by centre), and from the slide bootstrap on SLN-Breast. Cost is in encoder evaluations, thumbnail included, as \% of the full bag's. In the budgets $(K,b)$, all means every region. A bound printed as $-0.000$ is below zero.}\label{tab:ruleoos}\end{table*}
The criterion is met in all 12 settings, the 11 out of sample and the control on CAMELYON17 centres seen in training. The rule minus the full bag is $-0.003$ to $+0.007$, and 97 of 100 runs are within $0.005$ of their own full bag, 83 of 85 out of sample. On the PANDA tasks the rule never keeps the full bag and reads $37$ to $58\%$ of the encoder evaluations. Its loss there is at most $0.003$, but on PANDA ISUP the interval excludes zero for all three aggregators, and on PANDA cancer for gated ABMIL. These slides are biopsies with few regions, where even $(100,10)$ reads $37\%$ (gated ABMIL on PANDA ISUP, which chose it in every run). On KIRC grade the rule keeps the full bag in 9 of 15 runs, at $60$ to $87\%$. On the CAMELYON17 centres it keeps the full bag in 10 and 12 of 15 runs, as it does on CAMELYON17 itself, and on SLN-Breast the CAMELYON17 choices cost $87\%$ and are less than $0.001$ below the full bag in every run. The 3 runs outside $0.005$ are TransMIL on KIRC grade ($-0.008$), the centre left out ($-0.008$) and the centre seen in training ($-0.021$). The criterion holds for the mean over a setting's runs, not for every run. Keeping the full bag would also meet it, so where the rule mostly keeps the full bag the test shows that it does not lose, not that it saves.

\paragraph{BRCA at every operating point.} Table~\ref{tab:brcaall} adds the rows of Table~\ref{tab:main} of the main paper that Table~\ref{tab:mainsix} does not give for BRCA. On BRCA, random $20\times$ tiles at the same encoder cost are as good as our selection, at $D$ and at $(1000,10)$.
\begin{table}[h]\centering\scriptsize\setlength\tabcolsep{3pt}\adjustbox{max width=\linewidth}{\begin{tabular}{@{}lccccc@{}}\toprule
Frozen gated ABMIL & AUROC & $-$ full bag & ours $-$ random [95\% CI] & F\% & $\times$ \\ \midrule
full bag & 0.974 & & & 100 & 1.0 \\
our selection, $D$ & 0.969 & $-0.005$ & $-0.004$ $[-0.012,+0.002]$ & 4.1 & 21.2 \\
random tiles, FLOPs of $D$ & 0.974 & & & 4.1 & 25.3 \\
our selection, $(1000,10)$ & 0.970 & $-0.005$ & $-0.004$ $[-0.011,+0.003]$ & 6.3 & 14.3 \\
random tiles, FLOPs of $(1000,10)$ & 0.974 & & & 6.3 & n/a \\
performance-first rule & & $-0.001$ & & 23 & 4.4 \\
\bottomrule\end{tabular}}
\caption{\textbf{BRCA rows of Table~\ref{tab:main}} (Virchow2, frozen gated ABMIL, test AUROC, mean over 25 runs). Random tiles here match the encoder FLOPs of our selection, as in Table~\ref{tab:main}, so they read more tiles than the count-matched random tiles of Table~\ref{tab:frozen}. The column $-$ full bag is that of Tables~\ref{tab:frozen} and~\ref{tab:k1000design}, and F\% and $\times$ are those of Tables~\ref{tab:costflops} and~\ref{tab:costtime}. Random tiles at the FLOPs of $(1000,10)$ were not timed. Intervals are from the within-run patient bootstrap. Differences are of unrounded means, so they can differ by $0.001$ from the difference of the printed AUROCs.}\label{tab:brcaall}\end{table}
The performance-first rule can also be run on random tiles, where at each budget of the ladder it reads as many random $20\times$ tiles as the encoder cost of our selection buys (each run the mean over 5 draws). On BRCA the rule on our selection costs $23.4\%$ of the full bag and is $0.001$ below it, and the rule on random tiles costs $14.9\%$ and is less than $0.001$ below it, with all 25 runs saving in both cases. Random minus ours is $-8.5$ points of cost $[-22.3,+4.7]$ and $+0.001$ AUROC $[-0.001,+0.002]$ ($95\%$ intervals from a bootstrap over runs, 10{,}000 draws). Both intervals include zero, so on BRCA the ranking is not shown to add anything under the rule, in cost or in accuracy.

\paragraph{$(1000,10)$ against random tiles out of sample.} Table~\ref{tab:k1000rand} gives the column $(1000,10)$ minus random of Table~\ref{tab:general} of the main paper. In five rows the value is recomputed from the saved predictions of the runs, with as many random $20\times$ tiles as $(1000,10)$ reads, and it reproduces the printed value. The two cross-cohort runs saved no random tiles at $(1000,10)$. Their printed values, $+0.167$ and $+0.184$, were derived from the other columns of that table as $(D-\text{random})+((1000,10)-\text{full})-(D-\text{full})$, which compares $(1000,10)$ against as many random tiles as $D$ reads, fewer than $(1000,10)$ reads. These two values therefore measure a different comparison from the other five, and no interval was computed for them, so the mark of an interval that excludes zero on them in the main paper is not backed by a computed interval. For these two rows the comparison with an interval is the one at $D$, $+0.146$ $[+0.095,+0.208]$ and $+0.170$ $[+0.115,+0.232]$ (Table~\ref{tab:lfcross}). Intervals for the $(1000,10)$ column are not computed in this supplement.
\begin{table}[h]\centering\scriptsize\setlength\tabcolsep{3pt}\adjustbox{max width=\linewidth}{\begin{tabular}{@{}lcccl@{}}\toprule
Setting & runs & $D-$random & $(1000,10)-$random & source of the $(1000,10)$ value \\ \midrule
PANDA cancer & 5 & $+0.011$ & $+0.011$ & recomputed \\
PANDA ISUP & 5 & $+0.011$ & $+0.010$ & recomputed \\
KIRC grade & 5 & $+0.026$ & $+0.025$ & recomputed \\
SLN-Breast, C17 models & 25 & $+0.052$ & $+0.059$ & recomputed \\
C17 centre left out & 15 & $+0.152$ & $+0.168$ & recomputed \\
C16 selector, C17 models & 25 & $+0.146$ & $+0.167$ & derived, random tiles at the count of $D$ \\
C17 selector, C16 models & 25 & $+0.170$ & $+0.184$ & derived, random tiles at the count of $D$ \\
\bottomrule\end{tabular}}
\caption{\textbf{Our selection minus random tiles on the settings of Table~\ref{tab:general}} (frozen gated ABMIL, Virchow2, test AUROC, mean over runs). Recomputed values use as many random $20\times$ tiles as the selection reads at that budget. Derived values follow from the other columns of Table~\ref{tab:general} and compare $(1000,10)$ against as many random tiles as $D$ reads.}\label{tab:k1000rand}\end{table}

\end{document}